\pdfoutput=1

\documentclass[11pt]{article}

\PassOptionsToPackage{table,xcdraw,dvipsnames}{xcolor}

\usepackage[final]{acl}

\usepackage{times}
\usepackage{tcolorbox}
\usepackage{tabularx}

\usepackage{latexsym}
\usepackage{booktabs}
\usepackage{multirow}
\usepackage{enumitem}
\usepackage[T1]{fontenc}
\usepackage{soul} 
\usepackage{subcaption}

\definecolor{mygreen}{HTML}{bfffbf}
\definecolor{mycyan}{HTML}{b7e5fa}

\newcommand{\hlgreen}[1]{\colorbox{mygreen}{#1}}
\newcommand{\hlcyan}[1]{\colorbox{mycyan}{#1}}

\usepackage[utf8]{inputenc}

\usepackage{microtype}

\usepackage{inconsolata}

\usepackage{graphicx}

\title{BigTokDetect: A Clinically‑Informed Vision–Language Modeling Framework for Detecting Pro‑Bigorexia Videos on TikTok}

\author{Minh Duc Chu\textsuperscript{1}, Kshitij Pawar\textsuperscript{1}, Zihao He\textsuperscript{1}, 
Roxanna Sharifi\textsuperscript{2}, 
Ross Sonnenblick\textsuperscript{3},\\
{\bf Magdalayna Curry\textsuperscript{4}},
{\bf Laura D'Adamo\textsuperscript{3}},
{\bf Lindsay Young\textsuperscript{4}},
{\bf Stuart B Murray\textsuperscript{5}},
{\bf Kristina Lerman\textsuperscript{1}}\\
        \textsuperscript{1}USC Information Sciences Institute \\ 
        \textsuperscript{2}Keck School of Medicine, USC \\
        \textsuperscript{3}Department of Clinical Psychology, Drexel University \\
        \textsuperscript{4}Annenberg School for Communication and Journalism, USC \\
        \textsuperscript{5}Department of Psychiatry and Biobehavioral Sciences, UCLA\\
        \texttt{\textsuperscript{1}mhchu@usc.edu}
    }

\begin{document}
\maketitle
\begin{abstract}
Social media platforms face escalating challenges in detecting harmful content promoting muscle dysmorphic behaviors and cognitions (bigorexia), which disproportionately affects adolescent males and evades moderation by camouflaging as legitimate fitness advice. We address this challenge with \textsc{BigTokDetect}, a clinically informed framework for identifying pro-bigorexia content on TikTok. We introduce \textsc{BigTok}, the first expert-annotated multimodal benchmark dataset of over 2,200 TikTok videos labeled by clinical psychiatrists across five categories and eighteen fine-grained subcategories. Comprehensive evaluation of state-of-the-art vision-language models (VLMs) reveals that while commercial zero-shot models achieve the highest accuracy on broad primary categories, supervised finetuning allows smaller open-source models to achieve better fine-grained subcategory detection. Ablation studies demonstrate that multimodal fusion improves performance by 5-15\%, with video features providing the most discriminative signals. These findings support a grounded moderation approach, automating detection of explicit harms while flagging ambiguous content for human review, and establish a scalable framework for harm mitigation in emerging mental health domains.
\end{abstract}

\section{Introduction}

\begin{figure*}[!ht]
  \centering
  \includegraphics[width=2\columnwidth]{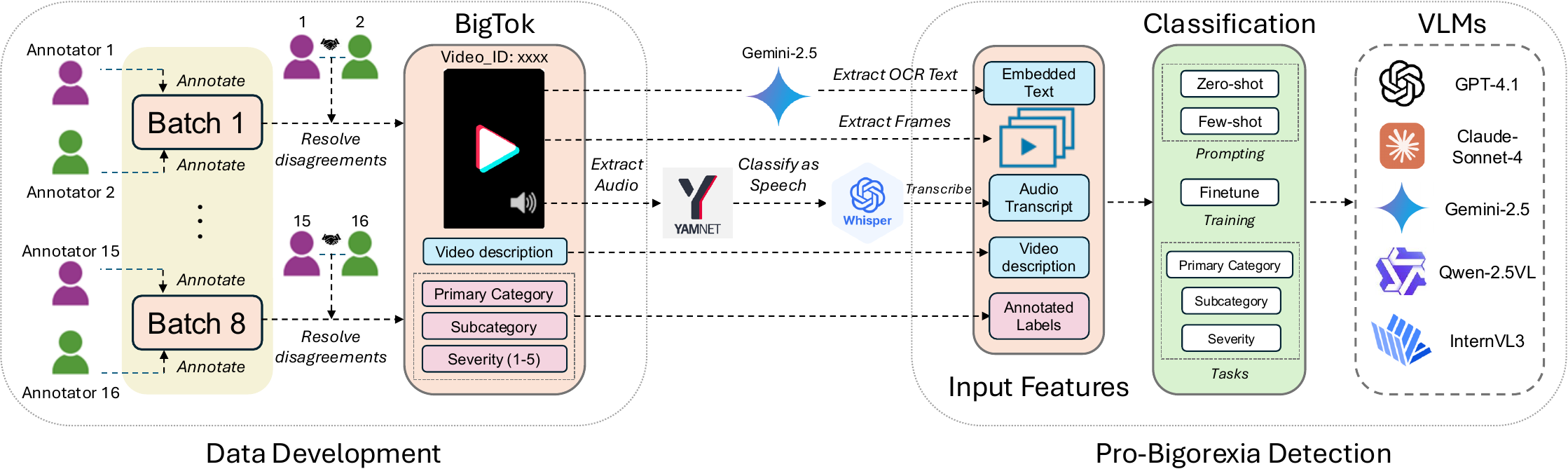}
  \caption{\textsc{BigTok} pipeline overview. Left: Expert annotation process with dual annotation and consensus resolution. Right: Multimodal feature extraction (visual, audio, text) and classification evaluation across VLMs using zero-shot, few-shot, and finetuning approaches for primary category, subcategory, and severity prediction tasks.}
  \label{fig:pipeline}
\end{figure*}

Social media platforms face mounting challenges in detecting harmful content that significantly impacts user mental health~\cite{chancellor2020methods,gorwa2020algorithmic}. Contemporary platforms like TikTok algorithmically amplify content promoting unrealistic body ideals~\cite{becker2004television,minadeo2022tiktok}, yet much of the harmful material exists in gray areas where legitimate discussions intersect with unsafe messages~\cite{gillespie2018custodians}. These narratives emerge through complex multimodal combinations, which include visual imagery, audio cues, and textual descriptions that traditional text-based detection systems do not adequately capture~\cite{kiela2020supervisedmultimodalbitransformersclassifying, Kiela2020}. Effective automated moderation requires comprehensive multimodal approaches that can parse nuanced signals across video, audio, and text~\cite{gimenogómez2024readingframesmultimodaldepression}.

Pro-bigorexia content, material promoting muscularity as an aspirational goal, exemplifies these multimodal detection challenges. This  content promotes compulsive muscle-building behaviors that disproportionately affect adolescent males~\cite{Pope1997,Mitchison2022}, yet remains severely understudied in computational research compared to traditional eating disorder detection~\cite{chancellor2020methods}. Pro-bigorexia material often masquerades as legitimate fitness content through subtle combinations of muscularity displays, extreme workout demonstrations, and coded supplement language~\cite{murray2017enigma,kamkari2025manosphere}. The visual similarity between harmful and beneficial fitness content, combined with evolving jargon, renders current detection approaches fundamentally inadequate.

Current automated detection systems face critical limitations while moderating such content. Foundation models trained on general corpora lack domain-specific knowledge to recognize subtle clinical markers and coded language patterns~\cite{murray2017enigma}. Systems relying on keyword filtering and text-based signals miss critical visual and behavioral cues embedded in video content~\cite{Kiela2020,gorwa2020algorithmic}. The dynamic nature of social media, where creators deliberately evolve language to evade detection, poses another  challenge to automated approaches~\cite{gillespie2018custodians}. Effective detection requires integrated analysis of visual behavioral cues, audio transcripts, and textual context, capabilities that remain underexplored in computational mental health research~\cite{gimenogómez2024readingframesmultimodaldepression}.

To address these detection challenges, we make the following contributions:
\begin{itemize}[noitemsep,topsep=0pt]
  \item We develop the first clinically-informed taxonomy for automated detection of pro-bigorexia content, establishing fine-grained categories spanning body image, nutrition, supplement abuse, exercise practices, and masculinity that enable systematic computational analysis.
  \item We introduce \textsc{BigTok}, a multimodal benchmark dataset of over 2,200 expert-annotated TikTok videos that enables robust evaluation of vision-language models (VLMs) on challenging harmful content detection tasks.
  \item We construct \textsc{BigTokDetect}, a detection framework that achieves state-of-the-art (SOTA) performance using proprietary and open-source VLMs, with our best models reaching 82.9\% accuracy on primary category classification and 69.0\% on fine-grained subcategory detection through finetuning.
  \item We demonstrate through comprehensive ablation studies that multimodal fusion improves detection performance by 5-15\% over text-only approaches, with video features providing the most discriminative signals for identifying subtle pro-bigorexia cues.
\end{itemize}

Figure~\ref{fig:pipeline} summarizes our comprehensive approach to multimodal pro-bigorexia detection. Our results demonstrate that clinically-informed VLMs can effectively identify subtle harmful content that evades traditional detection systems. Through systematic evaluation across multiple model architectures and training paradigms, we achieve substantial improvements over existing methods, establishing a new paradigm for detecting nuanced harmful content on social media platforms. The \textsc{BigTokDetect} framework provides both the computational tools and reproducible methodology needed to address the growing challenge of automated content moderation in mental health domains.

\section{Related Work}

\paragraph{Muscle Dysmorphic Disorder on Social Media}
Muscle dysmorphic disorder, or ``bigorexia,'' involves preoccupation with insufficient muscularity, driving compulsive behaviors including exercise, rigid dieting, and supplement abuse~\cite{Pope1997, Cooper2020}. Frequent exposure to muscularity-oriented TikTok and Instagram content, including body-transformation videos, supplement promotions, and steroid use, has been linked to elevated rates of probable muscle dysmorphia~\cite{Mitchison2022, ganson2025associations}. Hypermasculine online subcultures (e.g., the ``manosphere'') intensify these pressures through toxic social comparisons and body-optimization trends~\cite{kamkari2025manosphere}. Despite this prevalence, both psychological research and platform moderation remain focused on anorexia and bulimia, leaving bigorexia largely unmoderated~\cite{Lookingbill2023}.

\paragraph{Multimodal Mental Health and Body Image Content Detection}

Automated detection of body image content has overwhelmingly targeted anorexia and bulimia~\cite{Chancellor2017,chancellor2020methods}, relying primarily on keyword filters and text-only classifiers despite evidence that vision-language fusion substantially improves detection of subtle harmful imagery~\cite{Kiela2020}. Short-form video platforms have become dominant for youth mental health discourse~\cite{corey2022}, yet eating disorder detection remains predominantly text-based~\cite{wang2017detecting,merhbene2024investigating}. Even dedicated TikTok corpora for eating disorders research exclude muscle dysmorphia labels~\cite{bickham2025edtokdataseteatingdisorder,donati-etal-2023-building}, while existing qualitative analyses of pro-muscularity forums~\cite{murray2015} lack the scale needed for robust computational detection. Modern vision-language models—Flamingo~\cite{alayrac2022flamingovisuallanguagemodel}, InstructBLIP~\cite{li2023blip2bootstrappinglanguageimagepretraining}, GPT-4V~\cite{OpenAI2023}—enable end-to-end multimodal detection but lack clinically grounded training data for specialized tasks. Existing mental health datasets rely on crowdsourced annotations rather than clinical expertise, potentially missing subtle markers needed to distinguish legitimate fitness content from harmful pro-bigorexia messaging that uses coded language around supplements and steroids~\cite{murray2017enigma}. To our knowledge, no large-scale expert-annotated datasets exist for this increasingly prevalent content type.

\section{Pro-bigorexia Content Taxonomy}
We develop an empirically grounded taxonomy for pro-bigorexia content characterization, following an iterative four-step process. First, we compile a seed list of keywords from fitness, bodybuilding, exercise, and diet domains, supplemented by terms from foundational muscle dysmorphia research and popular media reports~\cite{Cafri2008,sanchez2024feelingsbodiesemotionsdiet, WP2023BodyDysmorphia}. 
Then, using these keywords, we retrieve 500 TikTok videos over 40 hours, and open-code emergent patterns of bigorexia, such as body-check demonstrations, supplement endorsements, extreme workout narratives, diet tutorials). 
Next, we draft preliminary categories based on the DSM-5 clinical criteria~\cite{APA2013DSM5}, validated measures such as the Drive for Muscularity Scale~\cite{mccreary2000drive} and the Muscle Dysmorphic Disorder Inventory~\cite{zeeck2018muscle}. 
Finally, we convene with clinical experts specializing in eating disorders and body image psychopathologies to refine categories, clarify specialized or coded language (especially around steroids and supplements), and disambiguate edge cases. 
We iterate to discover new potential categories and keywords.

Our final taxonomy for pro-bigorexia TikTok content consists of five primary categories: \textit{Relationship to Body}, \textit{Relationship to Food}, \textit{Relationship to Exercise}, \textit{Supplements}, and \textit{Masculinity} (Table~\ref{tab:taxonomy_full}, Appendix~\ref{sec:cat_definition}). Each is further divided into clinically grounded subcategories. For each category and subcategory, we provide clear definitions, representative keywords, practical guidelines for recognizing problematic content, and sample videos. 
The 
taxonomy facilitates a consistent, multimodal annotation of overt and implicit signals of pro-bigorexia behaviors. 

\section{BigTok: Expert-Curated Multimodal Dataset Construction}

\subsection{Data Collection}
We source our corpus from TikTok via its official API~\cite{TikTokResearchAPI}, querying videos from January 2019 to January 2025 to capture pre- and post-pandemic trends. Guided by domain experts, we curate 40 high-precision query terms mapped to taxonomy subcategories (Table~\ref{tab:taxonomy_full}, Appendix~\ref{sec:cat_definition}), retrieving up to 1,000 videos per term, from which we randomly sample 2,400 videos for annotation. To protect user privacy, only video content and captions are exposed to annotators (anonymization steps are described in Appendix~\ref{sec:data_collection}). To ensure robust supervised classification, we construct a negative control group (the \textit{Irrelevant} category) via two mechanisms: (1) proactive sampling using 42 generic trending hashtags (e.g., \#viral, \#travel) from TikTok's Creative Center~\cite{TikTokCreativeCenter}; and (2) re-assigning videos flagged by experts during annotation as unrelated to bigorexia. This strategy ensures the model learns to distinguish harmful content from both general entertainment and benign daily activities, following established supervised classification practices~\cite{Kiela2020}.

\subsection{Annotating Bigorexia Content}

\paragraph{Annotator Recruitment and Demographics}
We recruit 16 subject matter experts (13 females and 3 males), including licensed clinical psychologists, social workers, and doctoral candidates with research and clinical experience in eating and body image disorders (Table~\ref{tab:annotators}, Appendix~\ref{sec:annotator} for profiles). Their combined expertise in empirical research and direct patient care ensures that our annotations are both theoretically grounded and have clinical relevance. 

\paragraph{Annotation Instructions}
We base our annotation interface on Amazon Mechanical Turk via an invitation‐only pool (Appendix~\ref{sec:annon_platform}). Annotators are instructed to select the first (mandatory) primary–subcategory pair ($[t_1, s_1]$) and the second (optional) primary–subcategory pair ($[t_2, s_2]$).  
Every video is also rated for harm on a discrete 5-point Likert scale with 0.5-point increments (1=not harmful and 5=very harmful). The annotation interface is shown in Appendix~\ref{sec:annon_instruction}.
 
\paragraph{Annotator Training} As a pilot study, each annotator independently labels a set of 20-30 videos to validate taxonomy coverage, identify potential edge cases, and familiarize themselves with the annotation interface. We monitor disagreements and encourage detailed note‐taking. Annotators then join a group session to discuss flagged issues, collaboratively annotate selected videos while explaining their reasoning, until consensus is reached.

\paragraph{Batchwise Annotation}
We split the dataset into 8 batches of \textasciitilde300 videos. 
For each batch, we filter out videos that are not available (e.g., they have been deleted) or are not in English. Then the batch is assigned to two different annotators, so that every video receives two independent annotations. 
To assess agreement between a pair of annotators A and B, we compare all combinations of labels and determine the level of agreement:
\begin{itemize}[noitemsep,topsep=0pt]
  \item \textbf{Perfect agreement:} both the first and second primary–subcategory pairs match exactly across annotators, i.e.
  $\displaystyle \forall i\in\{1,2\}:\;[t_i,s_i]^A = [t_i,s_i]^B$
  \item \textbf{Strong agreement:} one primary–subcategory pair matches exactly across annotators, i.e.
  $\displaystyle \exists i\in\{1,2\}:\;[t_i,s_i]^A = [t_i,s_i]^B$
  \item \textbf{Weak agreement:} at least one primary category matches across any labels, but the subcategories differ, i.e. $\displaystyle \exists i\in\{1,2\}:\;t_i^A = t_i^B \;\wedge\; s_i^A \neq s_i^B$
  \item \textbf{Disagreement:} no common primary category across labels, i.e. $\displaystyle \forall i\in\{1,2\},\;t_i^A \neq t_i^B \;\wedge\; s_i^A \neq s_i^B$.
\end{itemize}
We then flag examples with weak agreement or disagreement on pro-bigorexia detection and examples where severity scores differ by more than two points. The two annotators are invited to a structured discussion to articulate their reasoning, review cases that need consultation, address disagreements, and eventually aim to reach consensus, although consensus is not mandatory. 
After the discussion, each annotator independently re‐annotates the flagged videos. Further details on this procedure are provided in Appendix ~\ref{sec:annon_proc}.

\paragraph{Label Aggregation}  
While each video may receive multiple label pairs (primary-subcategory), to ensure consistent and reproducible evaluation, we assign a single label pair to each video. For videos with perfect agreement, we randomly choose one of the label pairs. For strong agreement, we adopt the shared primary–subcategory pair; for weak agreement, we choose the mandatory (first) label over the optional (second) one; remaining disagreements are resolved by random tie‐breaking. The final harm severity score is the average of the two annotators’ ratings. This simplification reflects current limitations of LLMs in reliable multi‐label classification~\cite{ma2025largelanguagemodelsmultilabel}.   

We assess inter‐rater reliability with Cohen’s~$\kappa$ ~\cite{Cohen1960} for two annotators. Initial agreement is moderate (strong~$\kappa=0.43$-$0.69$; weak~$\kappa=0.59$-$0.83$; Table~\ref{tab:kappa}). After structured consensus discussions and re‐annotation, reliability increases substantially (strong~$\kappa=0.58$-$0.81$; weak~$\kappa=0.78$-$0.94$), ensuring that our iterative process yields high‐quality, consistent labels. 
All Cohen’s~$\kappa$ values for each batch are shown in Table \ref{tab:kappa} (Appendix \ref{sec:annon_stats}). Severity ratings showed moderate agreement (ICC = 0.413, 69.4\% within one point), typical for subjective clinical assessments and highlighting the complexity of harm evaluation in this domain (Figure~\ref{fig:sev_diff}, Appendix~\ref{sec:annon_stats}).

\subsection{Data Statistics}

\begin{table}[t]
\centering
\resizebox{0.5\textwidth}{!}{%
\begin{tabular}{l r r r}
\toprule
\textbf{Primary Category} & \textbf{Train} & \textbf{Test} & \textbf{Total} \\
\midrule
Relationship to Body & 489 & 98 & 587 \\
Relationship to Food & 310 & 98 & 408 \\
Relationship to Exercise & 450 & 98 & 548 \\
Supplement Abuse & 201 & 98 & 299 \\
Relationship to Masculinity & 66 & 98 & 164 \\
Irrelevant & 450 & 98 & 548 \\
\midrule
\textbf{Total} & \textbf{1,966} & \textbf{588} & \textbf{2,554} \\
\bottomrule
\end{tabular}
}

\vspace{1em}

\resizebox{0.5\textwidth}{!}{%
\begin{tabular}{l r r r}
\toprule
\textbf{Subcategory} & \textbf{Train} & \textbf{Test} & \textbf{Total} \\
\midrule
Muscularity Self‐objectification & 170 & 34 & 204 \\
Leanness Self‐objectification & 59 & 34 & 93 \\
Muscle Dissatisfaction & 72 & 34 & 106 \\
\cmidrule(lr){1-4}
Rigid Food Rules & 144 & 34 & 178 \\
Unsafe Food & 59 & 34 & 93 \\
Cheat Meals & 60 & 34 & 94 \\
\cmidrule(lr){1-4}
Excessive Exercise & 129 & 34 & 163 \\
Predebting Exercise & 28 & 4 & 32 \\
Maladaptive Coping & 82 & 34 & 116 \\
Exercise‐Induced Functional Impairment & 43 & 6 & 49 \\
Toxic Motivation & 102 & 34 & 136 \\
\cmidrule(lr){1-4}
Anabolic Steroids & 61 & 34 & 95 \\
Legal APEDs & 56 & 10 & 66 \\
Hormone Therapy & 77 & 34 & 111 \\
\cmidrule(lr){1-4}
Relationship to Masculinity & 130 & 34 & 164 \\
\cmidrule(lr){1-4}
Irrelevant & 200 & 34 & 234 \\
\midrule
\textbf{Total} & \textbf{1,472} & \textbf{462} & \textbf{1,934} \\
\bottomrule
\end{tabular}%
}
\caption{\textsc{BigTok} Classification Benchmark statistics showing the final train/test split distribution after data processing.}
\label{tab:label_distribution}
\end{table}

Our \textsc{BigTok} dataset originates from a raw core set of 2,210 expert-annotated TikTok videos (median duration 20.6 seconds, 59.8\% under 30 seconds). To construct robust classification benchmarks, we apply a strict preprocessing pipeline detailed in Section~\ref{sec:annon_proc} (Appendix), which includes removing corrupted or non-English files and supplementing the dataset with verified irrelevant videos sampled via popular keywords (Table~\ref{tab:taxonomy_full}). Table~\ref{tab:label_distribution} presents the final train/test split statistics resulting from this processing. Specifically, the primary category benchmark (Task 1) comprises 2,554 videos split into 1,966 training and 588 balanced test samples, while the subcategory benchmark (Task 2) utilizes a relevance-filtered subset of 1,934 videos (1,472 training, 462 test). Within the processed primary categories (excluding \textit{Irrelevant}), \emph{Relationship to Body} (587 videos) and \emph{Relationship to Exercise} (548 videos) represent the largest classes. At the subcategory level, \emph{Muscularity Self-objectification} (204 videos) predominates, reflecting the prevalence of physique-display content on fitness-oriented TikTok. Examples from the final dataset appear in Tables~\ref{tab:type_classification} and  \ref{tab:subtype} in Appendix~\ref{sec:cat_definition}, with additional annotation statistics provided in Appendix~\ref{sec:annon_stats}. The raw annotation counts are shown in Table~\ref{tab:annotation_count} in the Appendix.

\section{Classifying Multimodal Pro-Bigorexia Content via VLMs}

To develop an automated detection system for pro-bigorexia content, we apply SOTA VLMs to this challenging multimodal classification task. We leverage both proprietary and open-source VLMs through zero-shot prompting, few-shot learning, and finetuning approaches to achieve high performance in identifying subtle pro-bigorexia behaviors across video, audio, and text modalities.

\subsection{Task Definition}
\label{sec:task_diff}

We define three evaluation tasks for TikTok video classification. \textbf{Task 1: Primary Category Classification} involves predicting one of the five primary categories of pro-bigorexia content (\textit{Body}, \textit{Food}, \textit{Supplements}, \textit{Exercise}, \textit{Masculinity}), or \textit{Irrelevant}. \textbf{Task 2: Subcategory Classification} involves predicting the specific subcategory within the selected primary category. \textbf{Task 3: Severity Estimation} involves predicting a continuous severity score on a 1-5 scale, where 1 indicates no harm and 5 indicates extreme harm. 

For each task, we split the data into training and test sets at a 3:1 ratio using stratified sampling. The test set is strictly balanced across all classes, while the training set is adjusted by downsampling the majority categories to mitigate class imbalance. The sampling procedures and statistics after sampling are provided in Appendix~\ref{sec:model_data}.

\subsection{Models}

Given the multimodal nature of our classification task, involving text from captions, audio transcripts, images, and video content, we require SOTA models designed for multimodal data, particularly video understanding. We evaluate three commercial API‐based VLMs: GPT‑4.1~\cite{openai2025gpt41} and Claude-Sonnet-4~\cite{anthropic2025systemcard} (both process images), while Gemini-2.5-Flash~\cite{comanici2025gemini25pushingfrontier} natively accepts video input.
We also evaluate two open-source models: Qwen2.5-VL~\cite{qwen2024qwen25}, and InternVL3~\cite{Zhu2025InternVL3}, which uses Qwen2.5 pre-trained base models and Variable Visual Position, which achieves strong performance on video benchmarks. 

Model sizes and versions are listed in Table~\ref{tab:models}, Appendix~\ref{sec:setup}. Due to computational constraints, we focus on small and medium-sized models.
For open-source models, we set \textit{temperature} = 0.1 to encourage deterministic outputs. For commercial API models, we utilize the provider defaults (\textit{temperature} = 1.0). To ensure the validity of this choice, we conducted a robustness check comparing $T=1.0$ vs $T=0.1$ for all commercial models (see Appendix~\ref{sec:robustness}). We observe negligible performance differences, confirming that our results are robust to hyperparameter variations.
Implementation details and hardware specifications are provided in Appendix~\ref{sec:setup}.
Details about the train and test datasets are in Appendix~\ref{sec:model_data}.

\subsection{Input Features}
To capture the full range of pro‑bigorexia signals, we extract four complementary input features:
\textbf{Visual:} We adapt visual inputs to model architectures. For Gemini-2.5-Flash and Qwen2.5-VL, which support native video processing, we provide the raw video stream. For InternVL3, GPT-4.1, and Claude-Sonnet-4, we treat video as a sequential image task. We extract 4 equally spaced frames per video (unless otherwise noted in ablation studies) to balance temporal coverage with context-window constraints.
\textbf{Audio:} We classify audio using YAMNet \cite{hershey2017cnnarchitectureslargescaleaudio} and transcribe detected speech via Whisper \cite{radford2022robustspeechrecognitionlargescale}. To prevent bias, we retain all videos regardless of audio type: 40.2\% contain intelligible speech, while the remaining 59.8\% (music or ambient noise) are assigned empty transcripts.
\textbf{On‑Screen Text:} We use Gemini‑2.5‑Flash to extract overlaid text, creators' captions and annotations, since many TikTok creators rely on on‑screen text for key messages.
\textbf{Caption:} We include the original TikTok description (user caption and hashtags) from the video metadata.

\begin{table*}[!ht]
\centering
\small
\begin{tabular}{l|l|cccc|cccc}
\toprule
\textbf{Model}                   & \textbf{Training} & \multicolumn{4}{c|}{\textbf{Primary Category}} & \multicolumn{4}{c}{\textbf{Subcategory}} \\
\cmidrule(lr){3-6}\cmidrule(lr){7-10}
                        &          & \textbf{Acc.} & \textbf{P$_{\mathrm{m}}$} & \textbf{R$_{\mathrm{m}}$} & \textbf{F1$_{\mathrm{m}}$} & \textbf{Acc.} & \textbf{P$_{\mathrm{m}}$} & \textbf{R$_{\mathrm{m}}$} & \textbf{F1$_{\mathrm{m}}$} \\
\midrule
GPT‐4.1                 & Zero-shot & 0.796 & 0.808 & 0.792 & 0.792 & 0.652 & 0.639 & 0.503 & 0.532 \\
                        & Few-shot  & 0.813 & 0.820 & 0.813 & 0.813 & 0.680 & 0.679 & 0.643 & 0.639 \\
\cmidrule(lr){1-10}
Claude‐Sonnet‐4         & Zero-shot & \cellcolor{green!25}\textbf{0.829} & \cellcolor{green!25}\textbf{0.832} & \cellcolor{green!25}\textbf{0.829} & \cellcolor{green!25}\textbf{0.827} & 0.670 & 0.679 & 0.636 & 0.640 \\
                        & Few-shot  & \cellcolor{cyan!25}\textbf{0.819} & \cellcolor{cyan!25}\textbf{0.829} & \cellcolor{cyan!25}\textbf{0.818} & \cellcolor{cyan!25}\textbf{0.818} & 0.665 & 0.674 & 0.627 & 0.632 \\
\cmidrule(lr){1-10}
Gemini‐2.5‐Flash        & Zero-shot & 0.805 & 0.807 & 0.805 & 0.805 & 0.666 & 0.665 & 0.616 & 0.621 \\
                        & Few-shot  & 0.776 & 0.782 & 0.776 & 0.775 & 0.663 & 0.645 & 0.613 & 0.621 \\
\cmidrule(lr){1-10}
Qwen2.5‐VL‐7B           & Zero-shot & 0.539 & 0.744 & 0.539 & 0.526 & 0.383 & 0.554 & 0.327 & 0.306 \\
                        & Few-shot  & 0.689 & 0.733 & 0.689 & 0.689 & 0.236 & 0.149 & 0.142 & 0.140 \\
                        & Finetuning& 0.784 & 0.810 & 0.784 & 0.776 & \cellcolor{cyan!25}\textbf{0.684} & \cellcolor{green!25}\textbf{0.687} & \cellcolor{cyan!25}\textbf{0.669} & \cellcolor{cyan!25}\textbf{0.667} \\
\cmidrule(lr){1-10}
Qwen2.5‐VL‐32B          & Zero-shot & 0.733 & 0.788 & 0.733 & 0.742 & 0.556 & 0.637 & 0.495 & 0.513 \\
                        & Few-shot  & 0.804 & 0.812 & 0.804 & 0.807 & 0.654 & 0.662 & 0.594 & 0.607 \\
                        & Finetuning& 0.776 & 0.800 & 0.776 & 0.775 & 0.658 & 0.662 & 0.647 & 0.645 \\
\cmidrule(lr){1-10}
InternVL3‐8B            & Zero-shot & 0.614 & 0.734 & 0.614 & 0.635 & 0.517 & 0.582 & 0.464 & 0.476 \\
                        & Few-shot  & 0.690 & 0.717 & 0.690 & 0.692 & 0.519 & 0.560 & 0.436 & 0.443 \\
                        & Finetuning& 0.765 & 0.789 & 0.765 & 0.758 & \cellcolor{green!25}\textbf{0.690} & \cellcolor{cyan!25}\textbf{0.686} & \cellcolor{green!25}\textbf{0.679} & \cellcolor{green!25}\textbf{0.675} \\
\cmidrule(lr){1-10}
InternVL3‐38B           & Zero-shot & 0.784 & 0.797 & 0.784 & 0.785 & 0.649 & 0.665 & 0.603 & 0.608 \\
                        & Few-shot  & 0.806 & 0.817 & 0.806 & 0.807 & 0.673 & 0.666 & 0.630 & 0.630 \\
                        & Finetuning& 0.767 & 0.795 & 0.767 & 0.758 & 0.636 & 0.540 & 0.535 & 0.528 \\
\bottomrule
\end{tabular}
\caption{Classification Macro Metrics for Primary Category and Subcategory by Model and Training Method. Highest values in each column are in \hlgreen{\textbf{green}}, second‑highest in \hlcyan{\textbf{cyan}}. P$_{\mathrm{m}}$, R$_{\mathrm{m}}$, and F1$_{\mathrm{m}}$ refer to macro precision, macro recall, and macro F1 scores.}
\label{tab:results}
\end{table*}

\subsection{Training and Inference Paradigms}

\paragraph{Zero-Shot Prompting}
We create prompts for zero‐shot classification by incorporating taxonomy definitions of the primary and subcategories and instructing the model to select the appropriate category as the label. The full zero-shot prompt templates for Tasks 1 and 2 are shown in Figure~\ref{fig:vlm_zeroshot_prompt_type} and \ref{fig:vlm_zeroshot_prompt_subtype}, Appendix~\ref{sec:prompt}. We evaluate the models on balanced test sets of 588 examples for primary classification (Task 1) and 466 examples for subcategory classification (Task 2), with each class evenly represented (Appendix~\ref{sec:model_data})

\paragraph{Few-Shot Prompting}
For few‐shot experiments, we sample a fixed set of in‐context examples from the training data: two videos per primary category (12 examples total) for Task 1 and one video per subcategory (16 examples) for Task 2 (prompt details are in Figures~\ref{fig:vlm_fewshot_prompt_type} and \ref{fig:vlm_fewshot_prompt_subtype}, Appendix). Each example is presented with its full multimodal features: video frames, audio transcript, on‐screen text, caption, and the corresponding label. We evaluate the VLMs on a balanced test set of 462 examples (Appendix~\ref{sec:model_data}).

\paragraph{Finetuning}
Due to cost constraints, we limit finetuning to open‐source VLMs. We instruction‐tune various model variants (up to 72B) on our annotated datasets. The batch size is 8, and the learning rate is 5e-5. The inference hyperparameters are similar to zero-shot and few-shot prompting (\texttt{temperature} = 0.1 and \texttt{top\_p} = 0.9). 

\subsection{Pro-Bigorexia Classification Results}
\label{sec:classification_results}

Table \ref{tab:results} shows that while large commercial models like Claude-Sonnet-4 achieve the highest zero-shot performance on broad primary categories, parameter-efficient finetuning allows smaller open-source models (e.g., InternVL3-8B) to surpass them on fine-grained subcategory detection by capturing domain-specific clinical nuances.

\paragraph{Primary vs. Subcategory Detection}
Commercial API-based models continue to lead in broad-category detection: Claude-Sonnet-4 (zero-shot) attains the highest accuracy of 0.829 and F1 of 0.827. Although few-shot open-source models (e.g., InternVL3-38B) reach competitive levels (0.806 accuracy), none yet surpass the commercial zero-shot benchmark on primary categories. However, the trend reverses for fine-grained subcategory detection tasks. InternVL3-8B (finetuned) achieves the top subcategory F1 of 0.675, outperforming commercial methods, which plateau around 0.640 F1. This demonstrates that while large-scale pretraining confers an advantage for general classification, parameter adaptation better captures nuanced subtype distinctions.

\paragraph{Impact of Model Scale}
Our experiments reveal a divergence in finetuning effectiveness relative to model scale. While finetuning dramatically boosts smaller models (7B--8B), for instance, raising Qwen2.5-VL-7B subcategory accuracy from 0.383 (zero-shot) to 0.684 (finetuned), it frequently degrades the primary classification performance of larger open-source models (32B--38B). For example, InternVL3-38B drops from 0.784 (zero-shot) to 0.767 (finetuned) on primary categories. This likely stems from overparameterization under data scarcity: our \textasciitilde1,500 training samples create a capacity mismatch for larger models, leading to overfitting, whereas 7B--8B models achieve a more optimal bias-variance tradeoff.

\paragraph{Impact of Prompting Strategies}
Few-shot prompting yields modest boosts in primary-category performance for some models (e.g., GPT-4.1 improves from 0.796 to 0.813 accuracy), highlighting the value of in-context learning for broad tasks. However, in subcategory detection, few-shot gains are minimal or inconsistent. This suggests that for fine-grained tasks, a small set of exemplars may act as noise, anchoring the model to idiosyncratic patterns rather than clarifying the complex boundaries between subtypes. By contrast, supervised finetuning delivers substantial improvements across subcategory tasks, underscoring that direct model adaptation is the most effective strategy for specialized harmful content detection.

\subsection{Per-Class Detection Performance}
\label{sec:per_class_performance}

\begin{table*}[!ht]
\centering
\small
\begin{tabular}{l|cc|cc|cc|ccc|ccc}
\toprule
& \multicolumn{6}{c|}{\textbf{Commercial Models}} & \multicolumn{6}{c}{\textbf{Open-Source Models}} \\
\cmidrule(lr){2-7} \cmidrule(lr){8-13}
& \multicolumn{2}{c|}{\textbf{Claude-4}} & \multicolumn{2}{c|}{\textbf{Gemini-2.5}} & \multicolumn{2}{c|}{\textbf{GPT-4.1}} & \multicolumn{3}{c|}{\textbf{InternVL3-38B}} & \multicolumn{3}{c}{\textbf{Qwen2.5VL-32B}} \\
\cmidrule(lr){2-3} \cmidrule(lr){4-5} \cmidrule(lr){6-7} \cmidrule(lr){8-10} \cmidrule(lr){11-13}
\textbf{Primary Category} & ZS & FS & ZS & FS & ZS & FS & ZS & FS & FT & ZS & FS & FT \\
\midrule
Body & \cellcolor{cyan!25}\textbf{0.749} & \cellcolor{green!25}\textbf{0.757} & 0.716 & 0.697 & 0.740 & 0.739 & 0.700 & 0.733 & 0.702 & 0.586 & 0.693 & 0.689 \\
Food & \cellcolor{green!25}\textbf{0.866} & \cellcolor{cyan!25}\textbf{0.856} & 0.829 & 0.809 & 0.807 & 0.840 & 0.832 & 0.842 & 0.845 & 0.780 & 0.854 & 0.829 \\
Exercise & \cellcolor{green!25}\textbf{0.762} & 0.727 & \cellcolor{cyan!25}\textbf{0.734} & 0.694 & 0.713 & 0.728 & 0.708 & 0.725 & 0.719 & 0.655 & 0.728 & 0.720 \\
Supplements & 0.922 & \cellcolor{green!25}\textbf{0.928} & 0.892 & 0.882 & 0.897 & \cellcolor{cyan!25}\textbf{0.924} & 0.898 & \cellcolor{green!25}\textbf{0.928} & 0.890 & 0.870 & 0.882 & 0.901 \\
Masculinity & \cellcolor{cyan!25}\textbf{0.867} & \cellcolor{green!25}\textbf{0.869} & 0.828 & 0.771 & 0.832 & 0.847 & 0.793 & 0.818 & 0.549 & 0.735 & 0.844 & 0.654 \\
Irrelevant & 0.794 & 0.769 & 0.826 & 0.796 & 0.766 & 0.796 & 0.777 & 0.798 & \cellcolor{cyan!25}\textbf{0.843} & 0.827 & 0.841 & \cellcolor{green!25}\textbf{0.863} \\
\bottomrule
\end{tabular}
\caption{Per-Class F1 Scores for Primary Category Classification by Model and Training Paradigm. ZS = Zero-Shot, FS = Few-Shot, FT = Finetuned. Highest scores per category in \hlgreen{\textbf{green}}, 2nd highest in \hlcyan{\textbf{blue}}.}
\label{tab:per_class_type}
\end{table*}

Per-class performance analysis reveals systematic variation in detection difficulty that validates the importance of expert clinical knowledge (Table~\ref{tab:per_class_type} for Task 1 and Table~\ref{tab:per_class_subtype}, Appendix~\ref{sec:per_type} for Task 2).

\paragraph{Content Explicitness Hierarchy}
A clear hierarchy emerges: \textit{Supplement Abuse} achieves the highest performance across all models (up to 0.928 F1) as harmful indicators like steroid mentions (``tren'', ``stacking'') provide unambiguous lexical and visual signals. In contrast, \textit{Relationship to Body} and \textit{Relationship to Exercise} categories prove most challenging (F1 $\approx$ 0.70--0.76). These categories require distinguishing legitimate fitness content from harmful messaging along subjective boundaries, creating a dependency on expert knowledge that even large commercial models struggle to navigate without specific tuning.

\paragraph{Cultural Nuance and Data Density}
Commercial models outperform open-source alternatives on culturally nuanced categories like \textit{Relationship to Masculinity}, likely due to broader pre-training on sociological concepts. However, expert-labeled finetuning allows open-source models to yield results competitive with commercial systems in specific domains (e.g., InternVL3-38B matches Claude-Sonnet-4 on \textit{Supplements}). Notably, finetuning effectiveness correlates directly with annotation density: while these models excel on data-rich \textit{Irrelevant} content ($N=548$), they struggle significantly on sparse categories like \textit{Masculinity} ($N=164$).

\subsection{Severity Score Prediction}

\begin{table}[!ht]
\centering
\small
\begin{tabular}{l|rr}
\toprule
Model                   & \multicolumn{2}{c}{Severity} \\
\cmidrule(lr){2-3}
                        & MAE $\downarrow$     & $\rho$ $\uparrow$\\
\midrule
Claude‐Sonnet‐4         & \cellcolor{green!25}\textbf{0.679}        & 0.675           \\
GPT‐4.1                 & \cellcolor{cyan!25}\textbf{0.690}        & \cellcolor{cyan!25}\textbf{0.690}              \\
Gemini‐2.5‐Flash        & 0.693        & \cellcolor{green!25}\textbf{0.691}              \\
\cmidrule{1-3}
InternVL3‐8B            & 0.805        & 0.474              \\
InternVL3‐38B           & 0.701        & 0.607              \\
Qwen2.5‐VL‐7B           & 0.794        & 0.484              \\
Qwen2.5‐VL‐32B          & 0.688        & 0.601              \\
\bottomrule
\end{tabular}
\caption{Zero-shot Severity Prediction Metrics. MAE: Mean Absolute Error (lower is better); $\rho$: Spearman's Rank Correlation (higher is better). Highest values in \colorbox{green!25}{\textbf{green}}, 2nd highest in \colorbox{cyan!25}{\textbf{blue}}.}
\label{tab:severity}
\end{table}

To assess the ability of VLMs to predict harm severity, we conducted zero-shot prompting experiments across multiple models. Results shown in Table~\ref{tab:severity} reveal that models typically misestimate severity by approximately half a scale point, indicating reasonable but imperfect alignment with expert clinical judgment. Notably, severity prediction proves more challenging than categorical classification, with all models showing only moderate correlation with ground truth ratings. While these findings demonstrate baseline competence in severity estimation, substantial improvement is needed for reliable moderation deployment.

\subsection{Error Analysis}
\label{sec:error_analysis}

Cross-model error analysis exposes fundamental challenges in characterizing pro-bigorexia content, particularly when visual signals conflict with textual context.

\paragraph{Qualitative Case Study}
We illustrate these challenges through a representative error in an example video (\texttt{Video\_ID 030}). The ground truth label is \textit{Relationship to Masculinity} because the on-screen text explicitly ties specific body metrics (``170-190 lbs'', ``1.6 shoulder-to-waist ratio'') to female desirability (``what women find''). While Claude-Sonnet-4 (zero-shot) correctly identifies this Masculinity context, both Gemini-2.5-Flash and GPT-4.1 (zero-shot) misclassify it as \textit{Relationship to Body}. This error suggests that the misclassifying models overweight the visual modality, which features prominent shirtless posing, while underweighting the crucial textual overlay that reframes the physique display as a prerequisite for male social worth. This exemplifies the difficulty of distinguishing between pure body objectification and masculinity-driven dysmorphia without deep multimodal reasoning.

\paragraph{Systematic Misclassification}
Broader quantitative analysis mirrors these qualitative findings. Models consistently misclassify exercise-related content as \textit{Irrelevant} (0.061--0.143) or confuse \textit{Relationship to Body} with \textit{Relationship to Exercise} (0.102--0.173), reflecting annotator disagreements. Videos demonstrating workout sessions simultaneously touch on exercise, body display, dieting, lifestyle, and motivations that single-label classification cannot capture. Classification consistently confuses \textit{Hormone Therapy} and \textit{Anabolic Steroids} (0.118--0.176 bidirectional misclassification), a clinically critical distinction. Models also frequently misclassify \textit{Supplement Abuse} as \textit{Relationship to Body} (0.082--0.224), suggesting difficulty in recognizing harmful supplement messaging. This likely stems from creators using coded language (e.g., ``tren'', ``stack''), requiring domain expertise that general-purpose models lack. Additionally, models show systematic bias toward predicting \textit{Muscularity Self-objectification} in subtype classification, with excessive exercise and toxic motivation content frequently misclassified into this category (0.147--0.471). These systematic errors highlight a fundamental challenge: effective pro-bigorexia detection demands not just multimodal capabilities, but clinical and social knowledge to navigate the blurred boundaries between fitness content and promoting unhealthy behaviors.

\subsection{Ablation Study}
\label{sec:ablation}

\subsubsection{Input Features: Text vs Video}

\begin{table}[t]
\centering
\small
\resizebox{\columnwidth}{!}{%
  \begin{tabular}{l|l|ccc}
    \toprule
    \textbf{Model} & \textbf{Modality} & \textbf{P$_{\mathrm{m}}$} & \textbf{R$_{\mathrm{m}}$} & \textbf{F1$_{\mathrm{m}}$} \\
    \midrule
    Gemini‑2.5‑Flash & Audio   & 0.700 & 0.423 & 0.424 \\
                     & Audio All   &  0.669 & 0.471 & 0.478  \\
                     & Caption & 0.732 & 0.707 & 0.709 \\
                     & OCR     & 0.717 & 0.661 & 0.664 \\
                     & Text    & 0.733 & 0.719 & 0.719 \\
                     & Video   & \cellcolor{green!25}\textbf{0.786} & \cellcolor{cyan!25}\textbf{0.765} & \cellcolor{cyan!25}\textbf{0.766} \\
                     & All     & \cellcolor{cyan!25}\textbf{0.782} & \cellcolor{green!25}\textbf{0.776} & \cellcolor{green!25}\textbf{0.775} \\
    \cmidrule(lr){1-5}
    GPT‑4.1          & Audio   & 0.752 & 0.444 & 0.453 \\
                     & Audio All & 0.724 & 0.488 & 0.505 \\
                     & Caption & 0.728 & 0.672 & 0.679 \\
                     & OCR     & 0.741 & 0.663 & 0.672 \\
                     & Text    & 0.749 & 0.675 & 0.681 \\
                     & Video   & \cellcolor{cyan!25}\textbf{0.755} & \cellcolor{cyan!25}\textbf{0.697} & \cellcolor{cyan!25}\textbf{0.701} \\
                     & All     & \cellcolor{green!25}\textbf{0.808} & \cellcolor{green!25}\textbf{0.792} & \cellcolor{green!25}\textbf{0.792} \\
    \cmidrule(lr){1-5}
    InternVL3‑38B    & Audio   & 0.632 & 0.405 & 0.416 \\
                     & Audio All & 0.630  & 0.454  & 0.470 \\
                     & Caption & 0.669 & 0.590 & 0.587 \\
                     & OCR     & 0.649 & 0.614 & 0.618 \\
                     & Text    & 0.691 & 0.630 & 0.635 \\
                     & Video   & \cellcolor{cyan!25}\textbf{0.697} & \cellcolor{cyan!25}\textbf{0.655} & \cellcolor{cyan!25}\textbf{0.660} \\
                     & All     & \cellcolor{green!25}\textbf{0.797} & \cellcolor{green!25}\textbf{0.784} & \cellcolor{green!25}\textbf{0.785} \\
    \bottomrule
  \end{tabular}%
}
\caption{Ablation results for task 1 using a single modality input features (text or video). \textit{Audio} modality refers to the audio transcript of the video whose audio is classified as speech; \textit{Audio All} modality refers to the audio transcript of all videos; \textit{Caption} refers to the video description; OCR is text within images, and \textit{Text} modality refers to \textit{Caption} + \textit{OCR}. Highest values in each column are in \hlgreen{\textbf{green}}, second‑best in \hlcyan{\textbf{cyan}}. P$_{\mathrm{m}}$, R$_{\mathrm{m}}$, and F1$_{\mathrm{m}}$ refer to macro precision, macro recall, and macro F1 scores. }
\label{tab:unimodal_ablation}
\end{table}

Table~\ref{tab:unimodal_ablation} reports results of ablation study to assess the contribution of each modality to classification performance on Task 1 with zero-shot prompting. We evaluate GPT-4.1 and InternVL3-38B (image-based) and Gemini-2.5-Flash (video-native). Multimodal fusion yields the best results, with GPT-4.1 and InternVL3-38B achieving F1 scores of 0.792 and 0.785, respectively. Video alone provides the strongest unimodal performance (F1 = 0.660--0.701), highlighting the discriminative power of visual behavioral cues. In contrast, audio consistently underperforms (F1 $<$ 0.453), likely due to background music and variable audio quality. Gemini-2.5-Flash shows narrower modality gaps, suggesting native video processing reduces the need for explicit fusion. Overall, results underscore the value of combining visual and textual inputs for robust pro-bigorexia detection.

\subsubsection{Number of Frames per Video}

\begin{table}[ht]
\centering
\small
\begin{tabular}{l| r | c c c}
  \toprule
  \textbf{Model}        & \textbf{Frames \#} & \textbf{P$_{\mathrm{m}}$}    & \textbf{R$_{\mathrm{m}}$}    & \textbf{F1$_{\mathrm{m}}$}   \\
  \midrule
  GPT‑4.1               & 4                  & \hlgreen{\textbf{0.832}}            & \hlgreen{\textbf{0.829}}            & \hlgreen{\textbf{0.827}}            \\
                        & 16                 & \hlcyan{\textbf{0.817}}            & \hlcyan{\textbf{0.793}}             & \hlcyan{\textbf{0.792}}            \\
                        & 32                 & 0.739            & 0.657            & 0.685            \\
  \cmidrule(lr){1-5}
  InternVL3‑38B         & 4                  & 0.7967        & 0.7840        & 0.7849        \\
                        & 16                 & 0.7988        & 0.7891        & 0.7896        \\
                        & 32                 & 0.7922        & 0.7840        & 0.7844        \\
  \bottomrule
\end{tabular}
\caption{Frame‑count ablation for GPT‑4.1 and InternVL3‑38B. P$_{\mathrm{m}}$, R$_{\mathrm{m}}$, and F1$_{\mathrm{m}}$ are macro precision, macro recall, and macro F1. Highest values in each column are in \hlgreen{\textbf{green}}, second‑best in \hlcyan{\textbf{cyan}}.}
\label{tab:frame_ablation}
\end{table}

We examine whether providing more video content as input features to the models helps improve results. 
Our frame-count ablation (Table~\ref{tab:frame_ablation}) reveals contrasting patterns between models. Performance of GPT-4.1 shows substantial degradation with increased frames (F1: 0.827$\rightarrow$0.792$\rightarrow$0.685), while InternVL3-38B maintains stable performance across all densities (F1: 0.785-0.790). This suggests that GPT-4.1 suffers from information overload when processing dense temporal sequences, whereas InternVL3-38B effectively captures the relatively static visual elements in TikTok pro-bigorexia content. The stability across frame counts indicates that complementary text and audio modalities provide sufficient dynamic contextual information, validating our cost-efficient 4-frame approach.

\section{Conclusion}

We introduce \textsc{BigTokDetect} to bridge the gap between generic content moderation and the clinical nuances of muscle dysmorphia. While focused on this specific pathology, our work targets a broader class of multimodal mental health risks that evade traditional guardrails. Crucially, our results demonstrate that identifying these subtle behavioral harms relies less on model size and more on domain-specific adaptation; supplying high-quality, expert-annotated data for finetuning proves more effective than relying on massive generalist models. Future work may extend this via RLHF to integrate annotator disagreement as meaningful signal, preserving model utility while capturing the ambiguity inherent in clinical assessment that rigid training often discards. This functionality is particularly critical as short-form video platforms like TikTok increasingly shape youth culture and to combat the use of generative AI in health misinformation. Our framework provides reproducible methodology for the dynamic, behavior-aware moderation systems necessary to address the growing complexity of modern online harms.

\section*{Limitations}

\paragraph{Dataset and Sampling Constraints} Our dataset is limited to English-only TikTok content from 2019-2025, potentially missing cultural variations and platform-specific differences. We focused on English to ensure annotation quality and TikTok as the dominant youth platform. Keyword-based sampling may overlook subtle or emerging pro-bigorexia content that avoids our taxonomy terms, though we exhaustively developed keywords through previous literature review and expert consultation to maximize coverage. 

We did not track creator gender due to the privacy and ethical risks of visual inference. Since pro-bigorexia content typically skews male, while our ``Irrelevant'' control is likely gender-diverse, this demographic discrepancy presents a potential confounder for future work to address via privacy-preserving methods.

\paragraph{Annotation and Taxonomy Limitations} Forced reduction from multi-faceted expert annotations to single labels loses nuanced information. We simplified to single labels due to current LLM limitations in reliable multi-label classification. Our 16 annotators are predominantly female (13/16), potentially introducing gender perspective bias in male-centric bigorexia evaluation; however, this field remains understudied, and we could not find sufficient male experts specializing in bigorexia. Our taxonomy represents a living document that may miss emerging patterns, though we strived for exhaustiveness through iterative expert refinement.

\paragraph{Model Selection Limitations} While we selected current leading VLMs across both commercial and open-source categories, the rapidly evolving landscape means we may be missing other capable models. We prioritized models with proven multimodal video understanding capabilities, but acknowledge that newer or specialized architectures might offer different performance characteristics.
\paragraph{Experimental Reproducibility} Due to computational cost constraints, we conduct single runs for each experimental configuration without multiple trials or statistical significance testing. While our results establish baseline performance across models and tasks, future work should include multiple experimental runs with statistical analysis to provide more robust performance comparisons and confidence intervals for the reported metrics.

\section*{Ethics Statement}

This research addresses harmful content classification, raising important ethical considerations. To protect annotators reviewing potentially disturbing pro-bigorexia content, we recruited clinical experts, licensed psychologists, psychiatrists, and doctoral candidates specializing in eating disorders, plus computational social scientists with extensive experience in body image research. All annotators participated as voluntary research collaborators, fully informed of the project scope, with explicit rights to withdraw without penalty.

To strictly protect user privacy, we maintain strict protocols. While annotators accessed original content to ensure accurate clinical coding, we implemented a comprehensive manual de-identification protocol for the public data release. Four authors conducted a frame-by-frame review of the entire dataset. For every video, we annotated the specific on-screen coordinates of embedded usernames (\texttt{@<username>}), which can shift across frames, and redacted these regions using a white overlay box. Furthermore, we excised all platform-generated outro segments that display creator handles. This manual redaction, combined with the removal of user IDs and metadata, ensures that the stored videos and extracted frames contain no visible creator usernames or TikTok account identifiers. The full dataset will be made available only upon eligible research requests with ethics approval and signed data use agreements.

While supporting mental health research, we acknowledge risks of censorship against legitimate fitness communities. We focus on detection research rather than deployment recommendations. Our work is intended for research purposes and should not be used for clinical diagnosis without proper clinical oversight. The study protocol was reviewed and approved by the University of Southern California Institutional Review Board (IRB application number UP-23-00750).

We acknowledge the assistance of an AI language model in writing and editing portions of this manuscript. All AI-generated content was reviewed and edited by authors. Research design, data collection, analysis, and primary contributions remain entirely human work.

\section*{Acknowledgments}

We are deeply grateful to our clinical collaborators at the University of California, Los Angeles Health, including Krista Tabuenca, Artha Gillis, Colleen McCord, Brandy Saccacio, Amanda Velkova, Chessa Kabiling, Jenny Nguyen, and Danielle Crowe, for generously volunteering their time and expertise to annotate this sensitive content, without whose clinical insights this work would not have been possible

\bibliography{references}

@article{becker2004television,
  title   = {Television, Disordered Eating, and Young Women in Fiji: Negotiating Body Image and Identity during Rapid Social Change},
  author  = {Becker, Anne E.},
  journal = {Culture, Medicine and Psychiatry},
  year    = {2004},
  volume  = {28},
  number  = {4},
  pages   = {533--559},
  month   = dec,
  doi     = {10.1007/s11013-004-1067-5},
  url     = {https://doi.org/10.1007/s11013-004-1067-5},
  pmid    = {15847053}
}

@article{Pope1997,
  author  = {Pope, Harrison G. Jr. and Gruber, Anthony J. and Choi, Paulette and Olivardia, Roberto and Phillips, Katharine A.},
  title   = {Muscle Dysmorphia: An Underrecognized Form of Body Dysmorphic Disorder},
  journal = {Psychosomatics},
  year    = {1997},
  volume  = {38},
  number  = {6},
  pages   = {548--557},
  month   = nov,
  note    = {Nov--Dec},
  doi     = {10.1016/S0033-3182(97)71400-2},
  url     = {https://doi.org/10.1016/S0033-3182%2897%2971400-2}
}

@article{Cooper2020,
  author = {Cooper, Marita and Eddy, Kamryn T. and Thomas, Jennifer J. and Franko, Debra L. and Carron-Arthur, Bradley and Keshishian, Ani C. and Griffiths, Kathleen M.},
  title = {Muscle dysmorphia: A systematic and meta-analytic review of the literature to assess diagnostic validity},
  journal = {International Journal of Eating Disorders},
  volume = {53},
  number = {10},
  pages = {1583--1604},
  year = {2020},
  doi = {10.1002/eat.23349}
}

@article{Mitchison2022,
  author = {Mitchison, Deborah and Mond, Jonathan and Griffiths, Scott and Hay, Phillipa and Nagata, Jason M. and Bussey, Kay and Trompeter, Nora and Lonergan, Alexandra and Murray, Stuart B.},
  title = {Prevalence of muscle dysmorphia in adolescents: findings from the EveryBODY Study},
  journal = {Psychological Medicine},
  volume = {52},
  number = {14},
  pages = {3142--3149},
  year = {2022},
  doi = {10.1017/S0033291720005206}
}

@article{Lookingbill2023,
  author = {Lookingbill, Valerie and Mohammadi, Ehsan and Cai, Yizhou},
  title = {Assessment of Accuracy, User Engagement, and Themes of Eating Disorder Content in Social Media Short Videos},
  journal = {JAMA Network Open},
  volume = {6},
  number = {4},
  pages = {e238897},
  year = {2023},
  doi = {10.1001/jamanetworkopen.2023.8897}
}

@inproceedings{Chancellor2017,
  author = {Chancellor, Stevie and Kalantidis, Yannis and Pater, Jessica A. and De Choudhury, Munmun and Shamma, David A.},
  title = {Multimodal Classification of Moderated Online Pro-Eating Disorder Content},
  booktitle = {Proceedings of the 2017 ACM CHI Conference on Human Factors in Computing Systems},
  pages = {3213--3226},
  year = {2017},
  publisher = {ACM},
  doi = {10.1145/3025453.3025985}
}

@inproceedings{Kiela2020,
 author = {Kiela, Douwe and Firooz, Hamed and Mohan, Aravind and Goswami, Vedanuj and Singh, Amanpreet and Ringshia, Pratik and Testuggine, Davide},
 booktitle = {Advances in Neural Information Processing Systems},
 editor = {H. Larochelle and M. Ranzato and R. Hadsell and M.F. Balcan and H. Lin},
 pages = {2611--2624},
 publisher = {Curran Associates, Inc.},
 title = {The Hateful Memes Challenge: Detecting Hate Speech in Multimodal Memes},
 url = {https://proceedings.neurips.cc/paper_files/paper/2020/file/1b84c4cee2b8b3d823b30e2d604b1878-Paper.pdf},
 volume = {33},
 year = {2020}
}

@article{chancellor2020methods,
  title={Methods in predictive techniques for mental health status on social media: a critical review},
  author = {Chancellor, Stevie and Choudhury, Munmun},
  journal={NPJ Digital Medicine},
  year={2020},
  volume={3},
  number={1},
  pages={43},
  doi={10.1038/s41746-020-0233-7}
}

@misc{bickham2025edtokdataseteatingdisorder,
      title={EDTok: A Dataset for Eating Disorder Content on TikTok}, 
      author={Charles Bickham and Bryan Ramirez-Gonzalez and Minh Duc Chu and Kristina Lerman and Emilio Ferrara},
      year={2025},
      eprint={2505.02250},
      archivePrefix={arXiv},
      primaryClass={cs.SI},
      url={https://arxiv.org/abs/2505.02250}, 
}

@inproceedings{donati-etal-2023-building,
    title = "Building a Corpus on Eating Disorders from {T}ik{T}ok: Challenges and Opportunities",
    author = "Donati, Melissa  and
      Polidori, Ludovica  and
      Vernillo, Paola  and
      Gagliardi, Gloria",
    editor = "Boschetti, Federico  and
      Lebani, Gianluca E.  and
      Magnini, Bernardo  and
      Novielli, Nicole",
    booktitle = "Proceedings of the 9th Italian Conference on Computational Linguistics (CLiC-it 2023)",
    month = nov,
    year = "2023",
    address = "Venice, Italy",
    publisher = "CEUR Workshop Proceedings",
    url = "https://aclanthology.org/2023.clicit-1.66/",
    pages = "515--520",
    ISBN = "979-12-550-0084-6"
}

@article{murray2017enigma,
  title={The enigma of male eating disorders: A critical review and synthesis},
  author={Murray, Stuart B and Nagata, Jason M and Griffiths, Scott and Calzo, Jerel P and Brown, Tiffany A and Mitchison, Deborah and Blashill, Aaron J and Mond, Jonathan M},
  journal={Clinical psychology review},
  volume={57},
  pages={1--11},
  year={2017},
  publisher={Elsevier}
}

@article{ganson2025associations,
  title={Associations between muscularity-oriented social media content and muscle dysmorphia among boys and men},
  author={Ganson, Kyle T. and Testa, Alexander and Rodgers, Rachel F. and Nagata, Jason M.},
  journal={Body Image},
  volume={53},
  pages={101903},
  year={2025},
  publisher={Elsevier},
  doi={10.1016/j.bodyim.2025.101903},
  url={https://doi.org/10.1016/j.bodyim.2025.101903}
}

@article{kamkari2025manosphere,
  title={The manosphere and men's health: unpacking the links between online communities, body dysmorphia and erectile dysfunction},
  author={Kamkari, Nick Ara},
  journal={Journal of Men's Health},
  volume={21},
  number={4},
  pages={123--125},
  year={2025}
}

@misc{alayrac2022flamingovisuallanguagemodel,
      title={Flamingo: a Visual Language Model for Few-Shot Learning}, 
      author={Jean-Baptiste Alayrac and Jeff Donahue and Pauline Luc and Antoine Miech and Iain Barr and Yana Hasson and Karel Lenc and Arthur Mensch and Katie Millican and Malcolm Reynolds and Roman Ring and Eliza Rutherford and Serkan Cabi and Tengda Han and Zhitao Gong and Sina Samangooei and Marianne Monteiro and Jacob Menick and Sebastian Borgeaud and Andrew Brock and Aida Nematzadeh and Sahand Sharifzadeh and Mikolaj Binkowski and Ricardo Barreira and Oriol Vinyals and Andrew Zisserman and Karen Simonyan},
      year={2022},
      eprint={2204.14198},
      archivePrefix={arXiv},
      primaryClass={cs.CV},
      url={https://arxiv.org/abs/2204.14198}, 
}

@misc{li2023blip2bootstrappinglanguageimagepretraining,
      title={BLIP-2: Bootstrapping Language-Image Pre-training with Frozen Image Encoders and Large Language Models}, 
      author={Junnan Li and Dongxu Li and Silvio Savarese and Steven Hoi},
      year={2023},
      eprint={2301.12597},
      archivePrefix={arXiv},
      primaryClass={cs.CV},
      url={https://arxiv.org/abs/2301.12597}, 
}

@misc{OpenAI2023,
  author       = {{OpenAI}},
  title        = {GPT-4 Technical Report},
  howpublished = {Online},
  year         = {2023},
  note         = {\url{https://openai.com/research/gpt-4}}
}

@article{Cafri2008,
author = {Cafri, Guy and Olivardia, Roberto and Thompson, Joel},
year = {2008},
month = {07},
pages = {374-9},
title = {Symptom characteristics and psychiatric comorbidity among males with muscle dysmorphia},
volume = {49},
journal = {Comprehensive psychiatry},
doi = {10.1016/j.comppsych.2008.01.003}
}

@misc{mccreary2000drive,
  title={Drive for Muscularity Scale (DMS)},
  author={McCreary, Donald R. and Sasse, Doris K.},
  year={2000},
  publisher={APA PsycTests},
  note={Database record},
  doi={10.1037/t39771-000},
  url={https://doi.org/10.1037/t39771-000}
}

@article{zeeck2018muscle,
  title={Muscle Dysmorphic Disorder Inventory (MDDI): Validation of a German version with a focus on gender},
  author={Zeeck, Almut and Welter, Viola and Alatas, Hasan and Hildebrandt, Tom and Lahmann, Claas and Hartmann, Armin},
  journal={PLoS ONE},
  volume={13},
  number={11},
  pages={e0207535},
  year={2018},
  month={11},
  day={16},
  publisher={Public Library of Science},
  doi={10.1371/journal.pone.0207535},
  url={https://doi.org/10.1371/journal.pone.0207535}
}

@misc{sanchez2024feelingsbodiesemotionsdiet,
  title         = {Leveraging Machine Learning to Identify Gendered Stereotypes and Body Image Concerns on Diet and Fitness Online Forums},
  author        = {Minh Duc Chu and Cinthia S{\'a}nchez and Zihao He and Rebecca Dorn and Stuart B. Murray and Kristina Lerman},
  year          = {2024},
  eprint        = {2407.03551},
  archivePrefix = {arXiv},
  primaryClass  = {cs.SI},
  doi           = {10.48550/arXiv.2407.03551},
  url           = {https://doi.org/10.48550/arXiv.2407.03551}
}

@article{WP2023BodyDysmorphia,
  author    = {Megan Willett},
  title     = {Body Dysmorphia in Boys and Men Can Fuel Muscle Obsession},
  journal   = {The Washington Post},
  year      = {2023},
  month     = {April},
  day       = {14},
  url       = {https://www.washingtonpost.com/wellness/2023/04/14/male-body-dysmorphia-bodybuilding/}
}

@misc{TikTokResearchAPI,
  author       = {{TikTok}},
  title        = {Query Videos – TikTok Research API},
  howpublished = {Online; accessed July 2025},
  year         = {2025},
  url          = {https://developers.tiktok.com/doc/research-api-specs-query-videos}
}

@Article{corey2022,
author="Basch, Corey H
and Donelle, Lorie
and Fera, Joseph
and Jaime, Christie",
title="Deconstructing TikTok Videos on Mental Health: Cross-sectional, Descriptive Content Analysis",
journal="JMIR Form Res",
year="2022",
month="May",
day="19",
volume="6",
number="5",
pages="e38340",
issn="2561-326X",
doi="10.2196/38340",
url="https://formative.jmir.org/2022/5/e38340",
url="https://doi.org/10.2196/38340",
url="http://www.ncbi.nlm.nih.gov/pubmed/35588057"
}

@article{murray2015,
author = {Murray, Stuart and Hazery, Leila and Shen, Tori and Wooldridge, Tom and Mond, Jonathan},
year = {2015},
month = {11},
pages = {17-20},
title = {Go big or go home: A thematic content analysis of pro-muscularity websites},
volume = {16},
journal = {Body image},
doi = {10.1016/j.bodyim.2015.10.002}
}

@article{Zhu2025InternVL3,
  title   = {InternVL3: Exploring Advanced Training and Test-Time Recipes for Open-Source Multimodal Models},
  author  = {Zhu, Jinguo and Wang, Weiyun and Chen, Zhe and Liu, Zhaoyang and Ye, Shenglong and Gu, Lixin and Duan, Yuchen and Tian, Hao and Su, Weijie and Shao, Jie and Gao, Zhangwei and Cui, Erfei and Liu, Yinan and Jiang, Tan and Luo, Jiapeng and Wang, Yi and He, Yinan and Lu, Tan and Lu, Lewei and Zhu, Xizhou and Lu, Tong and Lin, Dahua and Qiao, Yu},
  journal = {arXiv preprint arXiv:2504.10479},
  year    = {2025},
  url     = {https://arxiv.org/abs/2504.10479}
}

@inproceedings{Liu2024LLaVANeXT,
  title     = {LLaVA-NeXT: Tackling Multi-image, Video, and 3D in Large Multimodal Models},
  author    = {Liu, Haotian and Li, Chunyuan and Yuheng Li and Bo Li and Yuanhan Zhang and Sheng Shen and Yong Jae Lee},
  booktitle = {arXiv preprint arXiv:2407.07895},
  year      = {2024},
  url       = {https://arxiv.org/abs/2407.07895}
}

@misc{qwen2024qwen25,
  title         = {Qwen2.5 Technical Report},
  author        = {{Qwen Team}},
  year          = {2025},
  month         = jan,
  eprint        = {2412.15115},
  archivePrefix = {arXiv},
  primaryClass  = {cs.CL},
  doi           = {10.48550/arXiv.2412.15115},
  url           = {https://doi.org/10.48550/arXiv.2412.15115},
  note          = {arXiv:2412.15115v2}
}

@misc{gimenogómez2024readingframesmultimodaldepression,
      title={Reading Between the Frames: Multi-Modal Depression Detection in Videos from Non-Verbal Cues}, 
      author={David Gimeno-Gómez and Ana-Maria Bucur and Adrian Cosma and Carlos-David Martínez-Hinarejos and Paolo Rosso},
      year={2024},
      eprint={2401.02746},
      archivePrefix={arXiv},
      primaryClass={cs.CV},
      url={https://arxiv.org/abs/2401.02746}, 
}

@article{minadeo2022tiktok,
  title={Weight-normative messaging predominates on TikTok—A qualitative content analysis},
  author={Minadeo, Maria and Pope, Lizzy},
  journal={PLOS ONE},
  volume={17},
  number={11},
  pages={e0267997},
  year={2022},
  publisher={Public Library of Science},
  doi={10.1371/journal.pone.0267997}
}

@misc{TikTokCreativeCenter,
  author       = {{TikTok Creative Center}},
  title        = {Top Hashtags Overview},
  howpublished = {\url{https://ads.tiktok.com/business/creativecenter/hashtag/pc/en}},
  note         = {Accessed January 2025}
}

@article{Cohen1960,
  author  = {Cohen, Jacob},
  title   = {A Coefficient of Agreement for Nominal Scales},
  journal = {Educational and Psychological Measurement},
  year    = {1960},
  volume  = {20},
  number  = {1},
  pages   = {37--46},
  month   = apr,
  doi     = {10.1177/001316446002000104},
  url     = {https://doi.org/10.1177/001316446002000104}
}

@misc{ma2025largelanguagemodelsmultilabel,
      title={Large Language Models Do Multi-Label Classification Differently}, 
      author={Marcus Ma and Georgios Chochlakis and Niyantha Maruthu Pandiyan and Jesse Thomason and Shrikanth Narayanan},
      year={2025},
      eprint={2505.17510},
      archivePrefix={arXiv},
      primaryClass={cs.CL},
      url={https://arxiv.org/abs/2505.17510}, 
}

@misc{hershey2017cnnarchitectureslargescaleaudio,
      title={CNN Architectures for Large-Scale Audio Classification}, 
      author={Shawn Hershey and Sourish Chaudhuri and Daniel P. W. Ellis and Jort F. Gemmeke and Aren Jansen and R. Channing Moore and Manoj Plakal and Devin Platt and Rif A. Saurous and Bryan Seybold and Malcolm Slaney and Ron J. Weiss and Kevin Wilson},
      year={2017},
      eprint={1609.09430},
      archivePrefix={arXiv},
      primaryClass={cs.SD},
      url={https://arxiv.org/abs/1609.09430}, 
}

@misc{radford2022robustspeechrecognitionlargescale,
      title={Robust Speech Recognition via Large-Scale Weak Supervision}, 
      author={Alec Radford and Jong Wook Kim and Tao Xu and Greg Brockman and Christine McLeavey and Ilya Sutskever},
      year={2022},
      eprint={2212.04356},
      archivePrefix={arXiv},
      primaryClass={eess.AS},
      url={https://arxiv.org/abs/2212.04356}, 
}

@misc{openai2025gpt41,
  author       = {{OpenAI}},
  title        = {GPT-4.1},
  howpublished = {\url{https://openai.com/index/gpt-4-1/}},
  year         = {2025},
  note         = {Accessed: 2025-07-27}
}

@misc{comanici2025gemini25pushingfrontier,
      title={Gemini 2.5: Pushing the Frontier with Advanced Reasoning, Multimodality, Long Context, and Next Generation Agentic Capabilities}, 
      author={Gheorghe Comanici and Eric Bieber and Mike Schaekermann and Ice Pasupat and Noveen Sachdeva and Inderjit Dhillon and Marcel Blistein and Ori Ram and Dan Zhang and Evan Rosen and Luke Marris and Sam Petulla and Colin Gaffney and Asaf Aharoni and Nathan Lintz and Tiago Cardal Pais and Henrik Jacobsson and Idan Szpektor and Nan-Jiang Jiang and Krishna Haridasan and Ahmed Omran and Nikunj Saunshi and Dara Bahri and Gaurav Mishra and Eric Chu and Toby Boyd and Brad Hekman and Aaron Parisi and Chaoyi Zhang and Kornraphop Kawintiranon and Tania Bedrax-Weiss and Oliver Wang and Ya Xu and Ollie Purkiss and Uri Mendlovic and Ilaï Deutel and Nam Nguyen and Adam Langley and Flip Korn and Lucia Rossazza and Alexandre Ramé and Sagar Waghmare and Helen Miller and Nathan Byrd and Ashrith Sheshan and Raia Hadsell Sangnie Bhardwaj and Pawel Janus and Tero Rissa and Dan Horgan and Sharon Silver and Ayzaan Wahid and Sergey Brin and Yves Raimond and Klemen Kloboves and Cindy Wang and Nitesh Bharadwaj Gundavarapu and Ilia Shumailov and Bo Wang and Mantas Pajarskas and Joe Heyward and Martin Nikoltchev and Maciej Kula and Hao Zhou and Zachary Garrett and Sushant Kafle and Sercan Arik and Ankita Goel and Mingyao Yang and Jiho Park and Koji Kojima and Parsa Mahmoudieh and Koray Kavukcuoglu and Grace Chen and Doug Fritz and Anton Bulyenov and Sudeshna Roy and Dimitris Paparas and Hadar Shemtov and Bo-Juen Chen and Robin Strudel and David Reitter and Aurko Roy and Andrey Vlasov and Changwan Ryu and Chas Leichner and Haichuan Yang and Zelda Mariet and Denis Vnukov and Tim Sohn and Amy Stuart and Wei Liang and Minmin Chen and Praynaa Rawlani and Christy Koh and JD Co-Reyes and Guangda Lai and Praseem Banzal and Dimitrios Vytiniotis and Jieru Mei and Mu Cai and Mohammed Badawi and Corey Fry and Ale Hartman and Daniel Zheng and Eric Jia and James Keeling and Annie Louis and Ying Chen and Efren Robles and Wei-Chih Hung and Howard Zhou and Nikita Saxena and Sonam Goenka and Olivia Ma and Zach Fisher and Mor Hazan Taege and Emily Graves and David Steiner and Yujia Li and Sarah Nguyen and Rahul Sukthankar and Joe Stanton and Ali Eslami and Gloria Shen and Berkin Akin and Alexey Guseynov and Yiqian Zhou and Jean-Baptiste Alayrac and Armand Joulin and Efrat Farkash and Ashish Thapliyal and Stephen Roller and Noam Shazeer and Todor Davchev and Terry Koo and Hannah Forbes-Pollard and Kartik Audhkhasi and Greg Farquhar and Adi Mayrav Gilady and Maggie Song and John Aslanides and Piermaria Mendolicchio and Alicia Parrish and John Blitzer and Pramod Gupta and Xiaoen Ju and Xiaochen Yang and Puranjay Datta and Andrea Tacchetti and Sanket Vaibhav Mehta and Gregory Dibb and Shubham Gupta and Federico Piccinini and Raia Hadsell and Sujee Rajayogam and Jiepu Jiang and Patrick Griffin and Patrik Sundberg and Jamie Hayes and Alexey Frolov and Tian Xie and Adam Zhang and Kingshuk Dasgupta and Uday Kalra and Lior Shani and Klaus Macherey and Tzu-Kuo Huang and Liam MacDermed and Karthik Duddu and Paulo Zacchello and Zi Yang and Jessica Lo and Kai Hui and Matej Kastelic and Derek Gasaway and Qijun Tan and Summer Yue and Pablo Barrio and John Wieting and Weel Yang and Andrew Nystrom and Solomon Demmessie and Anselm Levskaya and Fabio Viola and Chetan Tekur and Greg Billock and George Necula and Mandar Joshi and Rylan Schaeffer and Swachhand Lokhande and Christina Sorokin and Pradeep Shenoy and Mia Chen and Mark Collier and Hongji Li and Taylor Bos and Nevan Wichers and Sun Jae Lee and Angéline Pouget and Santhosh Thangaraj and Kyriakos Axiotis and Phil Crone and Rachel Sterneck and Nikolai Chinaev and Victoria Krakovna and Oleksandr Ferludin and Ian Gemp and Stephanie Winkler and Dan Goldberg and Ivan Korotkov and Kefan Xiao and Malika Mehrotra and Sandeep Mariserla and Vihari Piratla and Terry Thurk and Khiem Pham and Hongxu Ma and Alexandre Senges and Ravi Kumar and Clemens Meyer and Ellie Talius and Nuo Wang Pierse and Ballie Sandhu and Horia Toma and Kuo Lin and Swaroop Nath and Tom Stone and Dorsa Sadigh and Nikita Gupta and Arthur Guez and Avi Singh and Matt Thomas and Tom Duerig and Yuan Gong and Richard Tanburn and Lydia Lihui Zhang and Phuong Dao and Mohamed Hammad and Sirui Xie and Shruti Rijhwani and Ben Murdoch and Duhyeon Kim and Will Thompson and Heng-Tze Cheng and Daniel Sohn and Pablo Sprechmann and Qiantong Xu and Srinivas Tadepalli and Peter Young and Ye Zhang and Hansa Srinivasan and Miranda Aperghis and Aditya Ayyar and Hen Fitoussi and Ryan Burnell and David Madras and Mike Dusenberry and Xi Xiong and Tayo Oguntebi and Ben Albrecht and Jörg Bornschein and Jovana Mitrović and Mason Dimarco and Bhargav Kanagal Shamanna and Premal Shah and Eren Sezener and Shyam Upadhyay and Dave Lacey and Craig Schiff and Sebastien Baur and Sanjay Ganapathy and Eva Schnider and Mateo Wirth and Connor Schenck and Andrey Simanovsky and Yi-Xuan Tan and Philipp Fränken and Dennis Duan and Bharath Mankalale and Nikhil Dhawan and Kevin Sequeira and Zichuan Wei and Shivanker Goel and Caglar Unlu and Yukun Zhu and Haitian Sun and Ananth Balashankar and Kurt Shuster and Megh Umekar and Mahmoud Alnahlawi and Aäron van den Oord and Kelly Chen and Yuexiang Zhai and Zihang Dai and Kuang-Huei Lee and Eric Doi and Lukas Zilka and Rohith Vallu and Disha Shrivastava and Jason Lee and Hisham Husain and Honglei Zhuang and Vincent Cohen-Addad and Jarred Barber and James Atwood and Adam Sadovsky and Quentin Wellens and Steven Hand and Arunkumar Rajendran and Aybuke Turker and CJ Carey and Yuanzhong Xu and Hagen Soltau and Zefei Li and Xinying Song and Conglong Li and Iurii Kemaev and Sasha Brown and Andrea Burns and Viorica Patraucean and Piotr Stanczyk and Renga Aravamudhan and Mathieu Blondel and Hila Noga and Lorenzo Blanco and Will Song and Michael Isard and Mandar Sharma and Reid Hayes and Dalia El Badawy and Avery Lamp and Itay Laish and Olga Kozlova and Kelvin Chan and Sahil Singla and Srinivas Sunkara and Mayank Upadhyay and Chang Liu and Aijun Bai and Jarek Wilkiewicz and Martin Zlocha and Jeremiah Liu and Zhuowan Li and Haiguang Li and Omer Barak and Ganna Raboshchuk and Jiho Choi and Fangyu Liu and Erik Jue and Mohit Sharma and Andreea Marzoca and Robert Busa-Fekete and Anna Korsun and Andre Elisseeff and Zhe Shen and Sara Mc Carthy and Kay Lamerigts and Anahita Hosseini and Hanzhao Lin and Charlie Chen and Fan Yang and Kushal Chauhan and Mark Omernick and Dawei Jia and Karina Zainullina and Demis Hassabis and Danny Vainstein and Ehsan Amid and Xiang Zhou and Ronny Votel and Eszter Vértes and Xinjian Li and Zongwei Zhou and Angeliki Lazaridou and Brendan McMahan and Arjun Narayanan and Hubert Soyer and Sujoy Basu and Kayi Lee and Bryan Perozzi and Qin Cao and Leonard Berrada and Rahul Arya and Ke Chen and Katrina and Xu and Matthias Lochbrunner and Alex Hofer and Sahand Sharifzadeh and Renjie Wu and Sally Goldman and Pranjal Awasthi and Xuezhi Wang and Yan Wu and Claire Sha and Biao Zhang and Maciej Mikuła and Filippo Graziano and Siobhan Mcloughlin and Irene Giannoumis and Youhei Namiki and Chase Malik and Carey Radebaugh and Jamie Hall and Ramiro Leal-Cavazos and Jianmin Chen and Vikas Sindhwani and David Kao and David Greene and Jordan Griffith and Chris Welty and Ceslee Montgomery and Toshihiro Yoshino and Liangzhe Yuan and Noah Goodman and Assaf Hurwitz Michaely and Kevin Lee and KP Sawhney and Wei Chen and Zheng Zheng and Megan Shum and Nikolay Savinov and Etienne Pot and Alex Pak and Morteza Zadimoghaddam and Sijal Bhatnagar and Yoad Lewenberg and Blair Kutzman and Ji Liu and Lesley Katzen and Jeremy Selier and Josip Djolonga and Dmitry Lepikhin and Kelvin Xu and Jacky Liang and Jiewen Tan and Benoit Schillings and Muge Ersoy and Pete Blois and Bernd Bandemer and Abhimanyu Singh and Sergei Lebedev and Pankaj Joshi and Adam R. Brown and Evan Palmer and Shreya Pathak and Komal Jalan and Fedir Zubach and Shuba Lall and Randall Parker and Alok Gunjan and Sergey Rogulenko and Sumit Sanghai and Zhaoqi Leng and Zoltan Egyed and Shixin Li and Maria Ivanova and Kostas Andriopoulos and Jin Xie and Elan Rosenfeld and Auriel Wright and Ankur Sharma and Xinyang Geng and Yicheng Wang and Sam Kwei and Renke Pan and Yujing Zhang and Gabby Wang and Xi Liu and Chak Yeung and Elizabeth Cole and Aviv Rosenberg and Zhen Yang and Phil Chen and George Polovets and Pranav Nair and Rohun Saxena and Josh Smith and Shuo-yiin Chang and Aroma Mahendru and Svetlana Grant and Anand Iyer and Irene Cai and Jed McGiffin and Jiaming Shen and Alanna Walton and Antonious Girgis and Oliver Woodman and Rosemary Ke and Mike Kwong and Louis Rouillard and Jinmeng Rao and Zhihao Li and Yuntao Xu and Flavien Prost and Chi Zou and Ziwei Ji and Alberto Magni and Tyler Liechty and Dan A. Calian and Deepak Ramachandran and Igor Krivokon and Hui Huang and Terry Chen and Anja Hauth and Anastasija Ilić and Weijuan Xi and Hyeontaek Lim and Vlad-Doru Ion and Pooya Moradi and Metin Toksoz-Exley and Kalesha Bullard and Miltos Allamanis and Xiaomeng Yang and Sophie Wang and Zhi Hong and Anita Gergely and Cheng Li and Bhavishya Mittal and Vitaly Kovalev and Victor Ungureanu and Jane Labanowski and Jan Wassenberg and Nicolas Lacasse and Geoffrey Cideron and Petar Dević and Annie Marsden and Lynn Nguyen and Michael Fink and Yin Zhong and Tatsuya Kiyono and Desi Ivanov and Sally Ma and Max Bain and Kiran Yalasangi and Jennifer She and Anastasia Petrushkina and Mayank Lunayach and Carla Bromberg and Sarah Hodkinson and Vilobh Meshram and Daniel Vlasic and Austin Kyker and Steve Xu and Jeff Stanway and Zuguang Yang and Kai Zhao and Matthew Tung and Seth Odoom and Yasuhisa Fujii and Justin Gilmer and Eunyoung Kim and Felix Halim and Quoc Le and Bernd Bohnet and Seliem El-Sayed and Behnam Neyshabur and Malcolm Reynolds and Dean Reich and Yang Xu and Erica Moreira and Anuj Sharma and Zeyu Liu and Mohammad Javad Hosseini and Naina Raisinghani and Yi Su and Ni Lao and Daniel Formoso and Marco Gelmi and Almog Gueta and Tapomay Dey and Elena Gribovskaya and Domagoj Ćevid and Sidharth Mudgal and Garrett Bingham and Jianling Wang and Anurag Kumar and Alex Cullum and Feng Han and Konstantinos Bousmalis and Diego Cedillo and Grace Chu and Vladimir Magay and Paul Michel and Ester Hlavnova and Daniele Calandriello and Setareh Ariafar and Kaisheng Yao and Vikash Sehwag and Arpi Vezer and Agustin Dal Lago and Zhenkai Zhu and Paul Kishan Rubenstein and Allen Porter and Anirudh Baddepudi and Oriana Riva and Mihai Dorin Istin and Chih-Kuan Yeh and Zhi Li and Andrew Howard and Nilpa Jha and Jeremy Chen and Raoul de Liedekerke and Zafarali Ahmed and Mikel Rodriguez and Tanuj Bhatia and Bangju Wang and Ali Elqursh and David Klinghoffer and Peter Chen and Pushmeet Kohli and Te I and Weiyang Zhang and Zack Nado and Jilin Chen and Maxwell Chen and George Zhang and Aayush Singh and Adam Hillier and Federico Lebron and Yiqing Tao and Ting Liu and Gabriel Dulac-Arnold and Jingwei Zhang and Shashi Narayan and Buhuang Liu and Orhan Firat and Abhishek Bhowmick and Bingyuan Liu and Hao Zhang and Zizhao Zhang and Georges Rotival and Nathan Howard and Anu Sinha and Alexander Grushetsky and Benjamin Beyret and Keerthana Gopalakrishnan and James Zhao and Kyle He and Szabolcs Payrits and Zaid Nabulsi and Zhaoyi Zhang and Weijie Chen and Edward Lee and Nova Fallen and Sreenivas Gollapudi and Aurick Zhou and Filip Pavetić and Thomas Köppe and Shiyu Huang and Rama Pasumarthi and Nick Fernando and Felix Fischer and Daria Ćurko and Yang Gao and James Svensson and Austin Stone and Haroon Qureshi and Abhishek Sinha and Apoorv Kulshreshtha and Martin Matysiak and Jieming Mao and Carl Saroufim and Aleksandra Faust and Qingnan Duan and Gil Fidel and Kaan Katircioglu and Raphaël Lopez Kaufman and Dhruv Shah and Weize Kong and Abhishek Bapna and Gellért Weisz and Emma Dunleavy and Praneet Dutta and Tianqi Liu and Rahma Chaabouni and Carolina Parada and Marcus Wu and Alexandra Belias and Alessandro Bissacco and Stanislav Fort and Li Xiao and Fantine Huot and Chris Knutsen and Yochai Blau and Gang Li and Jennifer Prendki and Juliette Love and Yinlam Chow and Pichi Charoenpanit and Hidetoshi Shimokawa and Vincent Coriou and Karol Gregor and Tomas Izo and Arjun Akula and Mario Pinto and Chris Hahn and Dominik Paulus and Jiaxian Guo and Neha Sharma and Cho-Jui Hsieh and Adaeze Chukwuka and Kazuma Hashimoto and Nathalie Rauschmayr and Ling Wu and Christof Angermueller and Yulong Wang and Sebastian Gerlach and Michael Pliskin and Daniil Mirylenka and Min Ma and Lexi Baugher and Bryan Gale and Shaan Bijwadia and Nemanja Rakićević and David Wood and Jane Park and Chung-Ching Chang and Babi Seal and Chris Tar and Kacper Krasowiak and Yiwen Song and Georgi Stephanov and Gary Wang and Marcello Maggioni and Stein Xudong Lin and Felix Wu and Shachi Paul and Zixuan Jiang and Shubham Agrawal and Bilal Piot and Alex Feng and Cheolmin Kim and Tulsee Doshi and Jonathan Lai and Chuqiao and Xu and Sharad Vikram and Ciprian Chelba and Sebastian Krause and Vincent Zhuang and Jack Rae and Timo Denk and Adrian Collister and Lotte Weerts and Xianghong Luo and Yifeng Lu and Håvard Garnes and Nitish Gupta and Terry Spitz and Avinatan Hassidim and Lihao Liang and Izhak Shafran and Peter Humphreys and Kenny Vassigh and Phil Wallis and Virat Shejwalkar and Nicolas Perez-Nieves and Rachel Hornung and Melissa Tan and Beka Westberg and Andy Ly and Richard Zhang and Brian Farris and Jongbin Park and Alec Kosik and Zeynep Cankara and Andrii Maksai and Yunhan Xu and Albin Cassirer and Sergi Caelles and Abbas Abdolmaleki and Mencher Chiang and Alex Fabrikant and Shravya Shetty and Luheng He and Mai Giménez and Hadi Hashemi and Sheena Panthaplackel and Yana Kulizhskaya and Salil Deshmukh and Daniele Pighin and Robin Alazard and Disha Jindal and Seb Noury and Pradeep Kumar S and Siyang Qin and Xerxes Dotiwalla and Stephen Spencer and Mohammad Babaeizadeh and Blake JianHang Chen and Vaibhav Mehta and Jennie Lees and Andrew Leach and Penporn Koanantakool and Ilia Akolzin and Ramona Comanescu and Junwhan Ahn and Alexey Svyatkovskiy and Basil Mustafa and David D'Ambrosio and Shiva Mohan Reddy Garlapati and Pascal Lamblin and Alekh Agarwal and Shuang Song and Pier Giuseppe Sessa and Pauline Coquinot and John Maggs and Hussain Masoom and Divya Pitta and Yaqing Wang and Patrick Morris-Suzuki and Billy Porter and Johnson Jia and Jeffrey Dudek and Raghavender R and Cosmin Paduraru and Alan Ansell and Tolga Bolukbasi and Tony Lu and Ramya Ganeshan and Zi Wang and Henry Griffiths and Rodrigo Benenson and Yifan He and James Swirhun and George Papamakarios and Aditya Chawla and Kuntal Sengupta and Yan Wang and Vedrana Milutinovic and Igor Mordatch and Zhipeng Jia and Jamie Smith and Will Ng and Shitij Nigam and Matt Young and Eugen Vušak and Blake Hechtman and Sheela Goenka and Avital Zipori and Kareem Ayoub and Ashok Popat and Trilok Acharya and Luo Yu and Dawn Bloxwich and Hugo Song and Paul Roit and Haiqiong Li and Aviel Boag and Nigamaa Nayakanti and Bilva Chandra and Tianli Ding and Aahil Mehta and Cath Hope and Jiageng Zhang and Idan Heimlich Shtacher and Kartikeya Badola and Ryo Nakashima and Andrei Sozanschi and Iulia Comşa and Ante Žužul and Emily Caveness and Julian Odell and Matthew Watson and Dario de Cesare and Phillip Lippe and Derek Lockhart and Siddharth Verma and Huizhong Chen and Sean Sun and Lin Zhuo and Aditya Shah and Prakhar Gupta and Alex Muzio and Ning Niu and Amir Zait and Abhinav Singh and Meenu Gaba and Fan Ye and Prajit Ramachandran and Mohammad Saleh and Raluca Ada Popa and Ayush Dubey and Frederick Liu and Sara Javanmardi and Mark Epstein and Ross Hemsley and Richard Green and Nishant Ranka and Eden Cohen and Chuyuan Kelly Fu and Sanjay Ghemawat and Jed Borovik and James Martens and Anthony Chen and Pranav Shyam and André Susano Pinto and Ming-Hsuan Yang and Alexandru Ţifrea and David Du and Boqing Gong and Ayushi Agarwal and Seungyeon Kim and Christian Frank and Saloni Shah and Xiaodan Song and Zhiwei Deng and Ales Mikhalap and Kleopatra Chatziprimou and Timothy Chung and Toni Creswell and Susan Zhang and Yennie Jun and Carl Lebsack and Will Truong and Slavica Andačić and Itay Yona and Marco Fornoni and Rong Rong and Serge Toropov and Afzal Shama Soudagar and Andrew Audibert and Salah Zaiem and Zaheer Abbas and Andrei Rusu and Sahitya Potluri and Shitao Weng and Anastasios Kementsietsidis and Anton Tsitsulin and Daiyi Peng and Natalie Ha and Sanil Jain and Tejasi Latkar and Simeon Ivanov and Cory McLean and Anirudh GP and Rajesh Venkataraman and Canoee Liu and Dilip Krishnan and Joel D'sa and Roey Yogev and Paul Collins and Benjamin Lee and Lewis Ho and Carl Doersch and Gal Yona and Shawn Gao and Felipe Tiengo Ferreira and Adnan Ozturel and Hannah Muckenhirn and Ce Zheng and Gargi Balasubramaniam and Mudit Bansal and George van den Driessche and Sivan Eiger and Salem Haykal and Vedant Misra and Abhimanyu Goyal and Danilo Martins and Gary Leung and Jonas Valfridsson and Four Flynn and Will Bishop and Chenxi Pang and Yoni Halpern and Honglin Yu and Lawrence Moore and Yuvein and Zhu and Sridhar Thiagarajan and Yoel Drori and Zhisheng Xiao and Lucio Dery and Rolf Jagerman and Jing Lu and Eric Ge and Vaibhav Aggarwal and Arjun Khare and Vinh Tran and Oded Elyada and Ferran Alet and James Rubin and Ian Chou and David Tian and Libin Bai and Lawrence Chan and Lukasz Lew and Karolis Misiunas and Taylan Bilal and Aniket Ray and Sindhu Raghuram and Alex Castro-Ros and Viral Carpenter and CJ Zheng and Michael Kilgore and Josef Broder and Emily Xue and Praveen Kallakuri and Dheeru Dua and Nancy Yuen and Steve Chien and John Schultz and Saurabh Agrawal and Reut Tsarfaty and Jingcao Hu and Ajay Kannan and Dror Marcus and Nisarg Kothari and Baochen Sun and Ben Horn and Matko Bošnjak and Ferjad Naeem and Dean Hirsch and Lewis Chiang and Boya Fang and Jie Han and Qifei Wang and Ben Hora and Antoine He and Mario Lučić and Beer Changpinyo and Anshuman Tripathi and John Youssef and Chester Kwak and Philippe Schlattner and Cat Graves and Rémi Leblond and Wenjun Zeng and Anders Andreassen and Gabriel Rasskin and Yue Song and Eddie Cao and Junhyuk Oh and Matt Hoffman and Wojtek Skut and Yichi Zhang and Jon Stritar and Xingyu Cai and Saarthak Khanna and Kathie Wang and Shriya Sharma and Christian Reisswig and Younghoon Jun and Aman Prasad and Tatiana Sholokhova and Preeti Singh and Adi Gerzi Rosenthal and Anian Ruoss and Françoise Beaufays and Sean Kirmani and Dongkai Chen and Johan Schalkwyk and Jonathan Herzig and Been Kim and Josh Jacob and Damien Vincent and Adrian N Reyes and Ivana Balazevic and Léonard Hussenot and Jon Schneider and Parker Barnes and Luis Castro and Spandana Raj Babbula and Simon Green and Serkan Cabi and Nico Duduta and Danny Driess and Rich Galt and Noam Velan and Junjie Wang and Hongyang Jiao and Matthew Mauger and Du Phan and Miteyan Patel and Vlado Galić and Jerry Chang and Eyal Marcus and Matt Harvey and Julian Salazar and Elahe Dabir and Suraj Satishkumar Sheth and Amol Mandhane and Hanie Sedghi and Jeremiah Willcock and Amir Zandieh and Shruthi Prabhakara and Aida Amini and Antoine Miech and Victor Stone and Massimo Nicosia and Paul Niemczyk and Ying Xiao and Lucy Kim and Sławek Kwasiborski and Vikas Verma and Ada Maksutaj Oflazer and Christoph Hirnschall and Peter Sung and Lu Liu and Richard Everett and Michiel Bakker and Ágoston Weisz and Yufei Wang and Vivek Sampathkumar and Uri Shaham and Bibo Xu and Yasemin Altun and Mingqiu Wang and Takaaki Saeki and Guanjie Chen and Emanuel Taropa and Shanthal Vasanth and Sophia Austin and Lu Huang and Goran Petrovic and Qingyun Dou and Daniel Golovin and Grigory Rozhdestvenskiy and Allie Culp and Will Wu and Motoki Sano and Divya Jain and Julia Proskurnia and Sébastien Cevey and Alejandro Cruzado Ruiz and Piyush Patil and Mahdi Mirzazadeh and Eric Ni and Javier Snaider and Lijie Fan and Alexandre Fréchette and AJ Pierigiovanni and Shariq Iqbal and Kenton Lee and Claudio Fantacci and Jinwei Xing and Lisa Wang and Alex Irpan and David Raposo and Yi Luan and Zhuoyuan Chen and Harish Ganapathy and Kevin Hui and Jiazhong Nie and Isabelle Guyon and Heming Ge and Roopali Vij and Hui Zheng and Dayeong Lee and Alfonso Castaño and Khuslen Baatarsukh and Gabriel Ibagon and Alexandra Chronopoulou and Nicholas FitzGerald and Shashank Viswanadha and Safeen Huda and Rivka Moroshko and Georgi Stoyanov and Prateek Kolhar and Alain Vaucher and Ishaan Watts and Adhi Kuncoro and Henryk Michalewski and Satish Kambala and Bat-Orgil Batsaikhan and Alek Andreev and Irina Jurenka and Maigo Le and Qihang Chen and Wael Al Jishi and Sarah Chakera and Zhe Chen and Aditya Kini and Vikas Yadav and Aditya Siddhant and Ilia Labzovsky and Balaji Lakshminarayanan and Carrie Grimes Bostock and Pankil Botadra and Ankesh Anand and Colton Bishop and Sam Conway-Rahman and Mohit Agarwal and Yani Donchev and Achintya Singhal and Félix de Chaumont Quitry and Natalia Ponomareva and Nishant Agrawal and Bin Ni and Kalpesh Krishna and Masha Samsikova and John Karro and Yilun Du and Tamara von Glehn and Caden Lu and Christopher A. Choquette-Choo and Zhen Qin and Tingnan Zhang and Sicheng Li and Divya Tyam and Swaroop Mishra and Wing Lowe and Colin Ji and Weiyi Wang and Manaal Faruqui and Ambrose Slone and Valentin Dalibard and Arunachalam Narayanaswamy and John Lambert and Pierre-Antoine Manzagol and Dan Karliner and Andrew Bolt and Ivan Lobov and Aditya Kusupati and Chang Ye and Xuan Yang and Heiga Zen and Nelson George and Mukul Bhutani and Olivier Lacombe and Robert Riachi and Gagan Bansal and Rachel Soh and Yue Gao and Yang Yu and Adams Yu and Emily Nottage and Tania Rojas-Esponda and James Noraky and Manish Gupta and Ragha Kotikalapudi and Jichuan Chang and Sanja Deur and Dan Graur and Alex Mossin and Erin Farnese and Ricardo Figueira and Alexandre Moufarek and Austin Huang and Patrik Zochbauer and Ben Ingram and Tongzhou Chen and Zelin Wu and Adrià Puigdomènech and Leland Rechis and Da Yu and Sri Gayatri Sundara Padmanabhan and Rui Zhu and Chu-ling Ko and Andrea Banino and Samira Daruki and Aarush Selvan and Dhruva Bhaswar and Daniel Hernandez Diaz and Chen Su and Salvatore Scellato and Jennifer Brennan and Woohyun Han and Grace Chung and Priyanka Agrawal and Urvashi Khandelwal and Khe Chai Sim and Morgane Lustman and Sam Ritter and Kelvin Guu and Jiawei Xia and Prateek Jain and Emma Wang and Tyrone Hill and Mirko Rossini and Marija Kostelac and Tautvydas Misiunas and Amit Sabne and Kyuyeun Kim and Ahmet Iscen and Congchao Wang and José Leal and Ashwin Sreevatsa and Utku Evci and Manfred Warmuth and Saket Joshi and Daniel Suo and James Lottes and Garrett Honke and Brendan Jou and Stefani Karp and Jieru Hu and Himanshu Sahni and Adrien Ali Taïga and William Kong and Samrat Ghosh and Renshen Wang and Jay Pavagadhi and Natalie Axelsson and Nikolai Grigorev and Patrick Siegler and Rebecca Lin and Guohui Wang and Emilio Parisotto and Sharath Maddineni and Krishan Subudhi and Eyal Ben-David and Elena Pochernina and Orgad Keller and Thi Avrahami and Zhe Yuan and Pulkit Mehta and Jialu Liu and Sherry Yang and Wendy Kan and Katherine Lee and Tom Funkhouser and Derek Cheng and Hongzhi Shi and Archit Sharma and Joe Kelley and Matan Eyal and Yury Malkov and Corentin Tallec and Yuval Bahat and Shen Yan and Xintian and Wu and David Lindner and Chengda Wu and Avi Caciularu and Xiyang Luo and Rodolphe Jenatton and Tim Zaman and Yingying Bi and Ilya Kornakov and Ganesh Mallya and Daisuke Ikeda and Itay Karo and Anima Singh and Colin Evans and Praneeth Netrapalli and Vincent Nallatamby and Isaac Tian and Yannis Assael and Vikas Raunak and Victor Carbune and Ioana Bica and Lior Madmoni and Dee Cattle and Snchit Grover and Krishna Somandepalli and Sid Lall and Amelio Vázquez-Reina and Riccardo Patana and Jiaqi Mu and Pranav Talluri and Maggie Tran and Rajeev Aggarwal and RJ Skerry-Ryan and Jun Xu and Mike Burrows and Xiaoyue Pan and Edouard Yvinec and Di Lu and Zhiying Zhang and Duc Dung Nguyen and Hairong Mu and Gabriel Barcik and Helen Ran and Lauren Beltrone and Krzysztof Choromanski and Dia Kharrat and Samuel Albanie and Sean Purser-haskell and David Bieber and Carrie Zhang and Jing Wang and Tom Hudson and Zhiyuan Zhang and Han Fu and Johannes Mauerer and Mohammad Hossein Bateni and AJ Maschinot and Bing Wang and Muye Zhu and Arjun Pillai and Tobias Weyand and Shuang Liu and Oscar Akerlund and Fred Bertsch and Vittal Premachandran and Alicia Jin and Vincent Roulet and Peter de Boursac and Shubham Mittal and Ndaba Ndebele and Georgi Karadzhov and Sahra Ghalebikesabi and Ricky Liang and Allen Wu and Yale Cong and Nimesh Ghelani and Sumeet Singh and Bahar Fatemi and Warren and Chen and Charles Kwong and Alexey Kolganov and Steve Li and Richard Song and Chenkai Kuang and Sobhan Miryoosefi and Dale Webster and James Wendt and Arkadiusz Socala and Guolong Su and Artur Mendonça and Abhinav Gupta and Xiaowei Li and Tomy Tsai and Qiong and Hu and Kai Kang and Angie Chen and Sertan Girgin and Yongqin Xian and Andrew Lee and Nolan Ramsden and Leslie Baker and Madeleine Clare Elish and Varvara Krayvanova and Rishabh Joshi and Jiri Simsa and Yao-Yuan Yang and Piotr Ambroszczyk and Dipankar Ghosh and Arjun Kar and Yuan Shangguan and Yumeya Yamamori and Yaroslav Akulov and Andy Brock and Haotian Tang and Siddharth Vashishtha and Rich Munoz and Andreas Steiner and Kalyan Andra and Daniel Eppens and Qixuan Feng and Hayato Kobayashi and Sasha Goldshtein and Mona El Mahdy and Xin Wang and Jilei and Wang and Richard Killam and Tom Kwiatkowski and Kavya Kopparapu and Serena Zhan and Chao Jia and Alexei Bendebury and Sheryl Luo and Adrià Recasens and Timothy Knight and Jing Chen and Mohak Patel and YaGuang Li and Ben Withbroe and Dean Weesner and Kush Bhatia and Jie Ren and Danielle Eisenbud and Ebrahim Songhori and Yanhua Sun and Travis Choma and Tasos Kementsietsidis and Lucas Manning and Brian Roark and Wael Farhan and Jie Feng and Susheel Tatineni and James Cobon-Kerr and Yunjie Li and Lisa Anne Hendricks and Isaac Noble and Chris Breaux and Nate Kushman and Liqian Peng and Fuzhao Xue and Taylor Tobin and Jamie Rogers and Josh Lipschultz and Chris Alberti and Alexey Vlaskin and Mostafa Dehghani and Roshan Sharma and Tris Warkentin and Chen-Yu Lee and Benigno Uria and Da-Cheng Juan and Angad Chandorkar and Hila Sheftel and Ruibo Liu and Elnaz Davoodi and Borja De Balle Pigem and Kedar Dhamdhere and David Ross and Jonathan Hoech and Mahdis Mahdieh and Li Liu and Qiujia Li and Liam McCafferty and Chenxi Liu and Markus Mircea and Yunting Song and Omkar Savant and Alaa Saade and Colin Cherry and Vincent Hellendoorn and Siddharth Goyal and Paul Pucciarelli and David Vilar Torres and Zohar Yahav and Hyo Lee and Lars Lowe Sjoesund and Christo Kirov and Bo Chang and Deepanway Ghoshal and Lu Li and Gilles Baechler and Sébastien Pereira and Tara Sainath and Anudhyan Boral and Dominik Grewe and Afief Halumi and Nguyet Minh Phu and Tianxiao Shen and Marco Tulio Ribeiro and Dhriti Varma and Alex Kaskasoli and Vlad Feinberg and Navneet Potti and Jarrod Kahn and Matheus Wisniewski and Shakir Mohamed and Arnar Mar Hrafnkelsson and Bobak Shahriari and Jean-Baptiste Lespiau and Lisa Patel and Legg Yeung and Tom Paine and Lantao Mei and Alex Ramirez and Rakesh Shivanna and Li Zhong and Josh Woodward and Guilherme Tubone and Samira Khan and Heng Chen and Elizabeth Nielsen and Catalin Ionescu and Utsav Prabhu and Mingcen Gao and Qingze Wang and Sean Augenstein and Neesha Subramaniam and Jason Chang and Fotis Iliopoulos and Jiaming Luo and Myriam Khan and Weicheng Kuo and Denis Teplyashin and Florence Perot and Logan Kilpatrick and Amir Globerson and Hongkun Yu and Anfal Siddiqui and Nick Sukhanov and Arun Kandoor and Umang Gupta and Marco Andreetto and Moran Ambar and Donnie Kim and Paweł Wesołowski and Sarah Perrin and Ben Limonchik and Wei Fan and Jim Stephan and Ian Stewart-Binks and Ryan Kappedal and Tong He and Sarah Cogan and Romina Datta and Tong Zhou and Jiayu Ye and Leandro Kieliger and Ana Ramalho and Kyle Kastner and Fabian Mentzer and Wei-Jen Ko and Arun Suggala and Tianhao Zhou and Shiraz Butt and Hana Strejček and Lior Belenki and Subhashini Venugopalan and Mingyang Ling and Evgenii Eltyshev and Yunxiao Deng and Geza Kovacs and Mukund Raghavachari and Hanjun Dai and Tal Schuster and Steven Schwarcz and Richard Nguyen and Arthur Nguyen and Gavin Buttimore and Shrestha Basu Mallick and Sudeep Gandhe and Seth Benjamin and Michal Jastrzebski and Le Yan and Sugato Basu and Chris Apps and Isabel Edkins and James Allingham and Immanuel Odisho and Tomas Kocisky and Jewel Zhao and Linting Xue and Apoorv Reddy and Chrysovalantis Anastasiou and Aviel Atias and Sam Redmond and Kieran Milan and Nicolas Heess and Herman Schmit and Allan Dafoe and Daniel Andor and Tynan Gangwani and Anca Dragan and Sheng Zhang and Ashyana Kachra and Gang Wu and Siyang Xue and Kevin Aydin and Siqi Liu and Yuxiang Zhou and Mahan Malihi and Austin Wu and Siddharth Gopal and Candice Schumann and Peter Stys and Alek Wang and Mirek Olšák and Dangyi Liu and Christian Schallhart and Yiran Mao and Demetra Brady and Hao Xu and Tomas Mery and Chawin Sitawarin and Siva Velusamy and Tom Cobley and Alex Zhai and Christian Walder and Nitzan Katz and Ganesh Jawahar and Chinmay Kulkarni and Antoine Yang and Adam Paszke and Yinan Wang and Bogdan Damoc and Zalán Borsos and Ray Smith and Jinning Li and Mansi Gupta and Andrei Kapishnikov and Sushant Prakash and Florian Luisier and Rishabh Agarwal and Will Grathwohl and Kuangyuan Chen and Kehang Han and Nikhil Mehta and Andrew Over and Shekoofeh Azizi and Lei Meng and Niccolò Dal Santo and Kelvin Zheng and Jane Shapiro and Igor Petrovski and Jeffrey Hui and Amin Ghafouri and Jasper Snoek and James Qin and Mandy Jordan and Caitlin Sikora and Jonathan Malmaud and Yuheng Kuang and Aga Świetlik and Ruoxin Sang and Chongyang Shi and Leon Li and Andrew Rosenberg and Shubin Zhao and Andy Crawford and Jan-Thorsten Peter and Yun Lei and Xavier Garcia and Long Le and Todd Wang and Julien Amelot and Dave Orr and Praneeth Kacham and Dana Alon and Gladys Tyen and Abhinav Arora and James Lyon and Alex Kurakin and Mimi Ly and Theo Guidroz and Zhipeng Yan and Rina Panigrahy and Pingmei Xu and Thais Kagohara and Yong Cheng and Eric Noland and Jinhyuk Lee and Jonathan Lee and Cathy Yip and Maria Wang and Efrat Nehoran and Alexander Bykovsky and Zhihao Shan and Ankit Bhagatwala and Chaochao Yan and Jie Tan and Guillermo Garrido and Dan Ethier and Nate Hurley and Grace Vesom and Xu Chen and Siyuan Qiao and Abhishek Nayyar and Julian Walker and Paramjit Sandhu and Mihaela Rosca and Danny Swisher and Mikhail Dektiarev and Josh Dillon and George-Cristian Muraru and Manuel Tragut and Artiom Myaskovsky and David Reid and Marko Velic and Owen Xiao and Jasmine George and Mark Brand and Jing Li and Wenhao Yu and Shane Gu and Xiang Deng and François-Xavier Aubet and Soheil Hassas Yeganeh and Fred Alcober and Celine Smith and Trevor Cohn and Kay McKinney and Michael Tschannen and Ramesh Sampath and Gowoon Cheon and Liangchen Luo and Luyang Liu and Jordi Orbay and Hui Peng and Gabriela Botea and Xiaofan Zhang and Charles Yoon and Cesar Magalhaes and Paweł Stradomski and Ian Mackinnon and Steven Hemingray and Kumaran Venkatesan and Rhys May and Jaeyoun Kim and Alex Druinsky and Jingchen Ye and Zheng Xu and Terry Huang and Jad Al Abdallah and Adil Dostmohamed and Rachana Fellinger and Tsendsuren Munkhdalai and Akanksha Maurya and Peter Garst and Yin Zhang and Maxim Krikun and Simon Bucher and Aditya Srikanth Veerubhotla and Yaxin Liu and Sheng Li and Nishesh Gupta and Jakub Adamek and Hanwen Chen and Bernett Orlando and Aleksandr Zaks and Joost van Amersfoort and Josh Camp and Hui Wan and HyunJeong Choe and Zhichun Wu and Kate Olszewska and Weiren Yu and Archita Vadali and Martin Scholz and Daniel De Freitas and Jason Lin and Amy Hua and Xin Liu and Frank Ding and Yichao Zhou and Boone Severson and Katerina Tsihlas and Samuel Yang and Tammo Spalink and Varun Yerram and Helena Pankov and Rory Blevins and Ben Vargas and Sarthak Jauhari and Matt Miecnikowski and Ming Zhang and Sandeep Kumar and Clement Farabet and Charline Le Lan and Sebastian Flennerhag and Yonatan Bitton and Ada Ma and Arthur Bražinskas and Eli Collins and Niharika Ahuja and Sneha Kudugunta and Anna Bortsova and Minh Giang and Wanzheng Zhu and Ed Chi and Scott Lundberg and Alexey Stern and Subha Puttagunta and Jing Xiong and Xiao Wu and Yash Pande and Amit Jhindal and Daniel Murphy and Jon Clark and Marc Brockschmidt and Maxine Deines and Kevin R. McKee and Dan Bahir and Jiajun Shen and Minh Truong and Daniel McDuff and Andrea Gesmundo and Edouard Rosseel and Bowen Liang and Ken Caluwaerts and Jessica Hamrick and Joseph Kready and Mary Cassin and Rishikesh Ingale and Li Lao and Scott Pollom and Yifan Ding and Wei He and Lizzetth Bellot and Joana Iljazi and Ramya Sree Boppana and Shan Han and Tara Thompson and Amr Khalifa and Anna Bulanova and Blagoj Mitrevski and Bo Pang and Emma Cooney and Tian Shi and Rey Coaguila and Tamar Yakar and Marc'aurelio Ranzato and Nikola Momchev and Chris Rawles and Zachary Charles and Young Maeng and Yuan Zhang and Rishabh Bansal and Xiaokai Zhao and Brian Albert and Yuan Yuan and Sudheendra Vijayanarasimhan and Roy Hirsch and Vinay Ramasesh and Kiran Vodrahalli and Xingyu Wang and Arushi Gupta and DJ Strouse and Jianmo Ni and Roma Patel and Gabe Taubman and Zhouyuan Huo and Dero Gharibian and Marianne Monteiro and Hoi Lam and Shobha Vasudevan and Aditi Chaudhary and Isabela Albuquerque and Kilol Gupta and Sebastian Riedel and Chaitra Hegde and Avraham Ruderman and András György and Marcus Wainwright and Ashwin Chaugule and Burcu Karagol Ayan and Tomer Levinboim and Sam Shleifer and Yogesh Kalley and Vahab Mirrokni and Abhishek Rao and Prabakar Radhakrishnan and Jay Hartford and Jialin Wu and Zhenhai Zhu and Francesco Bertolini and Hao Xiong and Nicolas Serrano and Hamish Tomlinson and Myle Ott and Yifan Chang and Mark Graham and Jian Li and Marco Liang and Xiangzhu Long and Sebastian Borgeaud and Yanif Ahmad and Alex Grills and Diana Mincu and Martin Izzard and Yuan Liu and Jinyu Xie and Louis O'Bryan and Sameera Ponda and Simon Tong and Michelle Liu and Dan Malkin and Khalid Salama and Yuankai Chen and Rohan Anil and Anand Rao and Rigel Swavely and Misha Bilenko and Nina Anderson and Tat Tan and Jing Xie and Xing Wu and Lijun Yu and Oriol Vinyals and Andrey Ryabtsev and Rumen Dangovski and Kate Baumli and Daniel Keysers and Christian Wright and Zoe Ashwood and Betty Chan and Artem Shtefan and Yaohui Guo and Ankur Bapna and Radu Soricut and Steven Pecht and Sabela Ramos and Rui Wang and Jiahao Cai and Trieu Trinh and Paul Barham and Linda Friso and Eli Stickgold and Xiangzhuo Ding and Siamak Shakeri and Diego Ardila and Eleftheria Briakou and Phil Culliton and Adam Raveret and Jingyu Cui and David Saxton and Subhrajit Roy and Javad Azizi and Pengcheng Yin and Lucia Loher and Andrew Bunner and Min Choi and Faruk Ahmed and Eric Li and Yin Li and Shengyang Dai and Michael Elabd and Sriram Ganapathy and Shivani Agrawal and Yiqing Hua and Paige Kunkle and Sujeevan Rajayogam and Arun Ahuja and Arthur Conmy and Alex Vasiloff and Parker Beak and Christopher Yew and Jayaram Mudigonda and Bartek Wydrowski and Jon Blanton and Zhengdong Wang and Yann Dauphin and Zhuo Xu and Martin Polacek and Xi Chen and Hexiang Hu and Pauline Sho and Markus Kunesch and Mehdi Hafezi Manshadi and Eliza Rutherford and Bo Li and Sissie Hsiao and Iain Barr and Alex Tudor and Matija Kecman and Arsha Nagrani and Vladimir Pchelin and Martin Sundermeyer and Aishwarya P S and Abhijit Karmarkar and Yi Gao and Grishma Chole and Olivier Bachem and Isabel Gao and Arturo BC and Matt Dibb and Mauro Verzetti and Felix Hernandez-Campos and Yana Lunts and Matthew Johnson and Julia Di Trapani and Raphael Koster and Idan Brusilovsky and Binbin Xiong and Megha Mohabey and Han Ke and Joe Zou and Tea Sabolić and Víctor Campos and John Palowitch and Alex Morris and Linhai Qiu and Pranavaraj Ponnuramu and Fangtao Li and Vivek Sharma and Kiranbir Sodhia and Kaan Tekelioglu and Aleksandr Chuklin and Madhavi Yenugula and Erika Gemzer and Theofilos Strinopoulos and Sam El-Husseini and Huiyu Wang and Yan Zhong and Edouard Leurent and Paul Natsev and Weijun Wang and Dre Mahaarachchi and Tao Zhu and Songyou Peng and Sami Alabed and Cheng-Chun Lee and Anthony Brohan and Arthur Szlam and GS Oh and Anton Kovsharov and Jenny Lee and Renee Wong and Megan Barnes and Gregory Thornton and Felix Gimeno and Omer Levy and Martin Sevenich and Melvin Johnson and Jonathan Mallinson and Robert Dadashi and Ziyue Wang and Qingchun Ren and Preethi Lahoti and Arka Dhar and Josh Feldman and Dan Zheng and Thatcher Ulrich and Liviu Panait and Michiel Blokzijl and Cip Baetu and Josip Matak and Jitendra Harlalka and Maulik Shah and Tal Marian and Daniel von Dincklage and Cosmo Du and Ruy Ley-Wild and Bethanie Brownfield and Max Schumacher and Yury Stuken and Shadi Noghabi and Sonal Gupta and Xiaoqi Ren and Eric Malmi and Felix Weissenberger and Blanca Huergo and Maria Bauza and Thomas Lampe and Arthur Douillard and Mojtaba Seyedhosseini and Roy Frostig and Zoubin Ghahramani and Kelvin Nguyen and Kashyap Krishnakumar and Chengxi Ye and Rahul Gupta and Alireza Nazari and Robert Geirhos and Pete Shaw and Ahmed Eleryan and Dima Damen and Jennimaria Palomaki and Ted Xiao and Qiyin Wu and Quan Yuan and Phoenix Meadowlark and Matthew Bilotti and Raymond Lin and Mukund Sridhar and Yannick Schroecker and Da-Woon Chung and Jincheng Luo and Trevor Strohman and Tianlin Liu and Anne Zheng and Jesse Emond and Wei Wang and Andrew Lampinen and Toshiyuki Fukuzawa and Folawiyo Campbell-Ajala and Monica Roy and James Lee-Thorp and Lily Wang and Iftekhar Naim and Tony and Nguy\~ên and Guy Bensky and Aditya Gupta and Dominika Rogozińska and Justin Fu and Thanumalayan Sankaranarayana Pillai and Petar Veličković and Shahar Drath and Philipp Neubeck and Vaibhav Tulsyan and Arseniy Klimovskiy and Don Metzler and Sage Stevens and Angel Yeh and Junwei Yuan and Tianhe Yu and Kelvin Zhang and Alec Go and Vincent Tsang and Ying Xu and Andy Wan and Isaac Galatzer-Levy and Sam Sobell and Abodunrinwa Toki and Elizabeth Salesky and Wenlei Zhou and Diego Antognini and Sholto Douglas and Shimu Wu and Adam Lelkes and Frank Kim and Paul Cavallaro and Ana Salazar and Yuchi Liu and James Besley and Tiziana Refice and Yiling Jia and Zhang Li and Michal Sokolik and Arvind Kannan and Jon Simon and Jo Chick and Avia Aharon and Meet Gandhi and Mayank Daswani and Keyvan Amiri and Vighnesh Birodkar and Abe Ittycheriah and Peter Grabowski and Oscar Chang and Charles Sutton and Zhixin and Lai and Umesh Telang and Susie Sargsyan and Tao Jiang and Raphael Hoffmann and Nicole Brichtova and Matteo Hessel and Jonathan Halcrow and Sammy Jerome and Geoff Brown and Alex Tomala and Elena Buchatskaya and Dian Yu and Sachit Menon and Pol Moreno and Yuguo Liao and Vicky Zayats and Luming Tang and SQ Mah and Ashish Shenoy and Alex Siegman and Majid Hadian and Okwan Kwon and Tao Tu and Nima Khajehnouri and Ryan Foley and Parisa Haghani and Zhongru Wu and Vaishakh Keshava and Khyatti Gupta and Tony Bruguier and Rui Yao and Danny Karmon and Luisa Zintgraf and Zhicheng Wang and Enrique Piqueras and Junehyuk Jung and Jenny Brennan and Diego Machado and Marissa Giustina and MH Tessler and Kamyu Lee and Qiao Zhang and Joss Moore and Kaspar Daugaard and Alexander Frömmgen and Jennifer Beattie and Fred Zhang and Daniel Kasenberg and Ty Geri and Danfeng Qin and Gaurav Singh Tomar and Tom Ouyang and Tianli Yu and Luowei Zhou and Rajiv Mathews and Andy Davis and Yaoyiran Li and Jai Gupta and Damion Yates and Linda Deng and Elizabeth Kemp and Ga-Young Joung and Sergei Vassilvitskii and Mandy Guo and Pallavi LV and Dave Dopson and Sami Lachgar and Lara McConnaughey and Himadri Choudhury and Dragos Dena and Aaron Cohen and Joshua Ainslie and Sergey Levi and Parthasarathy Gopavarapu and Polina Zablotskaia and Hugo Vallet and Sanaz Bahargam and Xiaodan Tang and Nenad Tomasev and Ethan Dyer and Daniel Balle and Hongrae Lee and William Bono and Jorge Gonzalez Mendez and Vadim Zubov and Shentao Yang and Ivor Rendulic and Yanyan Zheng and Andrew Hogue and Golan Pundak and Ralph Leith and Avishkar Bhoopchand and Michael Han and Mislav Žanić and Tom Schaul and Manolis Delakis and Tejas Iyer and Guanyu Wang and Harman Singh and Abdelrahman Abdelhamed and Tara Thomas and Siddhartha Brahma and Hilal Dib and Naveen Kumar and Wenxuan Zhou and Liang Bai and Pushkar Mishra and Jiao Sun and Valentin Anklin and Roykrong Sukkerd and Lauren Agubuzu and Anton Briukhov and Anmol Gulati and Maximilian Sieb and Fabio Pardo and Sara Nasso and Junquan Chen and Kexin Zhu and Tiberiu Sosea and Alex Goldin and Keith Rush and Spurthi Amba Hombaiah and Andreas Noever and Allan Zhou and Sam Haves and Mary Phuong and Jake Ades and Yi-ting Chen and Lin Yang and Joseph Pagadora and Stan Bileschi and Victor Cotruta and Rachel Saputro and Arijit Pramanik and Sean Ammirati and Dan Garrette and Kevin Villela and Tim Blyth and Canfer Akbulut and Neha Jha and Alban Rrustemi and Arissa Wongpanich and Chirag Nagpal and Yonghui Wu and Morgane Rivière and Sergey Kishchenko and Pranesh Srinivasan and Alice Chen and Animesh Sinha and Trang Pham and Bill Jia and Tom Hennigan and Anton Bakalov and Nithya Attaluri and Drew Garmon and Daniel Rodriguez and Dawid Wegner and Wenhao Jia and Evan Senter and Noah Fiedel and Denis Petek and Yuchuan Liu and Cassidy Hardin and Harshal Tushar Lehri and Joao Carreira and Sara Smoot and Marcel Prasetya and Nami Akazawa and Anca Stefanoiu and Chia-Hua Ho and Anelia Angelova and Kate Lin and Min Kim and Charles Chen and Marcin Sieniek and Alice Li and Tongfei Guo and Sorin Baltateanu and Pouya Tafti and Michael Wunder and Nadav Olmert and Divyansh Shukla and Jingwei Shen and Neel Kovelamudi and Balaji Venkatraman and Seth Neel and Romal Thoppilan and Jerome Connor and Frederik Benzing and Axel Stjerngren and Golnaz Ghiasi and Alex Polozov and Joshua Howland and Theophane Weber and Justin Chiu and Ganesh Poomal Girirajan and Andreas Terzis and Pidong Wang and Fangda Li and Yoav Ben Shalom and Dinesh Tewari and Matthew Denton and Roee Aharoni and Norbert Kalb and Heri Zhao and Junlin Zhang and Angelos Filos and Matthew Rahtz and Lalit Jain and Connie Fan and Vitor Rodrigues and Ruth Wang and Richard Shin and Jacob Austin and Roman Ring and Mariella Sanchez-Vargas and Mehadi Hassen and Ido Kessler and Uri Alon and Gufeng Zhang and Wenhu Chen and Yenai Ma and Xiance Si and Le Hou and Azalia Mirhoseini and Marc Wilson and Geoff Bacon and Becca Roelofs and Lei Shu and Gautam Vasudevan and Jonas Adler and Artur Dwornik and Tayfun Terzi and Matt Lawlor and Harry Askham and Mike Bernico and Xuanyi Dong and Chris Hidey and Kevin Kilgour and Gaël Liu and Surya Bhupatiraju and Luke Leonhard and Siqi Zuo and Partha Talukdar and Qing Wei and Aliaksei Severyn and Vít Listík and Jong Lee and Aditya Tripathi and SK Park and Yossi Matias and Hao Liu and Alex Ruiz and Rajesh Jayaram and Jackson Tolins and Pierre Marcenac and Yiming Wang and Bryan Seybold and Henry Prior and Deepak Sharma and Jack Weber and Mikhail Sirotenko and Yunhsuan Sung and Dayou Du and Ellie Pavlick and Stefan Zinke and Markus Freitag and Max Dylla and Montse Gonzalez Arenas and Natan Potikha and Omer Goldman and Connie Tao and Rachita Chhaparia and Maria Voitovich and Pawan Dogra and Andrija Ražnatović and Zak Tsai and Chong You and Oleaser Johnson and George Tucker and Chenjie Gu and Jae Yoo and Maryam Majzoubi and Valentin Gabeur and Bahram Raad and Rocky Rhodes and Kashyap Kolipaka and Heidi Howard and Geta Sampemane and Benny Li and Chulayuth Asawaroengchai and Duy Nguyen and Chiyuan Zhang and Timothee Cour and Xinxin Yu and Zhao Fu and Joe Jiang and Po-Sen Huang and Gabriela Surita and Iñaki Iturrate and Yael Karov and Michael Collins and Martin Baeuml and Fabian Fuchs and Shilpa Shetty and Swaroop Ramaswamy and Sayna Ebrahimi and Qiuchen Guo and Jeremy Shar and Gabe Barth-Maron and Sravanti Addepalli and Bryan Richter and Chin-Yi Cheng and Eugénie Rives and Fei Zheng and Johannes Griesser and Nishanth Dikkala and Yoel Zeldes and Ilkin Safarli and Dipanjan Das and Himanshu Srivastava and Sadh MNM Khan and Xin Li and Aditya Pandey and Larisa Markeeva and Dan Belov and Qiqi Yan and Mikołaj Rybiński and Tao Chen and Megha Nawhal and Michael Quinn and Vineetha Govindaraj and Sarah York and Reed Roberts and Roopal Garg and Namrata Godbole and Jake Abernethy and Anil Das and Lam Nguyen Thiet and Jonathan Tompson and John Nham and Neera Vats and Ben Caine and Wesley Helmholz and Francesco Pongetti and Yeongil Ko and James An and Clara Huiyi Hu and Yu-Cheng Ling and Julia Pawar and Robert Leland and Keisuke Kinoshita and Waleed Khawaja and Marco Selvi and Eugene Ie and Danila Sinopalnikov and Lev Proleev and Nilesh Tripuraneni and Michele Bevilacqua and Seungji Lee and Clayton Sanford and Dan Suh and Dustin Tran and Jeff Dean and Simon Baumgartner and Jens Heitkaemper and Sagar Gubbi and Kristina Toutanova and Yichong Xu and Chandu Thekkath and Keran Rong and Palak Jain and Annie Xie and Yan Virin and Yang Li and Lubo Litchev and Richard Powell and Tarun Bharti and Adam Kraft and Nan Hua and Marissa Ikonomidis and Ayal Hitron and Sanjiv Kumar and Loic Matthey and Sophie Bridgers and Lauren Lax and Ishaan Malhi and Ondrej Skopek and Ashish Gupta and Jiawei Cao and Mitchelle Rasquinha and Siim Põder and Wojciech Stokowiec and Nicholas Roth and Guowang Li and Michaël Sander and Joshua Kessinger and Vihan Jain and Edward Loper and Wonpyo Park and Michal Yarom and Liqun Cheng and Guru Guruganesh and Kanishka Rao and Yan Li and Catarina Barros and Mikhail Sushkov and Chun-Sung Ferng and Rohin Shah and Ophir Aharoni and Ravin Kumar and Tim McConnell and Peiran Li and Chen Wang and Fernando Pereira and Craig Swanson and Fayaz Jamil and Yan Xiong and Anitha Vijayakumar and Prakash Shroff and Kedar Soparkar and Jindong Gu and Livio Baldini Soares and Eric Wang and Kushal Majmundar and Aurora Wei and Kai Bailey and Nora Kassner and Chizu Kawamoto and Goran Žužić and Victor Gomes and Abhirut Gupta and Michael Guzman and Ishita Dasgupta and Xinyi Bai and Zhufeng Pan and Francesco Piccinno and Hadas Natalie Vogel and Octavio Ponce and Adrian Hutter and Paul Chang and Pan-Pan Jiang and Ionel Gog and Vlad Ionescu and James Manyika and Fabian Pedregosa and Harry Ragan and Zach Behrman and Ryan Mullins and Coline Devin and Aroonalok Pyne and Swapnil Gawde and Martin Chadwick and Yiming Gu and Sasan Tavakkol and Andy Twigg and Naman Goyal and Ndidi Elue and Anna Goldie and Srinivasan Venkatachary and Hongliang Fei and Ziqiang Feng and Marvin Ritter and Isabel Leal and Sudeep Dasari and Pei Sun and Alif Raditya Rochman and Brendan O'Donoghue and Yuchen Liu and Jim Sproch and Kai Chen and Natalie Clay and Slav Petrov and Sailesh Sidhwani and Ioana Mihailescu and Alex Panagopoulos and AJ Piergiovanni and Yunfei Bai and George Powell and Deep Karkhanis and Trevor Yacovone and Petr Mitrichev and Joe Kovac and Dave Uthus and Amir Yazdanbakhsh and David Amos and Steven Zheng and Bing Zhang and Jin Miao and Bhuvana Ramabhadran and Soroush Radpour and Shantanu Thakoor and Josh Newlan and Oran Lang and Orion Jankowski and Shikhar Bharadwaj and Jean-Michel Sarr and Shereen Ashraf and Sneha Mondal and Jun Yan and Ankit Singh Rawat and Sarmishta Velury and Greg Kochanski and Tom Eccles and Franz Och and Abhanshu Sharma and Ethan Mahintorabi and Alex Gurney and Carrie Muir and Vered Cohen and Saksham Thakur and Adam Bloniarz and Asier Mujika and Alexander Pritzel and Paul Caron and Altaf Rahman and Fiona Lang and Yasumasa Onoe and Petar Sirkovic and Jay Hoover and Ying Jian and Pablo Duque and Arun Narayanan and David Soergel and Alex Haig and Loren Maggiore and Shyamal Buch and Josef Dean and Ilya Figotin and Igor Karpov and Shaleen Gupta and Denny Zhou and Muhuan Huang and Ashwin Vaswani and Christopher Semturs and Kaushik Shivakumar and Yu Watanabe and Vinodh Kumar Rajendran and Eva Lu and Yanhan Hou and Wenting Ye and Shikhar Vashishth and Nana Nti and Vytenis Sakenas and Darren Ni and Doug DeCarlo and Michael Bendersky and Sumit Bagri and Nacho Cano and Elijah Peake and Simon Tokumine and Varun Godbole and Carlos Guía and Tanya Lando and Vittorio Selo and Seher Ellis and Danny Tarlow and Daniel Gillick and Alessandro Epasto and Siddhartha Reddy Jonnalagadda and Meng Wei and Meiyan Xie and Ankur Taly and Michela Paganini and Mukund Sundararajan and Daniel Toyama and Ting Yu and Dessie Petrova and Aneesh Pappu and Rohan Agrawal and Senaka Buthpitiya and Justin Frye and Thomas Buschmann and Remi Crocker and Marco Tagliasacchi and Mengchao Wang and Da Huang and Sagi Perel and Brian Wieder and Hideto Kazawa and Weiyue Wang and Jeremy Cole and Himanshu Gupta and Ben Golan and Seojin Bang and Nitish Kulkarni and Ken Franko and Casper Liu and Doug Reid and Sid Dalmia and Jay Whang and Kevin Cen and Prasha Sundaram and Johan Ferret and Berivan Isik and Lucian Ionita and Guan Sun and Anna Shekhawat and Muqthar Mohammad and Philip Pham and Ronny Huang and Karthik Raman and Xingyi Zhou and Ross Mcilroy and Austin Myers and Sheng Peng and Jacob Scott and Paul Covington and Sofia Erell and Pratik Joshi and João Gabriel Oliveira and Natasha Noy and Tajwar Nasir and Jake Walker and Vera Axelrod and Tim Dozat and Pu Han and Chun-Te Chu and Eugene Weinstein and Anand Shukla and Shreyas Chandrakaladharan and Petra Poklukar and Bonnie Li and Ye Jin and Prem Eruvbetine and Steven Hansen and Avigail Dabush and Alon Jacovi and Samrat Phatale and Chen Zhu and Steven Baker and Mo Shomrat and Yang Xiao and Jean Pouget-Abadie and Mingyang Zhang and Fanny Wei and Yang Song and Helen King and Yiling Huang and Yun Zhu and Ruoxi Sun and Juliana Vicente Franco and Chu-Cheng Lin and Sho Arora and Hui and Li and Vivian Xia and Luke Vilnis and Mariano Schain and Kaiz Alarakyia and Laurel Prince and Aaron Phillips and Caleb Habtegebriel and Luyao Xu and Huan Gui and Santiago Ontanon and Lora Aroyo and Karan Gill and Peggy Lu and Yash Katariya and Dhruv Madeka and Shankar Krishnan and Shubha Srinivas Raghvendra and James Freedman and Yi Tay and Gaurav Menghani and Peter Choy and Nishita Shetty and Dan Abolafia and Doron Kukliansky and Edward Chou and Jared Lichtarge and Ken Burke and Ben Coleman and Dee Guo and Larry Jin and Indro Bhattacharya and Victoria Langston and Yiming Li and Suyog Kotecha and Alex Yakubovich and Xinyun Chen and Petre Petrov and Tolly Powell and Yanzhang He and Corbin Quick and Kanav Garg and Dawsen Hwang and Yang Lu and Srinadh Bhojanapalli and Kristian Kjems and Ramin Mehran and Aaron Archer and Hado van Hasselt and Ashwin Balakrishna and JK Kearns and Meiqi Guo and Jason Riesa and Mikita Sazanovich and Xu Gao and Chris Sauer and Chengrun Yang and XiangHai Sheng and Thomas Jimma and Wouter Van Gansbeke and Vitaly Nikolaev and Wei Wei and Katie Millican and Ruizhe Zhao and Justin Snyder and Levent Bolelli and Maura O'Brien and Shawn Xu and Fei Xia and Wentao Yuan and Arvind Neelakantan and David Barker and Sachin Yadav and Hannah Kirkwood and Farooq Ahmad and Joel Wee and Jordan Grimstad and Boyu Wang and Matthew Wiethoff and Shane Settle and Miaosen Wang and Charles Blundell and Jingjing Chen and Chris Duvarney and Grace Hu and Olaf Ronneberger and Alex Lee and Yuanzhen Li and Abhishek Chakladar and Alena Butryna and Georgios Evangelopoulos and Guillaume Desjardins and Jonni Kanerva and Henry Wang and Averi Nowak and Nick Li and Alyssa Loo and Art Khurshudov and Laurent El Shafey and Nagabhushan Baddi and Karel Lenc and Yasaman Razeghi and Tom Lieber and Amer Sinha and Xiao Ma and Yao Su and James Huang and Asahi Ushio and Hanna Klimczak-Plucińska and Kareem Mohamed and JD Chen and Simon Osindero and Stav Ginzburg and Lampros Lamprou and Vasilisa Bashlovkina and Duc-Hieu Tran and Ali Khodaei and Ankit Anand and Yixian Di and Ramy Eskander and Manish Reddy Vuyyuru and Jasmine Liu and Aishwarya Kamath and Roman Goldenberg and Mathias Bellaiche and Juliette Pluto and Bill Rosgen and Hassan Mansoor and William Wong and Suhas Ganesh and Eric Bailey and Scott Baird and Dan Deutsch and Jinoo Baek and Xuhui Jia and Chansoo Lee and Abe Friesen and Nathaniel Braun and Kate Lee and Amayika Panda and Steven M. Hernandez and Duncan Williams and Jianqiao Liu and Ethan Liang and Arnaud Autef and Emily Pitler and Deepali Jain and Phoebe Kirk and Oskar Bunyan and Jaume Sanchez Elias and Tongxin Yin and Machel Reid and Aedan Pope and Nikita Putikhin and Bidisha Samanta and Sergio Guadarrama and Dahun Kim and Simon Rowe and Marcella Valentine and Geng Yan and Alex Salcianu and David Silver and Gan Song and Richa Singh and Shuai Ye and Hannah DeBalsi and Majd Al Merey and Eran Ofek and Albert Webson and Shibl Mourad and Ashwin Kakarla and Silvio Lattanzi and Nick Roy and Evgeny Sluzhaev and Christina Butterfield and Alessio Tonioni and Nathan Waters and Sudhindra Kopalle and Jason Chase and James Cohan and Girish Ramchandra Rao and Robert Berry and Michael Voznesensky and Shuguang Hu and Kristen Chiafullo and Sharat Chikkerur and George Scrivener and Ivy Zheng and Jeremy Wiesner and Wolfgang Macherey and Timothy Lillicrap and Fei Liu and Brian Walker and David Welling and Elinor Davies and Yangsibo Huang and Lijie Ren and Nir Shabat and Alessandro Agostini and Mariko Iinuma and Dustin Zelle and Rohit Sathyanarayana and Andrea D'olimpio and Morgan Redshaw and Matt Ginsberg and Ashwin Murthy and Mark Geller and Tatiana Matejovicova and Ayan Chakrabarti and Ryan Julian and Christine Chan and Qiong Hu and Daniel Jarrett and Manu Agarwal and Jeshwanth Challagundla and Tao Li and Sandeep Tata and Wen Ding and Maya Meng and Zhuyun Dai and Giulia Vezzani and Shefali Garg and Jannis Bulian and Mary Jasarevic and Honglong Cai and Harish Rajamani and Adam Santoro and Florian Hartmann and Chen Liang and Bartek Perz and Apoorv Jindal and Fan Bu and Sungyong Seo and Ryan Poplin and Adrian Goedeckemeyer and Badih Ghazi and Nikhil Khadke and Leon Liu and Kevin Mather and Mingda Zhang and Ali Shah and Alex Chen and Jinliang Wei and Keshav Shivam and Yuan Cao and Donghyun Cho and Angelo Scorza Scarpati and Michael Moffitt and Clara Barbu and Ivan Jurin and Ming-Wei Chang and Hongbin Liu and Hao Zheng and Shachi Dave and Christine Kaeser-Chen and Xiaobin Yu and Alvin Abdagic and Lucas Gonzalez and Yanping Huang and Peilin Zhong and Cordelia Schmid and Bryce Petrini and Alex Wertheim and Jifan Zhu and Hoang Nguyen and Kaiyang Ji and Yanqi Zhou and Tao Zhou and Fangxiaoyu Feng and Regev Cohen and David Rim and Shubham Milind Phal and Petko Georgiev and Ariel Brand and Yue Ma and Wei Li and Somit Gupta and Chao Wang and Pavel Dubov and Jean Tarbouriech and Kingshuk Majumder and Huijian Li and Norman Rink and Apurv Suman and Yang Guo and Yinghao Sun and Arun Nair and Xiaowei Xu and Mohamed Elhawaty and Rodrigo Cabrera and Guangxing Han and Julian Eisenschlos and Junwen Bai and Yuqi Li and Yamini Bansal and Thibault Sellam and Mina Khan and Hung Nguyen and Justin Mao-Jones and Nikos Parotsidis and Jake Marcus and Cindy Fan and Roland Zimmermann and Yony Kochinski and Laura Graesser and Feryal Behbahani and Alvaro Caceres and Michael Riley and Patrick Kane and Sandra Lefdal and Rob Willoughby and Paul Vicol and Lun Wang and Shujian Zhang and Ashleah Gill and Yu Liang and Gautam Prasad and Soroosh Mariooryad and Mehran Kazemi and Zifeng Wang and Kritika Muralidharan and Paul Voigtlaender and Jeffrey Zhao and Huanjie Zhou and Nina D'Souza and Aditi Mavalankar and Séb Arnold and Nick Young and Obaid Sarvana and Chace Lee and Milad Nasr and Tingting Zou and Seokhwan Kim and Lukas Haas and Kaushal Patel and Neslihan Bulut and David Parkinson and Courtney Biles and Dmitry Kalashnikov and Chi Ming To and Aviral Kumar and Jessica Austin and Alex Greve and Lei Zhang and Megha Goel and Yeqing Li and Sergey Yaroshenko and Max Chang and Abhishek Jindal and Geoff Clark and Hagai Taitelbaum and Dale Johnson and Ofir Roval and Jeongwoo Ko and Anhad Mohananey and Christian Schuler and Shenil Dodhia and Ruichao Li and Kazuki Osawa and Claire Cui and Peng Xu and Rushin Shah and Tao Huang and Ela Gruzewska and Nathan Clement and Mudit Verma and Olcan Sercinoglu and Hai Qian and Viral Shah and Masa Yamaguchi and Abhinit Modi and Takahiro Kosakai and Thomas Strohmann and Junhao Zeng and Beliz Gunel and Jun Qian and Austin Tarango and Krzysztof Jastrzębski and Robert David and Jyn Shan and Parker Schuh and Kunal Lad and Willi Gierke and Mukundan Madhavan and Xinyi Chen and Mark Kurzeja and Rebeca Santamaria-Fernandez and Dawn Chen and Alexandra Cordell and Yuri Chervonyi and Frankie Garcia and Nithish Kannen and Vincent Perot and Nan Ding and Shlomi Cohen-Ganor and Victor Lavrenko and Junru Wu and Georgie Evans and Cicero Nogueira dos Santos and Madhavi Sewak and Ashley Brown and Andrew Hard and Joan Puigcerver and Zeyu Zheng and Yizhong Liang and Evgeny Gladchenko and Reeve Ingle and Uri First and Pierre Sermanet and Charlotte Magister and Mihajlo Velimirović and Sashank Reddi and Susanna Ricco and Eirikur Agustsson and Hartwig Adam and Nir Levine and David Gaddy and Dan Holtmann-Rice and Xuanhui Wang and Ashutosh Sathe and Abhijit Guha Roy and Blaž Bratanič and Alen Carin and Harsh Mehta and Silvano Bonacina and Nicola De Cao and Mara Finkelstein and Verena Rieser and Xinyi Wu and Florent Altché and Dylan Scandinaro and Li Li and Nino Vieillard and Nikhil Sethi and Garrett Tanzer and Zhi Xing and Shibo Wang and Parul Bhatia and Gui Citovsky and Thomas Anthony and Sharon Lin and Tianze Shi and Shoshana Jakobovits and Gena Gibson and Raj Apte and Lisa Lee and Mingqing Chen and Arunkumar Byravan and Petros Maniatis and Kellie Webster and Andrew Dai and Pu-Chin Chen and Jiaqi Pan and Asya Fadeeva and Zach Gleicher and Thang Luong and Niket Kumar Bhumihar},
      year={2025},
      eprint={2507.06261},
      archivePrefix={arXiv},
      primaryClass={cs.CL},
      url={https://arxiv.org/abs/2507.06261}, 
}

@misc{fu2025videommefirstevercomprehensiveevaluation,
      title={Video-MME: The First-Ever Comprehensive Evaluation Benchmark of Multi-modal LLMs in Video Analysis}, 
      author={Chaoyou Fu and Yuhan Dai and Yongdong Luo and Lei Li and Shuhuai Ren and Renrui Zhang and Zihan Wang and Chenyu Zhou and Yunhang Shen and Mengdan Zhang and Peixian Chen and Yanwei Li and Shaohui Lin and Sirui Zhao and Ke Li and Tong Xu and Xiawu Zheng and Enhong Chen and Caifeng Shan and Ran He and Xing Sun},
      year={2025},
      eprint={2405.21075},
      archivePrefix={arXiv},
      primaryClass={cs.CV},
      url={https://arxiv.org/abs/2405.21075}, 
}

@techreport{anthropic2025systemcard,
  title        = {System Card: Claude Opus 4 \& Claude Sonnet 4},
  author       = {{Anthropic PBC}},
  institution  = {Anthropic PBC},
  year         = {2025},
  month        = may,
  url          = {https://www-cdn.anthropic.com/4263b940cabb546aa0e3283f35b686f4f3b2ff47.pdf},
  note         = {Accessed: 2025-07-27}
}

@misc{kwon2023efficientmemorymanagementlarge,
      title={Efficient Memory Management for Large Language Model Serving with PagedAttention}, 
      author={Woosuk Kwon and Zhuohan Li and Siyuan Zhuang and Ying Sheng and Lianmin Zheng and Cody Hao Yu and Joseph E. Gonzalez and Hao Zhang and Ion Stoica},
      year={2023},
      eprint={2309.06180},
      archivePrefix={arXiv},
      primaryClass={cs.LG},
      url={https://arxiv.org/abs/2309.06180}, 
}

@misc{zheng2024llamafactoryunifiedefficientfinetuning,
      title={LlamaFactory: Unified Efficient Fine-Tuning of 100+ Language Models}, 
      author={Yaowei Zheng and Richong Zhang and Junhao Zhang and Yanhan Ye and Zheyan Luo and Zhangchi Feng and Yongqiang Ma},
      year={2024},
      eprint={2403.13372},
      archivePrefix={arXiv},
      primaryClass={cs.CL},
      url={https://arxiv.org/abs/2403.13372}, 
}

@book{APA2013DSM5,
  author       = {{American Psychiatric Association}},
  title        = {Diagnostic and Statistical Manual of Mental Disorders (DSM-5\textregistered)},
  series       = {Diagnostic and Statistical Manual of Mental Disorders},
  edition      = {5th},
  publisher    = {American Psychiatric Publishing},
  address      = {Arlington, VA},
  year         = {2013},
  month        = May,
  pages        = {947},
  isbn         = {9780890425541},
  doi          = {10.1176/appi.books.9780890425596},
  url          = {https://doi.org/10.1176/appi.books.9780890425596}
}

@article{gorwa2020algorithmic,
author = {Gorwa, Robert and Binns, Reuben and Katzenbach, Christian},
year = {2020},
month = {02},
pages = {205395171989794},
title = {Algorithmic content moderation: Technical and political challenges in the automation of platform governance},
volume = {7},
journal = {Big Data \& Society},
doi = {10.1177/2053951719897945}
}

@book{gillespie2018custodians,
  author    = {Gillespie, Tarleton},
  title     = {Custodians of the Internet: Platforms, Content Moderation, and the Hidden Decisions That Shape Social Media},
  year      = {2018},
  month     = jun,
  publisher = {Yale University Press},
  address   = {New Haven, Connecticut},
  pagetotal = {288},
  isbn      = {9780300235029},
  doi       = {10.12987/9780300235029},
  url       = {https://yalebooks.yale.edu/book/9780300235029/custodians-of-the-internet/}
}

@misc{kiela2020supervisedmultimodalbitransformersclassifying,
      title={Supervised Multimodal Bitransformers for Classifying Images and Text}, 
      author={Douwe Kiela and Suvrat Bhooshan and Hamed Firooz and Ethan Perez and Davide Testuggine},
      year={2020},
      eprint={1909.02950},
      archivePrefix={arXiv},
      primaryClass={cs.CL},
      url={https://arxiv.org/abs/1909.02950}, 
}

@inproceedings{wang2017detecting,
  title={Detecting and characterizing eating-disorder communities on social media},
  author={Wang, Tao and Brede, Markus and Ianni, Antonella and Mentzakis, Emmanouil},
  booktitle={Proceedings of the Tenth ACM International Conference on Web Search and Data Mining},
  pages={91--100},
  year={2017},
  doi={10.1145/3018661.3018706}
}

@article{merhbene2024investigating,
  title   = {Investigating machine learning and natural language processing techniques applied for detecting eating disorders: a systematic literature review},
  author  = {Merhbene, Ghofrane and Puttick, Alexandre and Kurpicz-Briki, Mascha},
  journal = {Frontiers in Psychiatry},
  year    = {2024},
  volume  = {15},
  pages   = {1319522},
  month   = mar,
  doi     = {10.3389/fpsyt.2024.1319522},
  url     = {https://doi.org/10.3389/fpsyt.2024.1319522}
}

\appendix

\section*{Appendix}

\section{Taxonomy}

\subsection{Categories Definition}
\label{sec:cat_definition}

A detailed definition of the categories in our taxonomy is provided in Table~\ref{tab:taxonomy_full}. These categories are carefully curated and iteratively refined in collaboration with clinical psychologists and domain researchers to ensure both validity and practicality.  

Throughout taxonomy development, clinical experts engage in extensive deliberation over edge cases that reveal the central challenge of drawing the line between harmful and legitimate fitness content---a distinction requiring specialized clinical expertise, real-world patient experience, and collaborative consensus-building. For instance, the \textit{Maladaptive Coping} subcategory sparks debate about whether content promoting exercise as emotional regulation constitutes harm, with experts weighing clinical distinctions between healthy stress management and problematic avoidance behaviors. Similarly, \textit{Legal APEDs} discussions center on sarcastic supplement content, where experts consider whether adult humor might inadvertently normalize excessive consumption among adolescent audiences. These deliberations require experts to balance contextual factors, including target audience, creator intent versus potential impact, and clinical thresholds between enthusiasm and pathology. Through structured consensus sessions examining ambiguous cases, annotators establish decision rules that balance clinical rigor with practical annotation consistency.

\begin{table*}[ht!]
\centering
\small
\begin{tabular}{p{0.12\textwidth}|p{0.15\textwidth}|p{0.35\textwidth}|p{0.25\textwidth}}
\toprule
\textbf{Primary} & \textbf{Secondary} & \textbf{Definition} & \textbf{Keywords} \\
\midrule
\multirow{3}{0.12\textwidth}{Relationship to Body} 
  & Muscularity Self‐Objectification 
    & Idealized imagery emphasizing muscular aesthetics as the primary source of value. 
    & shredded, swole, mensphysique, bodybuilding \\
\cmidrule(lr){2-4}
  & Leanness Self‐Objectification 
    & Self‐evaluation against a lean, low‐fat, highly‐toned ideal promoted online. 
    & flat tummy, skinny men physique, small waist fitness \\
\cmidrule(lr){2-4}
  & Muscle Dissatisfaction 
    & Expressing perceived insufficient muscularity despite fitness, fueling negative self‐view. 
    & muscles never big enough, muscles not big enough, muscle dysmorphia, bdd men \\
\midrule
\multirow{3}{0.12\textwidth}{Relationship to Food} 
  & Rigid Food Rules 
    & Obsessive macro/micronutrient tracking and extreme bulking or cutting diets. 
    & aggressive cut, aggressive bulk, shredded diet, macro tracking \\
\cmidrule(lr){2-4}
  & Unsafe Food 
    & Promotion of raw or unsafe foods believed to enhance muscle growth. 
    & liver king diet, raw meat diet for gym, dog food to gain muscles \\
\cmidrule(lr){2-4}
  & Cheat Meals 
    & Large ``reward'' meals after restrictive dieting that reinforce binge–compensation cycles. 
    & cheatday food, cheatmeal \\
\midrule
\multirow{5}{0.12\textwidth}{Relationship to Exercise} 
  & Excessive Exercise 
    & Obsessive routines exceeding healthy limits despite injury or life interference. 
    & no rest day, david goggins mentality, train until failure, push your limit \\
\cmidrule(lr){2-4}
  & Predebting 
    & Treating exercise as punishment or permission to eat, creating guilt cycles. 
    & exercise so I can eat, workout so I can eat, earn your food, train so I can eat \\
\cmidrule(lr){2-4}
  & Maladaptive Coping 
    & Using exercise as the sole coping mechanism to avoid emotional distress. 
    & gym therapy, workout breakup, workout heartbreak, gym mental health, gym fixes everything \\
\cmidrule(lr){2-4}
  & Exercise‐Induced  Functional Impairment 
    & Prioritizing exercise over essential duties, harming daily functioning. 
    & gym over everything, gym or nothing, gym over friends, skip school for gym \\
\cmidrule(lr){2-4}
  & Toxic Motivation 
    & Demeaning communication that pressures unrealistic fitness standards via shaming or slurs. 
    & gym masculinity, aggressive gym motivation, they don't know me son, get your ass to the gym \\
\midrule
\multirow{3}{0.12\textwidth}{Supplements} 
  & Anabolic Steroids 
    & Normalization or endorsement of anabolic‐androgenic steroid use with downplayed risks. 
    & tren, anabolic stack, anabolic gear, steroids \\
\cmidrule(lr){2-4}
  & Legal APEDs 
    & Overuse of legal supplements (e.g., creatine, protein, preworkout) beyond recommended doses. 
    & protein powder, whey, creatine, pre workout \\
\cmidrule(lr){2-4}
  & Hormone Therapy 
    & Downplaying risks and spreading misinformation about testosterone replacement therapy. 
    & TRT for gains, testosterone, hormoneboost \\
\midrule
\multirow{1}{0.12\textwidth}{Relationship to Masculinity} 
  & 
    & Links muscle‐building and exercise performance to male identity, sexuality, and self‐worth. 
    & gym masculinity, be a man gym, embrace masculinity, alpha male gym \\
\midrule
\multirow{1}{0.12\textwidth}{Irrelevant} 
  & 
    & General fitness or lifestyle content unrelated to pro–bigorexia harm. 
    & fyp, tiktok, foryoupage, viral, funny, duet, trending, love, meme, followme, repost, new, awesome, music, cute, video, foryou, fun, diy, ootd, family, lifehack, photography, usa, college, travel, christmas, sport, party, popular, clip, movie, star, moment, tiktokviral, tiktokfamous, tiktokmusic, likeforfollow, recipe, quotes, TikTokChallenge, memories \\
\bottomrule
\end{tabular}
\caption{Primary and secondary categories of the Harm Taxonomy for Pro‑Bigorexia Content on TikTok, with definitions and search keywords.} 
\label{tab:taxonomy_full}
\end{table*}

\section{Annotation}

\subsection{Annotator Profiles}
\label{sec:annotator}

We enlist 16 subject‐matter experts, ranging from clinical psychologists and social workers to computational social scientists, for video annotation. Our annotators are research collaborators and co-authors who volunteered their clinical expertise with full knowledge of the project's scope and scientific objectives, ensuring informed participation in this sensitive content annotation task. Table~\ref{tab:annotators} details each annotator’s ID, batch assignment, area of expertise, and gender.

\subsection{Annotation Instructions}
\label{sec:annon_instruction}

Annotators received comprehensive guidelines emphasizing safety protocols and decision-making frameworks for handling potentially disturbing content. Key instructions included: (1) \textbf{Content Safety}: annotators were advised to work at their own pace, take regular breaks, and prioritize mental well-being when reviewing harmful content; (2) \textbf{Decision Framework}: when encountering ambiguous cases, annotators were instructed to select ``Unsure, need consultation'' and provide explanatory notes, with textual information (captions, on-screen text) prioritized over conflicting visual content; (3) \textbf{Consistency Maintenance}: annotators were asked to reflect on their annotation patterns periodically to avoid drift and maintain consistency across the 300-video batches; (4) \textbf{Expert Support}: availability of research coordinators via dedicated communication channels for immediate consultation on challenging cases. 

Annotators assess harm severity based on message intensity, including explicit display of muscular physiques, toxic motivational pressure toward unattainable muscularity, rigid and unrealistic dietary demands, endorsement of high-risk behaviors such as steroid use or dangerous workout practices, and the degree of pathological behavioral patterns exhibited in the content.

These protocols ensured both annotator safety and annotation quality while acknowledging the inherently difficult nature of distinguishing harmful pro-bigorexia content from benign fitness material.

\subsection{Annotation Platform}
\label{sec:annon_platform}

The HTML interface (Figure~\ref{fig:mturk_interface}) presents each TikTok video with its caption and hashtags (if any). The form beneath the video prompts the annotator to mark if the video is: (1) \textit{Showing One’s Body} (yes/no); includes (2) \textit{Commercial Sponsorship} (yes/no); (3) 2 boxes for \textit{Type of harm} (1 mandatory, 1 optional), consisting of primary categories and subcategories, plus ``Relevant but Not Listed,'' ``Irrelevant,'' and ``Unsure—Needs Consultation'' options; and (4) \textit{Severity of Harm} on a discrete scale ranging from 1 (Not Harmful) to 5 (Extremely Harmful). 
A text box allows annotators to write additional comments and edge‐case observations.  

\begin{figure}[!ht]
  \centering
  \includegraphics[width=0.5\textwidth]{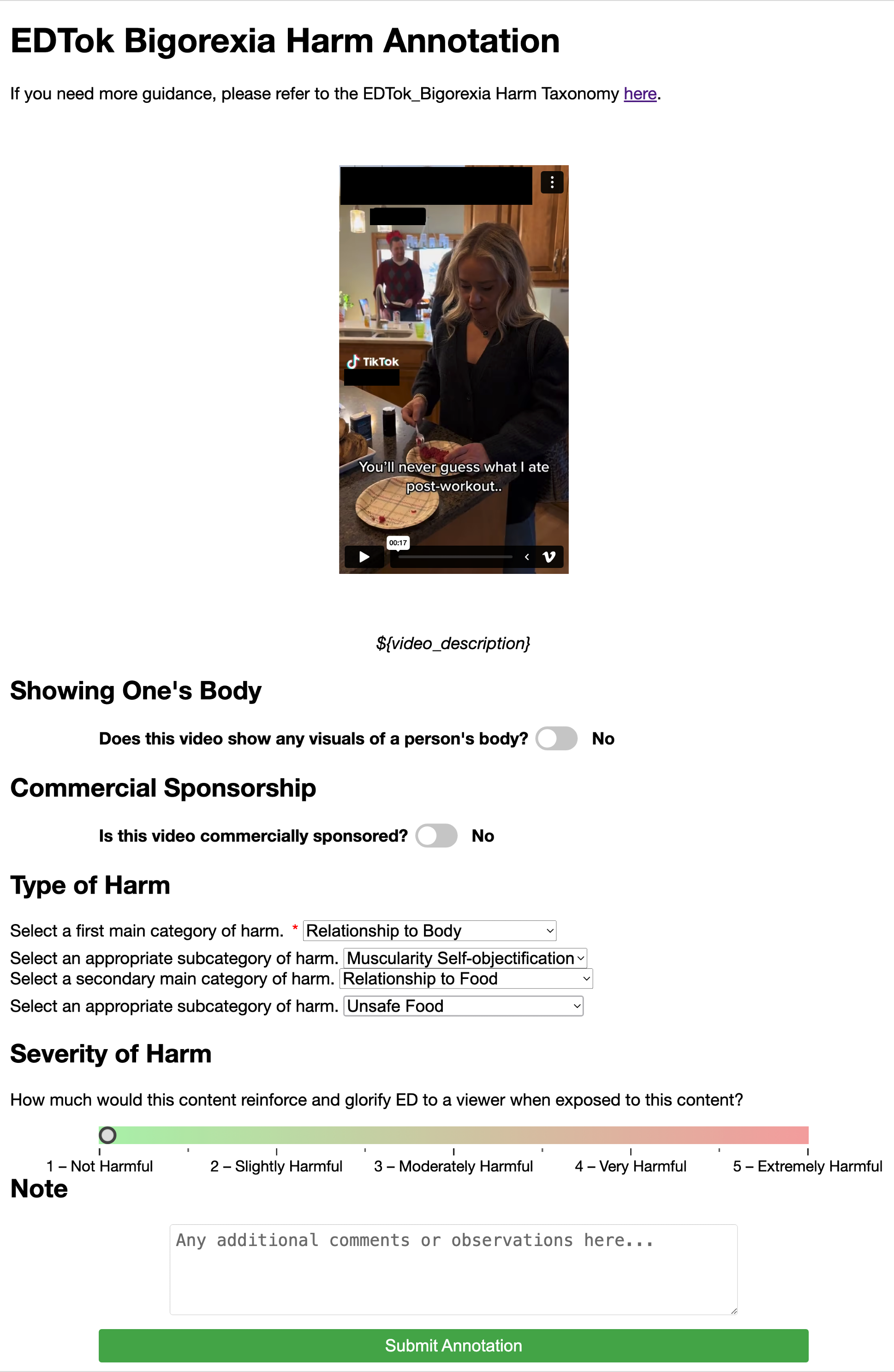}
  \caption{Annotation Interface on Amazon Mechanical Turk platform.}
  \label{fig:mturk_interface}
\end{figure}

\subsection{Annotation Processing}
\label{sec:annon_proc}

After Round 2, we apply a five‐step pipeline to prepare the data for consensus review and final curation:
\begin{enumerate}[noitemsep,topsep=0pt]
  \item \textbf{Annotation Completeness:} verify that each video has exactly two independent annotations.
  \item \textbf{Data Filtering:} remove videos that are unavailable (deleted) or non‐English, as indicated in annotators’ notes.
  \item \textbf{Reliability Assessment:} compute Cohen’s $\kappa$ for both strong (exact primary–secondary match) and weak (primary‐only match) agreement to quantify inter‐rater reliability.
  \item \textbf{Category Refinement Candidates:} collect all videos flagged as ``Unsure—Needs Consultation'' or ``Relevant but Not Listed.'' For these, we consult with the annotators to either reassign them to existing categories, extend the taxonomy to cover frequently emerging content types, or mark them as irrelevant.
  \item \textbf{Consensus Review and Re‐annotation:} filter out all items by agreement level (strong, weak, disagreement), sample a subset of weak and disagreement cases for each annotator pair, hold a batch‐specific meeting to discuss and jointly re‐annotate selected cases to reach consensus, then return the remaining weak and disagreement items to annotators for independent re‐annotation.
\end{enumerate}

This pipeline ensures data consistency, highlights contentious cases for Round 3 consensus sessions, and informs any necessary taxonomy updates.

\subsection{Annotation Statistics}
\label{sec:annon_stats}

Inter-annotator agreement substantially improves from Round 2 to Final (Tables~\ref{fig:annon_dis_type} and \ref{fig:annon_dis_subtype}), with Cohen's $\kappa$ values increasing from moderate agreement (0.43-0.69 strong $\kappa$, 0.59-0.83 weak $\kappa$) to good-to-excellent agreement (0.58-0.81 strong $\kappa$, 0.78-0.94 weak $\kappa$). The consistently high weak $\kappa$ values (>0.9 in most batches) in the Final round indicate that annotators achieved near-perfect agreement on the broader annotation categories.

\begin{table}[ht]
\centering
\small
\begin{tabular}{c|cc|cc}
\hline
Batch & \multicolumn{2}{c|}{Round 2} & \multicolumn{2}{c}{Final} \\ 
      & Strong $k$      & Weak $k$      & Strong $k$     & Weak $k$     \\ 
\hline
1     & 0.551           & 0.613         & 0.616          & 0.775        \\ 
2     & 0.570           & 0.588         & 0.762          & 0.841        \\ 
3     & 0.510           & 0.612         & 0.680          & 0.815        \\ 
4     & 0.551           & 0.785         & 0.811          & 0.937        \\ 
5     & 0.563           & 0.685         & 0.787          & 0.924        \\ 
6     & 0.555           & 0.723         & 0.742          & 0.938        \\ 
7     & 0.430           & 0.739         & 0.577          & 0.931        \\ 
8     & 0.685           & 0.825         & 0.769          & 0.938        \\ 
\hline
\end{tabular}
\caption{Cohen's $\kappa$ coefficients for inter-annotator agreement across annotation rounds. \textbf{Strong $\kappa$} denotes exact agreement on the Subcategory label; \textbf{Weak $\kappa$} denotes agreement on the Primary Category only. Reliability improved substantially in the Final round following consensus adjudication.}
\label{tab:kappa}
\end{table}

\begin{table}[ht]
  \centering
  \small
  \begin{subtable}[t]{0.45\textwidth}
    \centering
    \begin{tabular}{l r}
      \toprule
      \textbf{Primary Category} & \textbf{Count} \\
      \midrule
      Relationship to Body                & 1442 \\
      Relationship to Exercise            & 1233 \\
      Relationship to Food                & 824  \\
      Supplement Abuse                    & 558  \\
      Relevant but Not Harmful            & 411  \\
      Relationship to Masculinity         & 406  \\
      Irrelevant                          & 401  \\
      \bottomrule
    \end{tabular}
    \caption{Primary Categories}
    \label{tab:primary_categories}
  \end{subtable}%
  \quad
  \begin{subtable}[t]{0.45\textwidth}
    \centering
    \begin{tabular}{l r}
      \toprule
      \textbf{Subcategory} & \textbf{Count} \\
      \midrule
      Muscularity Self-objectification & 905 \\
      Leanness Self-objectification   & 238 \\
      Muscle Dissatisfaction          & 240 \\
      \cmidrule(lr){1-2}
      Rigid Food Rules                & 375 \\
      Cheat Meals                     & 193 \\
      \cmidrule(lr){1-2}
      Excessive Exercise              & 379 \\
      Maladaptive Coping              & 213 \\
      Toxic Motivation                & 308 \\
      \cmidrule(lr){1-2}
      Anabolic Steroids               & 195 \\
      \cmidrule(lr){1-2}
      Other (please specify in Note)  & 217 \\
      \bottomrule
    \end{tabular}
    \caption{Subcategories}
    \label{tab:subcategories}
  \end{subtable}
  \caption{Distribution of raw individual annotations (before consensus aggregation). Since each video received two independent annotations, these counts reflect the total number of labels assigned by annotators, whereas Table~\ref{tab:label_distribution} reflects the final unique video counts after aggregation.}
  \label{tab:annotation_count}
\end{table}

\section{Data}
\label{sec:data_collection}
\label{sec:model_data}

We collected videos through the official TikTok Research API after submitting a research proposal that was reviewed and approved by TikTok. This API accesses publicly posted content where users have consented to public visibility through TikTok's terms of service. We implemented additional privacy protections by anonymizing user identifiers and exposing only video content and captions to annotators.

Our dataset comprises TikTok videos annotated for hierarchical classification across two levels of granularity. The primary category classification categorizes content into six broad categories related to body image and health behaviors, while the subcategory classification provides fine-grained classification into 16 specific subcategories. The severity score scale defines the intensity of the harm emotion. The severity scores are most heavily concentrated in the 1.0 to 1.5 range (Figure~\ref{fig:sev_scores}), indicating that lower severity levels are the most common. The frequency then gradually decreases as scores increase, with relatively few cases exceeding 4.0. This suggests that mild severity is predominant in the dataset. Figure~\ref{fig:sev_by_type} shows that most content falls in the moderate-harm range, though \textit{Supplement Abuse} and \textit{Relationship to Food} score higher on average, likely because markers such as extreme dieting practices or steroid use are more readily identifiable to annotators.

We remove corrupted files (e.g., missing metadata) and non-English videos during preprocessing. The dataset is split into train/test sets, maintaining approximately a 3:1 ratio with stratified sampling based on task labels and downsampling of dominant categories to improve balance.

As shown in Table~\ref{tab:label_distribution}, the primary category task contains 1,966 training and 588 test samples. Training data exhibits natural class imbalance reflecting real-world distributions, ranging from 489 samples (``Relationship to Body'') to 66 samples (``Relationship to Masculinity''), while the test set maintains a strictly balanced representation (98 samples per category) for fair evaluation. The subcategory task operates on a filtered subset of 1,472 training and 462 test samples. Training samples range from 200 (``Irrelevant'') to 28 (``Predebting Exercise''), while test data maintains a balanced distribution for major categories (34 samples each) with reduced representation for rare categories (4--10 samples).

Zero-shot and few-shot approaches are evaluated solely on test sets, while finetuning models are trained on the respective task's training set and evaluated on the corresponding test set.

\begin{figure}[h]
  \centering
  \includegraphics[width=\columnwidth]{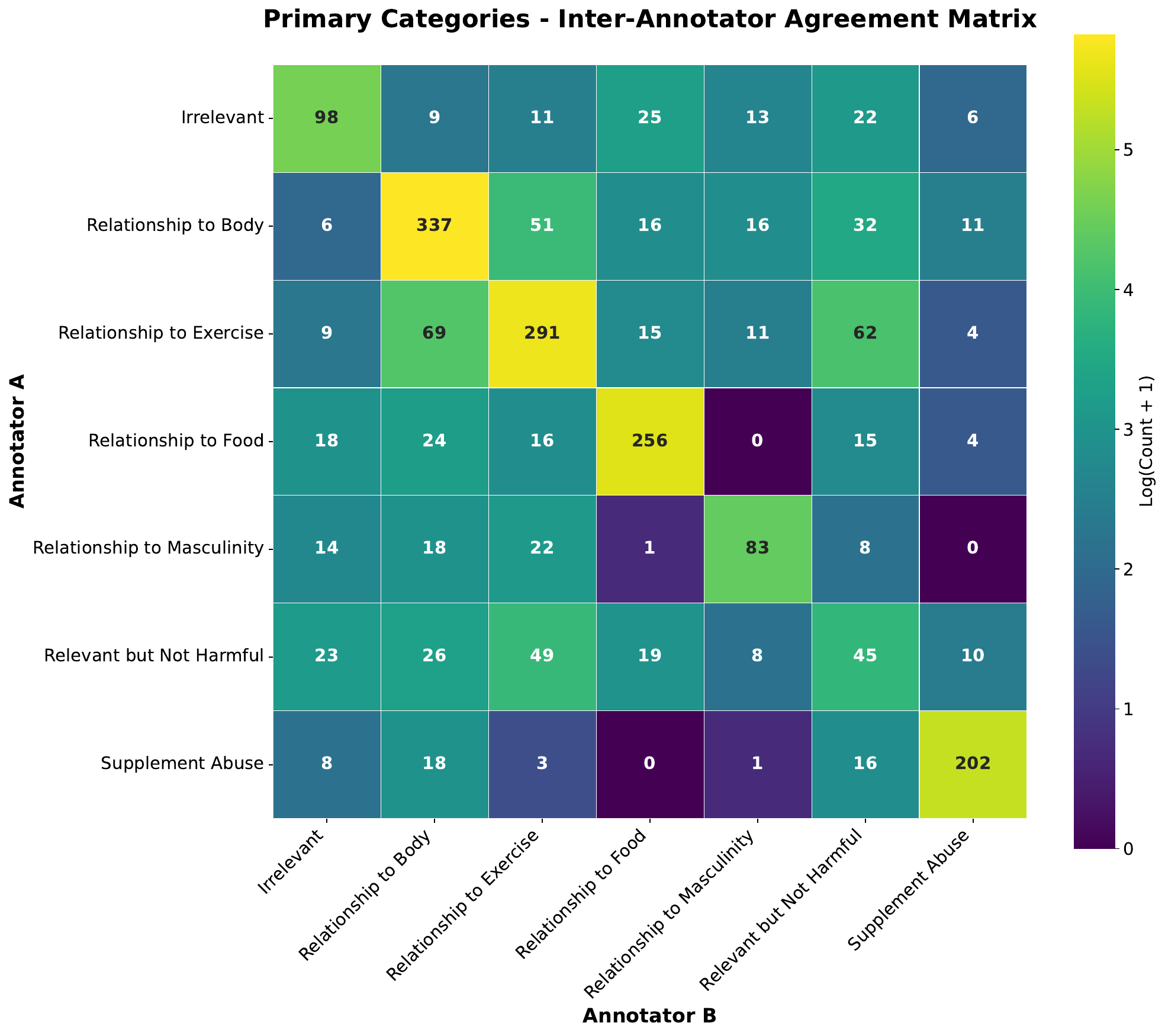}
  \caption{Inter-annotator agreement matrix for primary categories. Each cell shows the number of videos where Annotator A assigned the row category and Annotator B assigned the column category. Diagonal elements represent perfect agreement, while off-diagonal elements indicate disagreements between annotators.}
  \label{fig:annon_dis_type}
\end{figure}

\begin{figure}[h]
  \centering
  \includegraphics[width=\columnwidth]{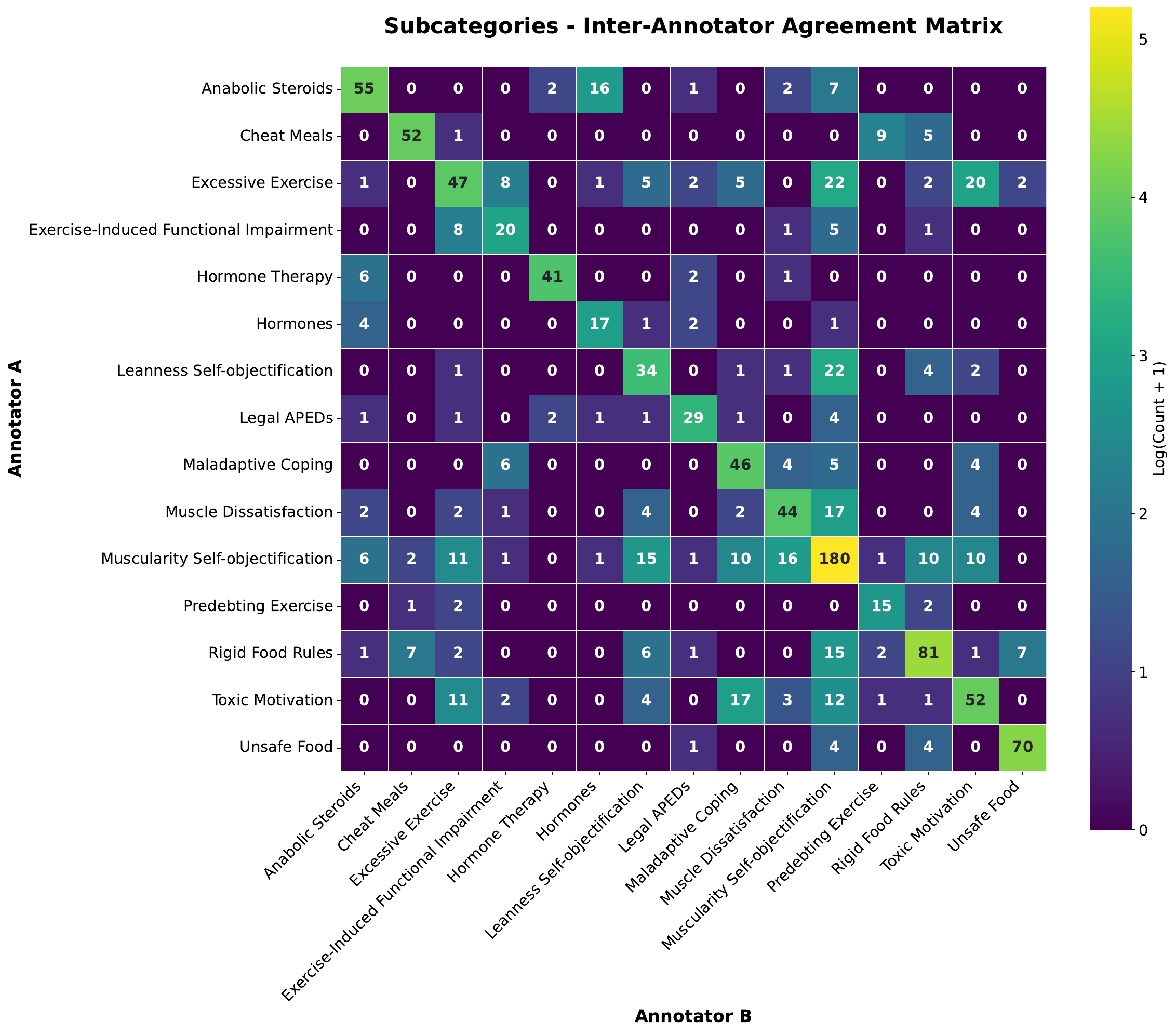}
  \caption{Inter-annotator agreement matrix for subcategories. Each cell shows the number of videos where Annotator A assigned the row subcategory and Annotator B assigned the column subcategory. The matrix reveals patterns of confusion between semantically related subcategories and overall annotation consistency.}
  \label{fig:annon_dis_subtype}
\end{figure}

\begin{figure}[h]
  \centering
  \includegraphics[width=\columnwidth]{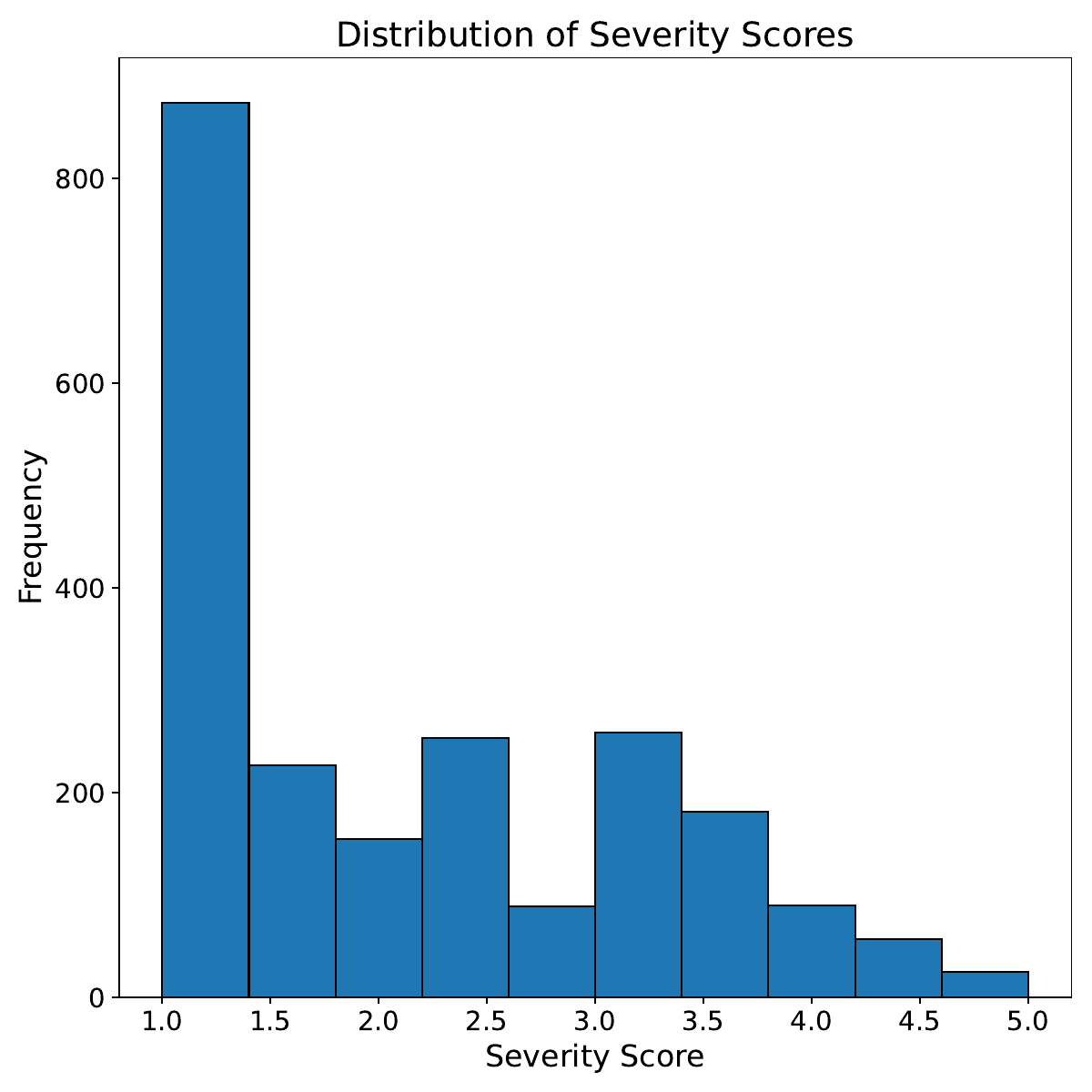}
  \caption{Distribution of Severity Scores. The histogram shows most values concentrated between 1.0 and 2.0, with fewer cases at higher severity levels.}
  \label{fig:sev_scores}
\end{figure}

\begin{figure}[h]
  \centering
  \includegraphics[width=\columnwidth]{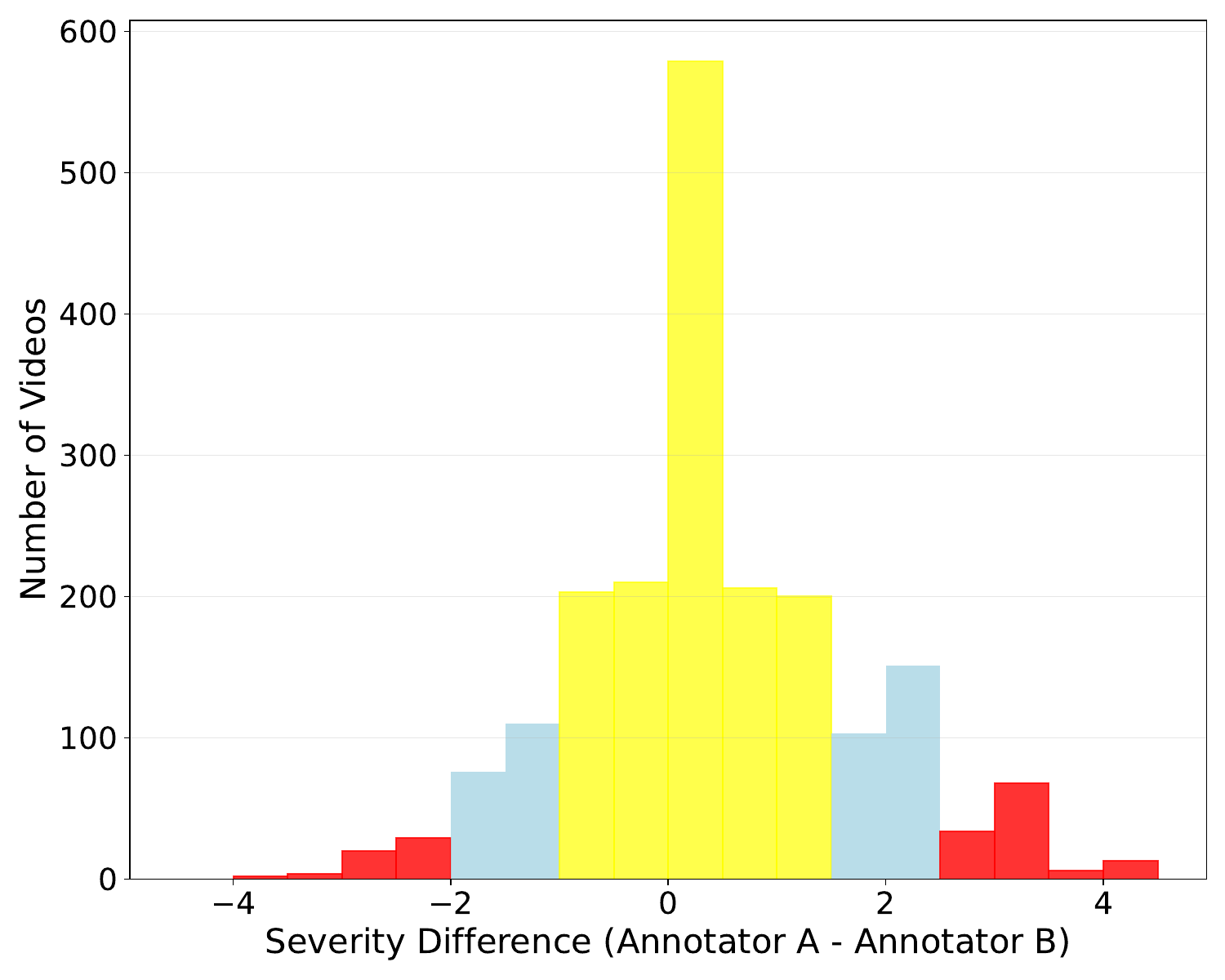}
  \caption{Histogram of severity score differences between annotators. Positive values indicate higher severity by Annotator A. Colors denote agreement levels: green (exact), yellow ($\pm$1 point), red ($>$2 points).}
  \label{fig:sev_diff}
\end{figure}

\begin{figure}[h]
  \centering
  \includegraphics[width=\columnwidth]{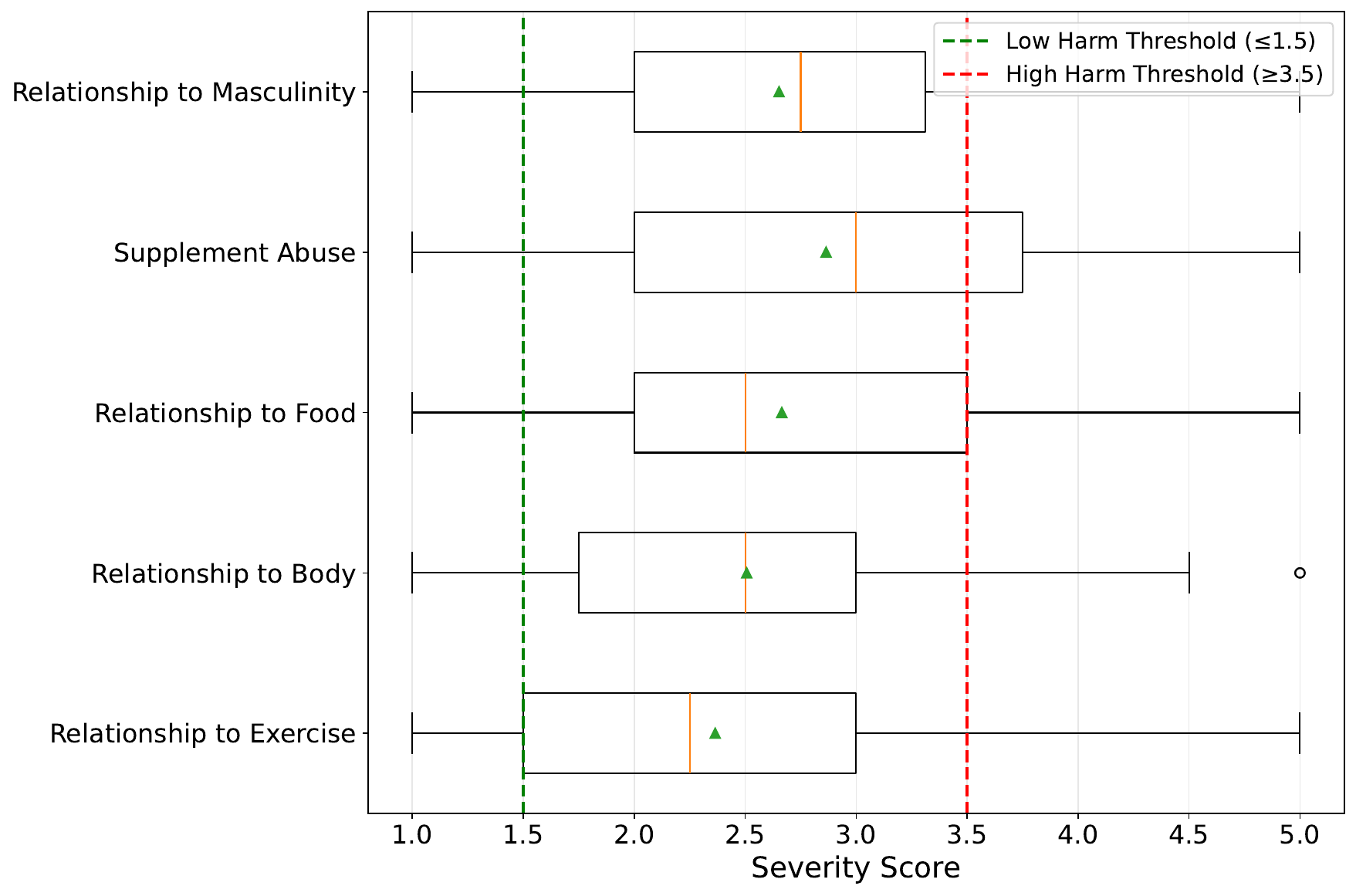}
  \caption{Box plot of severity scores across the primary types. Each box shows the distribution of annotated severity scores, with mean markers included. Dashed vertical lines indicate the harm thresholds, distinguishing low-harm content ($\le$ 1.5) from high-harm content ($\ge$ 3.5)}
  \label{fig:sev_by_type}
\end{figure}

\section{Modeling}

\subsection{Experiment Setup}
\label{sec:setup}

We implement the VLMs using Transformers-based (for InternVL3) and vLLM~\cite{kwon2023efficientmemorymanagementlarge} (for Qwen-2.5VL) implementations through the LlamaFactory framework~\cite{zheng2024llamafactoryunifiedefficientfinetuning}, which provides efficient inference and serving capabilities for VLMs. We used pre-trained VLMs according to their respective licenses and terms of use: commercial API-based models (GPT-4.1, Claude-Sonnet-4, Gemini-2.5-Flash) under their standard API terms, and open-source models (InternVL3, Qwen2.5-VL) under their permissive licenses for research use.

Due to hardware constraints, we exclude InternVL-38B and InternVL-78VL from certain experiments. Our current infrastructure is incompatible with the vLLM version required for multi-node deployment of these larger models. Since these models exceed the memory capacity available on single nodes in our cluster, we cannot accommodate their substantial memory requirements within our computational environment.

We consider conducting experiments with LLaVA-NeXT-Video~\cite{Liu2024LLaVANeXT}, which achieved SOTA open-source performance on Video-MME~\cite{fu2025videommefirstevercomprehensiveevaluation}. However, we exclude LLaVA-NeXT-Video from our study since its context window limitation (4,096 tokens) cannot accommodate our prompting setup (text + video/frames), which often exceeds that limit.

All zero-shot experiments on open-source VLMs are conducted on 8 $\times$ NVIDIA H100 GPUs. For few-shot and finetuning experiments, we use 32 $\times$ NVIDIA A100 GPUs for Qwen2.5-VL-7B, Qwen2.5-VL-32B, InternVL3-8B, and InternVL3-38B, and 64 $\times$ NVIDIA A100 GPUs for Qwen2.5-VL-72B.

We focus on evaluating two tasks: primary category and subcategory classification. For each task, we split the data into training and test sets at a 3:1 ratio using stratified sampling. The test set is strictly balanced across all classes, while the training set is adjusted by downsampling the majority categories to mitigate class imbalance. Detailed split statistics and sampling procedures are provided in Appendix~\ref{sec:model_data}.

\begin{table}[ht]
\centering
\small
\begin{tabular}{p{2.1cm}|p{1.5cm}|p{1.3cm}|p{1cm}}
\toprule
\textbf{Model} & \textbf{Version} & \textbf{Size (B)} & \textbf{Type} \\
\midrule
GPT-4o              & 2024-08-06       & —             & API           \\
Claude Sonnet 4     & 2024-02-24       & —             & API           \\
Gemini 2.5 Pro      & 2025-06-17       & —             & API           \\
\cmidrule(lr){1-4}
InternVL3           & 2024-10-04       & 8, 38  & Open  \\
Qwen2.5‑VL          & 2025-06-05       & 7, 32, 72  & Open   \\
\bottomrule
\end{tabular}
\caption{Model version dates, parameter counts, and types (API vs.\ Open-source) for video‐based VLMs.}
\label{tab:models}
\end{table}

\subsection{Prompt Templates}
\label{sec:prompt}

We use a JSON-based prompt structure following the standard role-content format for chat-based LLMs called `sharegpt'. Each prompt consists of:
\begin{itemize}
  \item A sequence of images from the video (sampled frames)
  \item A user message containing the video caption, embedded text and audio transcription
  \item A single assistant response with the predicted label
\end{itemize}

An example format of prompt used for zero-shot inference/supervised finetuning and few-shot inference is shown in Figures~\ref{fig:vlm_zeroshot_prompt_type} and \ref{fig:vlm_zeroshot_prompt_subtype} and  Figure~\ref{fig:vlm_fewshot_prompt_type} and \ref{fig:vlm_fewshot_prompt_subtype}.

\section{Results}

\subsection{Hyperparameter Robustness Check}
\label{sec:robustness}

To validate our choice of using default temperature settings ($T=1.0$) for commercial API models, we re-evaluated Task 1 (Primary Category Classification) using a deterministic setting ($T=0.1$). Table~\ref{tab:robustness_check} compares the performance across both settings.

The results demonstrate that performance remains highly stable, with F1 score variations confined within a narrow range ($\pm 0.01$). For instance, Claude-Sonnet-4 shows a slight improvement at $T=0.1$ (+0.008 F1), while GPT-4.1 shows a minor decrease (-0.004 F1). These marginal differences indicate that the specific temperature choice for high-performing commercial models does not significantly alter the comparative conclusions drawn in the main paper.

\begin{table}[h]
\centering
\small
\setlength{\tabcolsep}{5pt}
\begin{tabular}{lccccc}
\toprule
\textbf{Model} & \textbf{Temp} & \textbf{Acc.} & \textbf{P\textsubscript{m}} & \textbf{R\textsubscript{m}} & \textbf{F1\textsubscript{m}} \\
\midrule
Claude-Sonnet-4 & 1.0 & 0.829 & 0.832 & 0.829 & 0.827 \\
Claude-Sonnet-4 & 0.1 & \textbf{0.837} & \textbf{0.841} & \textbf{0.837} & \textbf{0.835} \\
\midrule
GPT-4.1 & 1.0 & \textbf{0.796} & 0.808 & \textbf{0.792} & \textbf{0.792} \\
GPT-4.1 & 0.1 & 0.787 & \textbf{0.809} & 0.787 & 0.788 \\
\midrule
Gemini-2.5-Flash & 1.0 & 0.805 & 0.807 & 0.805 & 0.805 \\
Gemini-2.5-Flash & 0.1 & \textbf{0.810} & \textbf{0.813} & \textbf{0.810} & \textbf{0.809} \\
\bottomrule
\end{tabular}
\caption{Robustness check for commercial API models on Task 1 (Primary Category) comparing default ($T=1.0$) vs. deterministic ($T=0.1$) temperature settings. $P_m$, $R_m$, and $F1_m$ denote macro precision, recall, and F1.}
\label{tab:robustness_check}
\end{table}

\subsection{Per-Subtype Detection Analysis}
\label{sec:per_type}

Table~\ref{tab:per_class_subtype} details the F1 scores for all 15 subcategories, revealing significant performance variance driven by content explicitness.

\paragraph{Explicitness Hierarchy}
Subcategories characterized by distinct lexical keywords or unambiguous visual objects achieve the highest detection rates. \textit{Cheat Meals} (Claude-4 ZS: 0.899) and \textit{Supplement Abuse} subtypes (e.g., \textit{Anabolic Steroids}, \textit{Hormone Therapy}) consistently achieve F1 scores above 0.80, benefiting from clear indicators like specific drug names (e.g., ``tren'', ``TRT''). Conversely, performance drops sharply for ambiguous categories requiring subjective behavioral interpretation, such as \textit{Muscularity Self-objectification} and \textit{Excessive Exercise} ($<0.60$ F1), reflecting the difficulty in distinguishing harmful behaviors from normative training without deep context.

\paragraph{Model-Specific Strengths}
Commercial models demonstrate superior reasoning capabilities on psychologically nuanced categories, with Claude-4 (Few-Shot) achieving the highest score on \textit{Maladaptive Coping} (0.787). In contrast, open-source models like InternVL3-38B excel in categories with strong visual or explicit signals, leading performance in \textit{Unsafe Food} and \textit{Anabolic Steroids} (Few-Shot: 0.831), confirming their utility for detecting overtly harmful content.

\begin{table*}[!ht]
\centering
\scriptsize
\begin{tabular}{l|cc|c|cc|ccc|ccc}
\toprule
& \multicolumn{5}{c|}{\textbf{Commercial Models}} & \multicolumn{6}{c}{\textbf{Open-Source Models}} \\
\cmidrule(lr){2-6} \cmidrule(lr){7-12}
& \multicolumn{2}{c|}{\textbf{Claude-4}} & \textbf{Gemini-2.5} & \multicolumn{2}{c|}{\textbf{GPT-4.1}} & \multicolumn{3}{c|}{\textbf{InternVL3-38B}} & \multicolumn{3}{c}{\textbf{Qwen2.5VL-32B}} \\
\cmidrule(lr){2-3} \cmidrule(lr){4-4} \cmidrule(lr){5-6} \cmidrule(lr){7-9} \cmidrule(lr){10-12}
\textbf{Subcategory} & ZS & FS & ZS & ZS & FS & ZS & FS & FT & ZS & FS & FT \\
\midrule
Muscularity Self-objectification & 0.451 & 0.431 & 0.512 & --- & \cellcolor{cyan!25}\textbf{0.562} & 0.495 & \cellcolor{green!25}\textbf{0.578} & 0.384 & 0.349 & 0.538 & 0.391 \\
Leanness Self-objectification & 0.676 & 0.635 & 0.645 & --- & \cellcolor{cyan!25}\textbf{0.706} & 0.636 & 0.590 & 0.500 & 0.533 & \cellcolor{green!25}\textbf{0.625} & 0.561 \\
Muscle Dissatisfaction & 0.655 & 0.738 & \cellcolor{green!25}\textbf{0.774} & 0.714 & \cellcolor{cyan!25}\textbf{0.765} & 0.677 & 0.765 & 0.654 & 0.676 & 0.754 & 0.806 \\
Rigid Food Rules & \cellcolor{cyan!25}\textbf{0.721} & 0.698 & 0.714 & 0.642 & 0.618 & \cellcolor{green!25}\textbf{0.769} & 0.750 & 0.703 & 0.607 & 0.725 & 0.730 \\
Unsafe Food & 0.812 & \cellcolor{cyan!25}\textbf{0.848} & 0.833 & 0.800 & 0.812 & \cellcolor{green!25}\textbf{0.889} & 0.857 & 0.899 & 0.691 & 0.759 & 0.879 \\
Cheat Meals & \cellcolor{green!25}\textbf{0.899} & 0.781 & \cellcolor{cyan!25}\textbf{0.866} & 0.742 & 0.750 & 0.836 & 0.889 & 0.700 & 0.712 & 0.901 & 0.818 \\
Anabolic Steroids & 0.706 & 0.735 & 0.750 & 0.765 & 0.730 & 0.781 & \cellcolor{green!25}\textbf{0.831} & 0.781 & 0.754 & \cellcolor{cyan!25}\textbf{0.825} & 0.690 \\
Legal APEDs & 0.700 & \cellcolor{green!25}\textbf{0.818} & 0.667 & \cellcolor{cyan!25}\textbf{0.818} & 0.800 & 0.833 & 0.696 & 0.667 & 0.556 & 0.636 & 0.696 \\
Hormone Therapy & 0.769 & 0.794 & 0.712 & 0.788 & \cellcolor{green!25}\textbf{0.824} & 0.783 & \cellcolor{cyan!25}\textbf{0.794} & 0.778 & 0.776 & 0.794 & 0.800 \\
Excessive Exercise & 0.449 & 0.510 & 0.408 & 0.507 & \cellcolor{green!25}\textbf{0.538} & 0.433 & 0.500 & 0.600 & 0.267 & 0.459 & 0.476 \\
Maladaptive Coping & 0.712 & \cellcolor{green!25}\textbf{0.787} & 0.655 & 0.560 & \cellcolor{cyan!25}\textbf{0.759} & 0.489 & 0.471 & 0.607 & 0.419 & 0.490 & 0.635 \\
Exercise-induced Functional Impairment & \cellcolor{green!25}\textbf{0.571} & \cellcolor{cyan!25}\textbf{0.533} & 0.500 & --- & 0.500 & 0.308 & 0.533 & 0.429 & 0.333 & 0.333 & 0.353 \\
Toxic Motivation & \cellcolor{green!25}\textbf{0.607} & 0.552 & \cellcolor{cyan!25}\textbf{0.603} & 0.571 & 0.594 & 0.563 & 0.523 & 0.448 & 0.533 & 0.508 & 0.516 \\
Relationship to Masculinity & 0.779 & 0.740 & 0.621 & \cellcolor{green!25}\textbf{0.806} & 0.712 & 0.618 & 0.691 & 0.691 & 0.588 & \cellcolor{cyan!25}\textbf{0.759} & 0.649 \\
Irrelevant & 0.513 & 0.470 & 0.629 & 0.475 & 0.532 & 0.559 & 0.568 & 0.658 & 0.527 & 0.545 & \cellcolor{green!25}\textbf{0.743} \\
\bottomrule
\end{tabular}
\caption{Per-Class F1 Scores for Subcategory Classification by Model and Training Paradigm. ZS = Zero-Shot, FS = Few-Shot, FT = FineTuned. Highest scores per subcategory in \cellcolor{green!25}\textbf{green}, second-highest in \cellcolor{cyan!25}\textbf{cyan}. Dashes indicate missing data.}
\label{tab:per_class_subtype}
\end{table*}

\begin{figure*}[ht]
  \begin{tcolorbox}[
      width=\textwidth,
      colback=gray!5,
      colframe=blue!60!black,
      title=\textbf{VLM Zero-Shot/SFT Prompt - Task 1: Primary Category Classification},
      fonttitle=\bfseries,
      boxrule=1pt,
      arc=2pt,
      outer arc=2pt,
    ]
\begin{verbatim}
<video>You are an expert classifier of harmful pro-muscle-dysmorphia content
on TikTok.
Taxonomy:
Type 1: Relationship to Body:
  • Subtype 1: Muscularity Self‑Objectification:Idealized imagery emphasizing 
  muscular aesthetics.
  • Subtype 2: Leanness Self‑Objectification: Self‑evaluation against a lean ideal.
  • Subtype 3: Muscle Dissatisfaction: Perceived insufficient muscularity.

Type 2: Relationship to Food:
  • Subtype 1: Rigid Food Rules: Obsessive tracking, bulking, or cutting.
  • Subtype 2: Unsafe Food: Promoting unconventional muscle-enhancing foods.
  • Subtype 3: Cheat Meals: Binge-compensation cycles.

Type 3: Supplement Abuse:
  • Subtype 1: Anabolic Steroids: Downplaying risks of steroid use.
  • Subtype 2: Legal APEDs: Excessive legal supplement usage.
  • Subtype 3: Hormone Therapy: Promoting TRT as performance enhancer.

Type 4: Relationship to Exercise:
  • Subtype 1: Excessive Exercise despite harm.
  • Subtype 2: Predebting: Exercising to justify eating.
  • Subtype 3: Maladaptive Coping: Sole coping method.
  • Subtype 4: Functional Impairment.
  • Subtype 5: Toxic Motivation.

Type 5: Relationship to Masculinity:
  • Subtype 1: Linking muscles to male identity and worth.

Type 6: Irrelevant:
  • Subtype 1: Content unrelated to the above.

Carefully consider all sources of information about this video:

VIDEO_ID: <Anonymized Video ID>
Caption: <TikTok Video Caption>
Audio transcript: <Audio Transcriptions using Whisper>
Embedded text: <OCR text from video>

###TASK###
Classify the video into one of the following types:
Relationship to Body, Relationship to Food, Supplement Abuse,
Relationship to Exercise, Relationship to Masculinity, or
Irrelevant.

Only output the type label, no explanations or subtypes.

\end{verbatim}
  \end{tcolorbox}
\caption{Prompt used for zero-shot inference and supervised finetuning for Task 1.}
\label{fig:vlm_zeroshot_prompt_type}
\end{figure*}

\begin{figure*}[ht]
  \begin{tcolorbox}[
      width=\textwidth,
      colback=gray!5,
      colframe=blue!60!black,
      title=\textbf{VLM Zero-Shot/SFT Prompt - Task 2:  Subcategory Classification},
      fonttitle=\bfseries,
      boxrule=1pt,
      arc=2pt,
      outer arc=2pt,
    ]
\footnotesize
\begin{verbatim}
<video>You are an expert classifier of harmful pro-muscle-dysmorphia content
on TikTok.
Taxonomy:
Type 1: Relationship to Body:
  • Subtype 1: Muscularity Self‑Objectification:Idealized imagery emphasizing 
  muscular aesthetics.
  • Subtype 2: Leanness Self‑Objectification: Self‑evaluation against a lean ideal.
  • Subtype 3: Muscle Dissatisfaction: Perceived insufficient muscularity.

Type 2: Relationship to Food:
  • Subtype 1: Rigid Food Rules: Obsessive tracking, bulking, or cutting.
  • Subtype 2: Unsafe Food: Promoting unconventional muscle-enhancing foods.
  • Subtype 3: Cheat Meals: Binge-compensation cycles.

Type 3: Supplement Abuse:
  • Subtype 1: Anabolic Steroids: Downplaying risks of steroid use.
  • Subtype 2: Legal APEDs: Excessive legal supplement usage.
  • Subtype 3: Hormone Therapy: Promoting TRT as performance enhancer.

Type 4: Relationship to Exercise:
  • Subtype 1: Excessive Exercise despite harm.
  • Subtype 2: Predebting: Exercising to justify eating.
  • Subtype 3: Maladaptive Coping: Sole coping method.
  • Subtype 4: Functional Impairment.
  • Subtype 5: Toxic Motivation.

Type 5: Relationship to Masculinity:
  • Subtype 1: Linking muscles to male identity and worth.

Type 6: Irrelevant:
  • Subtype 1: Content unrelated to the above.

Carefully consider all sources of information about this video:

VIDEO_ID: <Anonymized Video ID>
Caption: <TikTok Video Caption>
Audio transcript: <Audio Transcriptions using Whisper>
Embedded text: <OCR text from video>

###TASK###
Classify the video into one of the following subtypes: Muscularity Self‑Objectification, 
Leanness Self‑Objectification, 
Muscle Dissatisfaction, Rigid Food Rules, 
Unsafe Food, Cheat Meals, 
Anabolic Steroids, Legal APEDs, 
Hormone Therapy, Excessive Exercise, 
Predebting, Maladaptive Coping, 
Exercise‑Induced Functional Impairment, 
Toxic Motivation, 
Relationship to Masculinity, or Irrelevant.

Only output the specific subtype label, no explanations or other text. 
If none apply, output Irrelevant.
Use the exact subtype names, not the type names.


\end{verbatim}
  \end{tcolorbox}
\caption{Prompt used for zero-shot inference and supervised finetuning for Task 2.}
\label{fig:vlm_zeroshot_prompt_subtype}
\end{figure*}

\begin{figure*}[ht]
  \begin{tcolorbox}[
      width=\textwidth,
      colback=gray!5,
      colframe=blue!60!black,
      title=\textbf{VLM Few-Shot Prompt. Task 1: Primary Category Classification},
      fonttitle=\bfseries,
      boxrule=1pt,
      arc=2pt,
      outer arc=2pt,
    ]
\scriptsize
\begin{verbatim}
You are an expert classifier of harmful pro-muscle-dysmorphia content on TikTok.
Taxonomy:
Type 1: Relationship to Body:
  • Subtype 1: Muscularity Self‑Objectification: Idealized imagery emphasizing muscular 
  aesthetics as the primary source of value.
  • Subtype 2: Leanness Self‑Objectification: Self‑evaluation against a lean, 
  low‑fat, highly‑toned ideal.
  • Subtype 3: Muscle Dissatisfaction: Expressing perceived insufficient muscularity despite 
  having a muscular physique.
Type 2: Relationship to Food:
  • Subtype 1: Rigid Food Rules: Obsessive macro/micronutrient tracking 
  and restrictive dietary practices to rapidly gain muscle mass (bulking) or lose fat (cutting).
  • Subtype 2: Unsafe Food: Promotion of unconventional foods believed 
  to enhance muscle growth.
  • Subtype 3: Cheat Meals: Large "reward" meals after restrictive dieting 
  that reinforce binge–compensation cycles.
Type 3: Supplement Abuse:
  • Subtype 1: Anabolic Steroids: Normalization or 
  endorsement of anabolic‑androgenic steroid use with downplayed risks.
  • Subtype 2: Legal APEDs: Overuse of legal supplements (e.g., creatine, 
  protein, pre-workout) beyond recommended doses.
  • Subtype 3: Hormone Therapy: Downplaying risks and spreading misinformation 
  about testosterone replacement therapy, promoted as a performance enhancer or "anti-aging" treatment 
  without proper medical diagnosis.
Type 4: Relationship to Exercise:
  • Subtype 1: Excessive Exercise: Extreme exercise routines exceeding healthy limits despite 
  injury or life interference.
  • Subtype 2: Predebting: Treating exercise as punishment or permission to eat.
  • Subtype 3: Maladaptive Coping: Using exercise as the sole coping mechanism to avoid 
  emotional distress.
  • Subtype 4: Exercise‑Induced Functional Impairment: Prioritizing exercise 
  over essential duties, harming daily functioning.
  • Subtype 5: Toxic Motivation: Demeaning communication that pressures 
  unrealistic fitness standards via shaming or slurs.
Type 5: Relationship to Masculinity:
  • Subtype 1: Linking muscle‑building and exercise performance to male identity, 
  sexuality, and self‑worth.
Type 6: Irrelevant:
  • Subtype 1: Content without muscle‑obsession, restrictive 
  diets, supplement/AAS promotion, or extreme exercise. Includes dance trends, 
  memes, travel/cooking vlogs, general wellness (e.g., yoga), and pure entertainment unrelated 
  to body‑image narratives.
Here are some examples from the training data:
Example 1:
<video>
Caption: <Video Caption>
Audio transcript: <Audio Transcription from Whisper>
Embedded text: <OCR Text>
Classification: <Type>
.
.
.
Example 12:
<video>
Caption: <Video Caption>
Audio transcript: <Audio Transcription from Whisper>
Embedded text: <OCR Text>
Classification: <Type>
<video>
VIDEO_ID: 
Caption: 
Audio transcript: 
Embedded text: 

###TASK###
Classify the video into one of the following types: Relationship to Body, Relationship to Food, 
Supplement Abuse, Relationship to Exercise, Relationship to Masculinity, or Irrelevant. 
Only output the type label, no explanations or other text. If none apply, output Irrelevant.
Don't use a subtype, such as "Muscularity Self‑Objectification" or "Unsafe Food".
Valid outputs: Relationship to Body, Relationship to Food, Supplement Abuse, Relationship to Exercise, 
Relationship to Masculinity, Irrelevant
\end{verbatim}
  \end{tcolorbox}
\caption{Prompt used for few-shot inference for Task 1.}
\label{fig:vlm_fewshot_prompt_type}
\end{figure*}

\begin{figure*}[ht]
  \begin{tcolorbox}[
      width=\textwidth,
      colback=gray!5,
      colframe=blue!60!black,
      title=\textbf{VLM Few-Shot Prompt. Task 2:  Subcategory Classification},
      fonttitle=\bfseries,
      boxrule=1pt,
      arc=2pt,
      outer arc=2pt,
    ]
\scriptsize
\begin{verbatim}
You are an expert classifier of harmful pro-muscle-dysmorphia content on TikTok.
Taxonomy:
Type 1: Relationship to Body:
  • Subtype 1: Muscularity Self‑Objectification: Idealized imagery emphasizing muscular 
  aesthetics as the primary source of value.
  • Subtype 2: Leanness Self‑Objectification: Self‑evaluation against a lean, 
  low‑fat, highly‑toned ideal.
  • Subtype 3: Muscle Dissatisfaction: Expressing perceived insufficient muscularity despite 
  having a muscular physique.
Type 2: Relationship to Food:
  • Subtype 1: Rigid Food Rules: Obsessive macro/micronutrient tracking 
  and restrictive dietary practices to rapidly gain muscle mass (bulking) or lose fat (cutting).
  • Subtype 2: Unsafe Food: Promotion of unconventional foods believed 
  to enhance muscle growth.
  • Subtype 3: Cheat Meals: Large "reward" meals after restrictive dieting 
  that reinforce binge–compensation cycles.
Type 3: Supplement Abuse:
  • Subtype 1: Anabolic Steroids: Normalization or 
  endorsement of anabolic‑androgenic steroid use with downplayed risks.
  • Subtype 2: Legal APEDs: Overuse of legal supplements (e.g., creatine, 
  protein, pre-workout) beyond recommended doses.
  • Subtype 3: Hormone Therapy: Downplaying risks and spreading misinformation 
  about testosterone replacement therapy, promoted as a performance enhancer or "anti-aging" treatment 
  without proper medical diagnosis.
Type 4: Relationship to Exercise:
  • Subtype 1: Excessive Exercise: Extreme exercise routines exceeding healthy limits despite 
  injury or life interference.
  • Subtype 2: Predebting: Treating exercise as punishment or permission to eat.
  • Subtype 3: Maladaptive Coping: Using exercise as the sole coping mechanism to avoid 
  emotional distress.
  • Subtype 4: Exercise‑Induced Functional Impairment: Prioritizing exercise 
  over essential duties, harming daily functioning.
  • Subtype 5: Toxic Motivation: Demeaning communication that pressures 
  unrealistic fitness standards via shaming or slurs.
Type 5: Relationship to Masculinity:
  • Subtype 1: Linking muscle‑building and exercise performance to male identity, 
  sexuality, and self‑worth.
Type 6: Irrelevant:
  • Subtype 1: Content without muscle‑obsession, restrictive 
  diets, supplement/AAS promotion, or extreme exercise. Includes dance trends, 
  memes, travel/cooking vlogs, general wellness (e.g., yoga), and pure entertainment unrelated 
  to body‑image narratives.
Here are some examples from the training data:
Example 1:
<video>
Caption: <Video Caption>
Audio transcript: <Audio Transcription from Whisper>
Embedded text: <OCR Text>
Classification: <Type>
.
.
.
Example 12:
<video>
Caption: <Video Caption>
Audio transcript: <Audio Transcription from Whisper>
Embedded text: <OCR Text>
Classification: <Type>
<video>
VIDEO_ID: 
Caption: 
Audio transcript: 
Embedded text: 

###TASK###
Classify the video into one of the following subtypes: Muscularity Self‑Objectification, 
Leanness Self‑Objectification, 
Muscle Dissatisfaction, Rigid Food Rules, 
Unsafe Food, Cheat Meals, 
Anabolic Steroids, Legal APEDs, 
Hormone Therapy, Excessive Exercise, 
Predebting, Maladaptive Coping, 
Exercise‑Induced Functional Impairment, 
Toxic Motivation, 
Relationship to Masculinity, or Irrelevant.

\end{verbatim}

  \end{tcolorbox}
\caption{Prompt used for few-shot inference for Task 2.}
\label{fig:vlm_fewshot_prompt_subtype}
\end{figure*}

\begin{table*}[ht]
\centering
\small
\begin{tabular}{p{0.03\textwidth}|p{0.05\textwidth}|p{0.65\textwidth}|p{0.1\textwidth}}
\toprule
\textbf{ID} & \textbf{Batch} & \textbf{Expertise} & \textbf{Gender} \\
\midrule
A1  & 1 & clinical psychology doctoral candidate specializing in male eating disorders and muscle dysmorphia intervention research. & Male \\
A2  & 1 & senior researcher in computational social science focused on social media dynamics and their implications for mental health. & Female \\
\cmidrule(lr){1-4}
A3  & 2 & health communication professor using computational social network methods to study online health behaviors. & Female \\
A4  & 2 & medical student focusing on clinical and translational research in body image disorders and digital health interventions. & Female \\
\cmidrule(lr){1-4}
A5  & 3 & clinical psychology doctoral candidate specializing in digital mental health and disordered eating prevention. & Female \\
A6  & 3 & communication doctoral candidate investigating how interactive media–driven social comparisons affect body image outcomes. & Female \\
\cmidrule(lr){1-4}
A7  & 4 & clinical professor and psychiatrist studying the psychopathology and treatment of eating disorders and muscle dysmorphia. & Male \\
A8  & 4 & clinical professor and pediatric psychologist specializing in adolescent body image disorders. & Female \\
\cmidrule(lr){1-4}
A9  & 5 & child and adolescent psychiatrist specializing in developmental psychopathology and body image disorders. & Female \\
A10 & 5 & clinical social worker specializing in child and adolescent body image disorders and psychosocial interventions in neuropsychiatric care. & Female \\
\cmidrule(lr){1-4}
A11 & 6 & occupational therapist specializing in functional rehabilitation and psychosocial support for eating‐disorder patients in a neuropsychiatric hospital setting. & Female \\
A12 & 6 & clinical dietitian specializing in nutritional assessment and dietary management for eating and body-image disorders. & Female \\
\cmidrule(lr){1-4}
A13 & 7 & mental health nurse practitioner specializing in the assessment and treatment of child and adolescent psychiatric disorders. & Female \\
A14 & 7 & board-certified child and adolescent psychiatrist specializing in mood and behavioral disorders. & Female \\
\cmidrule(lr){1-4}
A15 & 8 & clinical social worker with inpatient psychiatric experience in providing psychotherapy to adult patients. & Female \\
A16 & 8 & doctoral candidate specializing in detecting disordered behaviors in social media. & Male \\
\bottomrule
\end{tabular}
\caption{Annotator assignments, areas of expertise, and gender.}
\label{tab:annotators}
\end{table*}

\begin{table*}[!ht]
\centering
\footnotesize
\begin{tabular}{p{0.08\textwidth}|p{0.15\textwidth}|p{0.15\textwidth}|p{0.22\textwidth}|p{0.3\textwidth}}
\toprule
\textbf{Video ID} & \textbf{Subcategory} & \textbf{Video Description} & \textbf{Embedded Text} & \textbf{Audio Transcript} \\
\midrule
001 & Relationship to Body 
  & my lil frame 
  & me being a 5'8 male but with lean body 
  & and my little frame and my sweet little girl voice. It exudes something in people that... \\
\hline
002 & Relationship to Masculinity 
  & \#menshealth \#menshealthtips \#cynic 
  & @<redacted> 
  & The reason that there is such a void of male role models for young men is because we literally do not have adult males in our population anymore. What do I mean by that? There's no adult males. Like, there's grown men everywhere, but they're not actually physiologically adult human males... \\
\hline
003 & Supplement Abuse 
  & this is totally healthy right? \#recipes \#creatine 
  & @<redacted>
  & So I told a co-worker I was just taking my creatine in water, just water, because it was unflavored. And they told me you're supposed to take it with like a meal or like sugar so then you can absorb it better... \\
\hline
004 & Relationship to Food 
  & <redacted> dinner! 
  & @<redacted> 
  & Pulled beef with hard shelled eggs Mmm It's crazy. It actually tastes like the delicious pulled beef That's in with the garlic oh and the onion and the lime juice booyah, yeah Sweet potatoes, the purple variety... \\
\hline
005 & Relationship to Exercise 
  & I am girl 
  & MAKING THE GYM YOUR WHOLE PERSONALITY no train legs 
  & girls who make going to the gym their entire personality it is just so crazy to me I genuinely do not know how you guys do it I could never be one of those girls anyway someone I know just asked me what day it was and I replied with leg day \\
\hline
006 & Irrelevant 
  & Two in \#shorts \#gymshorts 
  & @<redacted> 
  & ugly, ugly, ugly aura. Bro, if your shorts, especially if you're hitting leg day, are below your knees, you're genuinely cooked, bro. These, I got them here on the TikTok shop. This is a size medium. I'm five foot 10, 165 pounds, and these things fit so good... \\
\bottomrule
\end{tabular}
\caption{Example entries for Type classification showing captions and TikTok metadata. The actual videos can be found in the data attachment in the submission platform.}
\label{tab:type_classification}
\end{table*}

\begin{table*}[!ht]
\centering
\small
\begin{tabular}{p{0.08\textwidth}|p{0.15\textwidth}|p{0.20\textwidth}|p{0.22\textwidth}|p{0.25\textwidth}}
\toprule
\textbf{Video ID} & \textbf{Subcategory} & \textbf{Video Description} & \textbf{Embedded Text} & \textbf{Audio Transcript} \\
\midrule
007 & Muscularity Self-objectification 
  & training hard or hardly training ? 
  & sounds of my workout on push day 
  & Yeah, we need some weight, should be a good one today. Bet you gotta hit me fast, you know old... \\

008 & Leanness Self-objectification 
  & my lil frame 
  & me being a 5'8 male but with lean body 
  & and my little frame and my sweet little girl voice. It exudes something in people that... \\

009 & Muscle Dissatisfaction 
  & \#gymtok 
  & When u start working out to look better but now u have body dysmorphia 
  & fucking go let's go I guess \\

\cmidrule(lr){1-5}
010 & Rigid Food Rules 
  & I gained 35lbs in 90 days \#lifting \#gym \#bulking 
  & 35lbs in 90 days I've considered ever since 7th grade but change 
  & I gained 35 pounds in 90 days. I've considered myself a runner ever since 7th grade... \\

011 & Unsafe Food 
  & Liver King Dinner to start the week 
  & <redacted> 
  & Mmm, steak and potatoes, sweet potato fritters, Liver King Chef Lionel, that's how he calls them. So good. We got... \\

012 & Cheat Meals 
  & NO MORE.. \#gym \#cheatday \#gains 
  & us after cheat day turned into cheat month 
  & No more cheese fillets. No more McDonald's. No more chicken wings. No more chicken snobby with some sour. No more... \\

013 & Excessive Exercise 
  & Gym rats! \#gym \#gymtok \#mothersoftiktok \#thick \#newyearsresolutions \#muscles \#gains \#proteinshake \#lawenforcement 
  & It's time for us normal gym peeps to get our routines back!!! 
  & Listen, I got my ass the gym today and I didn't want to go. So did you go today? Well... \\

\cmidrule(lr){1-5}
014 & Predebting Exercise 
  & You CANNOT our train a bad diet \#diet \#fitness \#exercise 
  & I CAN EAT WHATEVER I WANT AS LONG AS I EXERCISE YUP YOU HEARD ME WHILE EXERCISE IS IMPORTANT... 
  & I can eat whatever I want as long as I exercise. You said what? Yup, you heard me. While exercise... \\

015 & Maladaptive Coping 
  & Gym Therapy \#gymtherapy \#gym \#doordie \#dobetter \#military 
  & @<redacted> 
  & Gym therapy is a term that was coined by Shredded Vets that describes the psychological effects of working out consistently... \\

016 & Exercise-Induced Functional Impairment 
  & @<redacted> - I would skip school just to go to the gym. I was going to do whatever it took. 
  & I would skip school to guru the gym. 
  & lucky enough like when I was in year 9 and year 10 the Knights were giving me a day off... \\

017 & Toxic Motivation 
  & put in the effort \#gym \#motivation \#fitnessmotivation \#workout 
  & TD PRESSURE 
  & It's fucking different. Gotta really come kill me. I'm built different. I train different. I work different. I am different... \\

\cmidrule(lr){1-5}
018 & Anabolic Steroids 
  & \#<redacted> 
  & POV: Steroids gives you acne HAVE FUCKING I MAN LIKE PIMPLE SHIT COLLAB MIGHT PIMPLE UH ASIAN 
  & dude I think I have like a fucking I got like a cyst man you know like where it's a... \\

019 & Legal APEDs 
  & Sustenanceee \#fyp \#xyzbca \#gymtok \#creatine 
  & POV: you get caught using creatine 
  & Are there any drugs in this house? If there are, you better find them and give them to me immediately... \\

020 & Hormone Therapy 
  & Day 5 on TRT \#trt \#lift \#weightlifting \#gym \#workout 
  & Day 5 on TRT 
  & What up guys, it's day five. So on day five, the doctor calls you with your full blood work done... \\

\bottomrule
\end{tabular}
\caption{Example entries for Subtype classification showing captions and TikTok metadata. The actual videos can be found in the data attachment in the submission platform.}
\label{tab:subtype}
\end{table*}

\end{document}